\documentclass[twoside,11pt]{article}
\usepackage[preprint]{jmlr2e}
\usepackage{lastpage}
\jmlrheading{27}{2026}{1-\pageref{LastPage}}{11/30}{00/00}{26-0000}{Hongyi Tao, Dingzhi Yu, Difan Zou, and Lijun Zhang}
\ShortHeadings{When and Why Muon Wins}{Tao, Yu, Zou, and Zhang}
\firstpageno{1}
\usepackage[utf8]{inputenc}
\usepackage{amsfonts}
\usepackage{nicefrac}
\usepackage{microtype}
\usepackage[normalem]{ulem}
\usepackage{tikz}
\usetikzlibrary{calc,arrows.meta,decorations.pathreplacing,shapes}
\usepackage{enumitem}
\setlist[itemize]{leftmargin=1.8em,topsep=3pt,itemsep=2pt,parsep=0pt,partopsep=0pt}
\setlist[enumerate]{leftmargin=1.8em,topsep=3pt,itemsep=2pt,parsep=0pt,partopsep=0pt}
\usepackage{graphicx}
\usepackage{url}
\usepackage{amsmath}
\usepackage{amssymb}
\usepackage{hyperref}
\usepackage[caption=false]{subfig}
\usepackage[capitalize,noabbrev]{cleveref}
\AddToHook{env/lemma/begin}{\crefalias{theorem}{lemma}}
\AddToHook{env/proposition/begin}{\crefalias{theorem}{proposition}}
\AddToHook{env/remark/begin}{\crefalias{theorem}{remark}}
\AddToHook{env/corollary/begin}{\crefalias{theorem}{corollary}}
\AddToHook{env/definition/begin}{\crefalias{theorem}{definition}}
\AddToHook{env/conjecture/begin}{\crefalias{theorem}{conjecture}}
\AddToHook{env/axiom/begin}{\crefalias{theorem}{axiom}}
\crefname{lemma}{Lemma}{Lemmas}
\crefname{proposition}{Proposition}{Propositions}
\crefname{remark}{Remark}{Remarks}
\crefname{corollary}{Corollary}{Corollaries}
\crefname{definition}{Definition}{Definitions}
\crefname{conjecture}{Conjecture}{Conjectures}
\crefname{axiom}{Axiom}{Axioms}
\usepackage{appendix}
\usepackage{apptools}
\usepackage{interval}
\usepackage{fancybox}
\usepackage{algorithm}
\usepackage{algorithmic}

\renewenvironment{proof}[1][Proof]{\par\vspace{1ex}\noindent{\bf #1.\ }\ignorespaces}{\hfill\BlackBox\\[2mm]}

\newtheorem{assumption}{Assumption}

\newcommand{\AssumptionSeries}[1]{%
  \renewcommand{\theassumption}{\arabic{assumption}#1}%
  \renewcommand{\theHassumption}{\arabic{assumption}#1}%
  \setcounter{assumption}{0}}

\makeatletter
\newcommand{\MatrixNoiseFloatLayout}{%
  \setlength{\@fptop}{0pt}%
  \setlength{\@fpbot}{0pt plus 1fil}}
\makeatother

\usepackage{tcolorbox}
\definecolor{comparisonblue}{HTML}{668FBC}
\definecolor{comparisongold}{HTML}{CF951F}
\newtcolorbox{comparisonpanel}[2][]{
    nobeforeafter,width=0.485\linewidth,
    colback=#2!10!white,colframe=#2,
    boxrule=0.5pt,arc=2mm,boxsep=0pt,
    left=4pt,right=4pt,top=4pt,bottom=4pt,
    halign=center,
    before upper={\setlength{\parskip}{0pt}%
        \setlength{\abovedisplayskip}{4pt}%
        \setlength{\abovedisplayshortskip}{4pt}%
        \setlength{\belowdisplayskip}{2pt}%
        \setlength{\belowdisplayshortskip}{2pt}},
    #1
}
\newenvironment{formulacomparison}
    {\par\addvspace{6pt}\setlength{\parskip}{0pt}%
        \nointerlineskip\noindent\ignorespaces}
    {\par\addvspace{6pt}\nointerlineskip\ignorespacesafterend}

\usepackage{booktabs}
\usepackage{threeparttable}
\usepackage{xcolor}         
\usepackage{colortbl}
\usepackage{makecell}
\usepackage{bbding}
\definecolor{customgray}{gray}{0.9}
\usepackage{xcolor}

\usepackage{bm}
\usepackage[all]{hypcap}

\newcommand{\op}{\textnormal{op}}
\DeclareMathOperator{\tr}{Tr}
\DeclareMathOperator{\rank}{rank}
\DeclareMathOperator{\erank}{erank}
\DeclareMathOperator{\diag}{diag}
\DeclareMathOperator{\diagvec}{diagvec}
\DeclareMathOperator{\spann}{span}

\DeclareMathOperator*{\argmax}{arg\,max}

\DeclareMathOperator*{\sgn}{sign}
\DeclareMathOperator*{\msgn}{msign}

\newcommand{\R}{\mathbb{R}}

\newcommand{\vzero}{\mathbf{0}}

\newcommand{\brac}[1]{\left(#1\right)}

\newcommand{\sqbrac}[1]{\left[#1\right]}
\newcommand{\bbrac}[1]{\left \{ #1 \right\}}
\newcommand{\abs}[1]{\left|#1\right|}
\newcommand{\norm}[1]{\left\lVert#1\right\rVert}
\newcommand{\inner}[2]{\left\langle #1, #2 \right\rangle}
\newcommand{\ud}{\,\mathrm{d}}

\newcommand{\Diag}[1]{\diag\left(#1\right)}
\newcommand{\Diagvec}[1]{\diagvec\left(#1\right)}

\newcommand{\expect}[1]{\E\sqbrac{#1}}
\newcommand{\bigO}[1]{\mathcal{O}\left(#1\right)}
\newcommand\vsigma{{\bm{\sigma}}}
\newcommand\vSigma{{\bm{\Sigma}}}
\newcommand{\msign}[1]{\msgn\left(#1\right)}
\newcommand{\sign}[1]{\sgn\left(#1\right)}

\def \y {\mathbf{y}}
\def \YB {\mathbf{Y}}
\def \E {\mathbb{E}}
\def \x {\mathbf{x}}
\def \XB {\mathbf{X}}

\def \v {\mathbf{v}}
\def \V {\mathbf{V}}
\def \g {\mathbf{g}}
\let\PolishL\L
\def \L {\ifmmode\mathbf{L}\else\PolishL\fi}

\def \Db {\mathbf{D}}
\def \O {\mathcal{O}}

\def \z {\mathbf{z}}

\def \u {\mathbf{u}}
\def \U {\mathbf{U}}

\def \H {\mathcal{H}}
\def \w {\mathbf{w}}
\def \WB {\mathbf{W}}

\def \PB {\mathbf{P}}
\def \F {\mathcal{F}}

\def \A {\mathcal{A}}
\def \Ab {\mathbf{A}}
\newcommand{\q}{\mathbf{q}}

\def \p {\mathbf{p}}

\def \Ib {\mathbf{I}}

\def \a {\mathbf{a}}
\def \b {\mathbf{b}}

\def \SBB {\mathbb{S}}

\def \Gb {\mathbf{G}}

\def \s {\mathbf{s}}
\def \SB {\mathbf{S}}

\def \bta {\bm \delta}

\def \m {\mathbf{m}}
\def \MB {\mathbf{M}}

\def \Rt {\mathcal{R}}

\def \QB {\mathbf{Q}}

\def \e {\mathbf{e}}
\def \EB {\mathbf{E}}

\def \N {\mathbb{N}}
\def \Nb {\mathbf{N}}

\def \Pr    {\mathbb{P}}

\def \SumT {\sum_{t=1}^T}

\def \Linf {L_{\infty}}

\crefname{assumption}{Assumption}{Assumptions}
\crefname{lemma}{Lemma}{Lemmas}
\crefname{theorem}{Theorem}{Theorems}
\crefname{appendix}{Appendix}{Appendices}
\crefname{subtheorem}{Theorem}{Theorems}
\crefname{algorithm}{Algorithm}{Algorithms}

\definecolor{muonlink}{HTML}{2E649E}
\definecolor{sgdmlink}{HTML}{A67314}
\newcommand{\MuonRef}[1][Muon]{\hyperref[alg:muon-step]{\textcolor{muonlink}{#1}}}
\newcommand{\SGDMRef}[1][SGDM]{\hyperref[alg:sgdm-step]{\textcolor{sgdmlink}{#1}}}
\newcommand{\SGDRef}[1][SGD]{\hyperref[alg:sgdm-step]{\textcolor{sgdmlink}{#1}}}

\begin{document}

\title{When and Why Muon Wins:\\ A Complexity-Theoretic Separation from SGD}

\author{\name Hongyi Tao\thanks{Equal contribution.} \email taohy@lamda.nju.edu.cn \\
  \name Dingzhi Yu\footnotemark[1] \email yudz@lamda.nju.edu.cn \\
  \addr State Key Laboratory of Novel Software Technology, Nanjing University \\ School of Artificial Intelligence, Nanjing University\\
  \name Difan Zou \email dzou@cs.hku.hk \\
  \addr School of Computing and Data Science, The University of Hong Kong \\
  \name Lijun Zhang \email zhanglj@lamda.nju.edu.cn \\
  \addr State Key Laboratory of Novel Software Technology, Nanjing University \\ School of Artificial Intelligence, Nanjing University
}

\editor{To be assigned}

\maketitle

\begin{abstract}
Matrix-sign optimizers such as Muon have demonstrated remarkable empirical gains in large language model (LLM) pretraining, yet a theoretical explanation for their superiority over standard Stochastic Gradient Descent (SGD) remains elusive. A fundamental barrier is that under classical Euclidean smoothness and finite variance assumptions, SGD is already minimax optimal, leaving no room for any worst-case complexity improvement. We show that a strict complexity separation emerges once the optimization framework aligns with the matrix structure of neural network parameters. Specifically, we study smooth matrix optimization under the operator norm, measure stationarity via the nuclear norm, and bound the covariance of stochastic gradient noise by a positive semidefinite matrix $\vSigma^2$. Under this formulation, we establish matching upper and lower complexity bounds for Muon without explicit dimension factors. By contrast, we prove matching upper and lower bounds for both vanilla SGD and SGD with momentum (SGDM), showing that Euclidean updates inevitably suffer dimension penalties. In particular, SGD and SGDM incur an order-$m$ slowdown in the deterministic term and an order-$m\chi(\vSigma)$ slowdown in the stochastic term, where $m$ is the smaller matrix dimension and $\chi(\vSigma) = m\norm{\vSigma}_F^2/\norm{\vSigma}_*^2 \in [1, m]$ characterizes the spectral concentration of the gradient noise. Consequently, Muon achieves a provable complexity speedup over SGD and SGDM ranging from $\Theta(m)$ under isotropic noise to $\Theta(m^2)$ under spectrally concentrated noise. Extensive experiments across CNNs and LLMs up to 7B scale confirm that LLM layers exhibit strong spectral noise concentration, where Muon substantially accelerates pretraining, whereas CNN layers remain nearly isotropic, where the performance gap indeed vanishes. Together, we provide a rigorous theoretical and empirical characterization of \emph{when and why} Muon outperforms SGD.
\end{abstract}


\section{Introduction}

Scalable stochastic optimization underpins the training of modern foundation models~\citep{devlin-etal-2019-bert,brown2020gpt3,achiam2023gpt4,touvron2023llama,team2023gemini}. Recently, the matrix-sign optimizer Muon~\citep{jordan2024muon} has emerged as a compelling alternative to AdamW~\citep{loshchilov2019adamw} and Stochastic Gradient Descent (SGD)~\citep{robbins1951stochastic}, delivering substantial speedups in large language model (LLM) pretraining~\citep{liu2025muon,shah2025practical,wen2025fantastic,semenov2025benchmarking,team2025kimi,zeng2025glm,kimi2026kimi2.5,zeng2026glm,cheng2026ngram,deepseekai2026deepseekv4}. Motivated by these empirical successes, a growing body of work has sought to theoretically understand the convergence behavior of Muon~\citep{li2025note,chen2025muon,qian2025muon,chang2025convergence,pan2025unbiased,kim2026convergence,shulgin2026beyond,yu2026signheavytails}. However, existing analyses focus primarily on establishing convergence upper bounds, leaving it theoretically unclear whether Muon can provably outperform SGD, and if so, under what conditions such a separation holds.

Establishing such a separation faces a fundamental obstacle. Under the standard Euclidean formulation of smooth non-convex stochastic optimization, where gradient noise is bounded by a single scalar variance $\sigma^2$ that treats all directions as isotropic, vanilla SGD already achieves the minimax optimal $\bigO{\epsilon^{-4}}$ complexity~\citep{ghadimi2013stochastic,arjevani2023lower}. Consequently, any theoretical explanation for Muon's advantage must move beyond this isotropic Euclidean setting and identify the structural properties of deep network weight matrices that Muon exploits. Related progress in vector optimization shows that coordinate-wise sign and adaptive methods (such as SignSGD and AdaGrad) can provably improve over vanilla SGD under $\ell_\infty$-smoothness and coordinate-wise noise~\citep{jiang2024convergence,tao2026when}. However, those comparisons are restricted to memoryless SGD without momentum, whereas Muon inherently incorporates momentum. Extending this perspective to matrix optimization therefore requires addressing two central questions, namely how the matrix-sign update interacts with the geometry of weight matrices and gradient noise, and whether momentum alone allows SGDM to match this advantage.

\subsection{Problem Geometry and Structured Matrix Noise}

To capture the matrix structure of neural network layers, we consider minimizing a differentiable, lower-bounded objective $F:\R^{m\times n}\to\R$ with $m\le n$ and initial objective gap $\Delta = F(\WB_1) - \inf F$. Muon updates weight matrices along the orthogonalized polar factor (matrix sign) of the momentum state. For a matrix $\Gb\in\R^{m\times n}$ of rank $r$ with reduced singular value decomposition $\Gb=\U_r\SB_r\V_r^\top$, its matrix sign is $\msign{\Gb}=\U_r\V_r^\top$. Geometrically, $\msign{\Gb}$ is the steepest descent direction under the operator norm $\norm{\cdot}_{\op}$, as it solves the linear maximization problem
\begin{equation}
\msign{\Gb}
\in
\argmax_{\norm{\Db}_{\op}\le1}
\inner{\Gb}{\Db},\qquad \text{with optimal value } \norm{\Gb}_*.
\end{equation}
This duality between the operator and nuclear norms~\citep{pethick2025training,bernstein2025modular} naturally motivates the geometry of our problem class. We assume $L_*$-smoothness with respect to the operator norm (\cref{ass:operator norm smooth}) and measure $\epsilon$-stationarity via the nuclear norm of the gradient, requiring $\expect{\min_{1\le t\le T}\norm{\nabla F(\WB_t)}_*}\le\epsilon$ after $T$ stochastic-gradient queries.

Equally important is how the stochastic gradient noise $\Nb_t=\Gb_t-\nabla F(\WB_t)\in\R^{m\times n}$ is modeled. A conventional finite-variance bound $\expect{\norm{\Nb_t}_F^2\mid\F_{t-1}}\le\sigma^2$ collapses the entire singular-value spectrum of the noise into a single number, effectively assuming an isotropic noise structure. Motivated by extensive empirical evidence that gradient noise in deep learning is highly anisotropic~\citep{sagun2017empirical,zhu19anisotropic,zhang2020whyheavytail,pan2022eigencurve,wu2022alignment,panICLR2024momentum,xie2025tale,yu2026signheavytails,tao2026when}, we bound the noise covariance by a prescribed positive semidefinite matrix $\vSigma\in\SBB_+^m$ (\cref{ass:Matrix-variance}) through
\begin{equation}
\expect{\Nb_t\Nb_t^\top\mid\F_{t-1}}\preceq\vSigma^2.
\end{equation}
While taking the trace recovers the total variance $\norm{\vSigma}_F^2$, this matrix inequality preserves the directional singular spectrum of the noise. For a nonzero noise-scale matrix $\vSigma$, we measure its \emph{spectral concentration} by
\begin{equation}\label{eq:intro-spectral-concentration}
  \chi(\vSigma)
  = \frac{m}{\erank(\vSigma)}
  = \frac{m\norm{\vSigma}_F^2}{\norm{\vSigma}_*^2}
  \in [1,m],
\end{equation}
where $\erank(\vSigma) := \norm{\vSigma}_*^2 / \norm{\vSigma}_F^2$ denotes the effective rank~\citep{bartlett2020benign}. Thus $\chi(\vSigma)$ equals $1$ when the noise is isotropic ($\erank(\vSigma)=m$, a flat spectrum) and approaches $m$ when the noise is strongly anisotropic ($\erank(\vSigma)\ll m$, concentrated along a few singular directions).

\subsection{Main Contributions}

Comparing Muon, vanilla SGD, and SGDM on this shared problem class under identical oracle assumptions and the same nuclear-norm stationarity criterion, we establish the following tight complexities.
\begin{formulacomparison}
\begin{comparisonpanel}{comparisonblue}
    \textbf{Muon}
    \[
    \Theta\brac{
        \frac{L_*\Delta}{\epsilon^2}
        +\frac{\textcolor{red}{\norm{\vSigma}_*^2}L_*\Delta}{\epsilon^4}
    }
    \]
\end{comparisonpanel}\hfill
\begin{comparisonpanel}{comparisongold}
    \textbf{SGD / SGDM}
    \[
    \Theta\brac{
        \frac{\textcolor{red}{m}L_*\Delta}{\epsilon^2}
        +\frac{\textcolor{red}{m^2\norm{\vSigma}_F^2}L_*\Delta}{\epsilon^4}
    }
    \]
\end{comparisonpanel}
\end{formulacomparison}
Specifically, our main contributions are three-fold.
\begin{enumerate}
    \item \textbf{Tight complexity bounds for Muon without explicit dimension factors.} We prove matching upper and lower complexity bounds for Muon (\cref{thm:muon_upper_infty,thm:muon_lower_infty}). Our upper bound depends on the noise through $\norm{\vSigma}_*^2$ without explicit dimension factors, strictly improving over existing bounds under scalarized noise assumptions~\citep{shen2025convergence,hubler2026free,zhang2026scale}. To establish the matching lower bound in the presence of momentum, we construct a \emph{3D rotating-gradient hard instance} for SignSGD in which rotating gradient components cancel in the momentum buffer, and then lift this construction to Muon. Notably, our lower bound holds for any fixed matrix dimension $m\ge3$ without requiring high-dimensional constructions where the dimension grows as $\epsilon$ decreases~\citep{arjevani2023lower}.

    \item \textbf{Sharp lower bounds for SGD and SGDM, establishing a $\Theta(m)$ to $\Theta(m^2)$ separation.} We establish matching upper and lower complexity bounds for vanilla SGD (\cref{thm:matrix-sgd-all-horizon}) and extend them to SGDM (\cref{thm:direct-matrix-sgdm,thm:direct-sgdm-upper}). By constructing a direct matrix-domain hard instance, we prove that Euclidean updates incur an unavoidable factor of $m$ in the deterministic term and $m^2\norm{\vSigma}_F^2$ in the stochastic term, and that adding momentum cannot close this gap. Contrasting these tight bounds reveals both \emph{why} and \emph{when} Muon wins. Eliminating the matrix dimension $m$ from the complexity gives Muon a baseline $\Theta(m)$ speedup even under isotropic noise ($\chi(\vSigma)=1$), while spectral concentration further amplifies its stochastic advantage by $\chi(\vSigma)$, yielding up to a $\Theta(m^2)$ speedup under strongly anisotropic noise ($\chi(\vSigma)\asymp m$).

    \item \textbf{Multi-scale empirical validation up to 7B scale.} We bridge theory and practice through systematic pretraining experiments and direct spectral noise measurements across CNNs (ResNet-20 on CIFAR-10), LLMs across three scales (nanoGPT-124M, LLaMA-130M, and LLaMA-1.34B on C4), and open checkpoints of Pythia-6.9B~\citep{biderman2023pythia}. Across architectures, empirical optimization gains closely track the measured spectral concentration $\chi_b$. CNN layers exhibit nearly isotropic noise spectra ($\chi_b \approx 1$) and show negligible final performance difference between Muon and SGDM. Conversely, LLM attention (\texttt{QKV}) and MLP expansion blocks exhibit strong spectral concentration ($\chi_b \gg 1$), where Muon substantially accelerates pretraining and reduces final validation loss on the 1.34B model from $6.22$ to $3.22$.
\end{enumerate}

\subsection{Comparison with the Preliminary Conference Version}

A preliminary version of this work~\citep{tao2026when} appeared at the 40th Annual Conference on Neural Information Processing Systems (NeurIPS 2026), which studied when and why SignSGD outperforms SGD in vector optimization. This journal article elevates the investigation to matrix optimization to answer when and why Muon outperforms SGD, advancing the conference version in three main directions.
\begin{itemize}
    \item \textbf{Establishing when and why Muon outperforms SGD.} Extending the vector separation of~\citet{tao2026when} to the matrix domain is nontrivial, as a direct reduction from the vector SGD lower bound only yields a stochastic coefficient of $\textcolor{blue}{m\norm{\vSigma}_F^2}$, which leaves no separation from Muon under isotropic noise ($\chi(\vSigma)=1$). By developing a direct matrix-domain hard instance, we establish a sharp stochastic complexity of $\textcolor{red}{m^2\norm{\vSigma}_F^2}$ for SGD and SGDM, proving that Muon strictly outperforms both methods by a factor of $m\chi(\vSigma)\in[m,m^2]$ across all noise spectra.
    \item \textbf{Incorporating momentum into SGD, SignSGD, and Muon.} While the conference version analyzed only momentum-free SignSGD, Muon, and vanilla SGD, practical deep learning optimizers rely on momentum. We establish matching lower bounds for momentum-based SignSGD and Muon via the 3D rotating-gradient construction, and extend our SGD lower bound from vanilla SGD to SGDM.
    \item \textbf{Large-scale LLM experiments and matrix-level spectral diagnostics.} We expand our empirical study from $130$M-scale models and vector noise measurements to pretraining a $1.34$B LLaMA model and probing $7$B-scale checkpoints of Pythia-6.9B~\citep{biderman2023pythia}, paired with matrix-covariance spectral concentration measurements ($\chi_b$) across architectures, layer depths, module types, and training stages.
\end{itemize}

\paragraph{Organization.}
\Cref{sec:related-work} reviews related work, and \cref{sec:preliminaries} introduces the notation, assumptions, and algorithmic setup. \Cref{sec:matrix-optimizers} establishes matching upper and lower complexity bounds for \MuonRef{} via SignSGD with momentum. \Cref{sec:compare} proves the sharp lower bounds for \SGDRef{} and \SGDMRef{} and characterizes the resulting complexity separation. \Cref{sec:experiments} presents our empirical pretraining results and spectral noise analyses from ResNet-20 up to the 7B scale, and \cref{sec:conclusion} concludes the paper. Complete proofs and experimental details are provided in the appendices.

\section{Related Work}\label{sec:related-work}

\paragraph{Sign-based vector optimizers.}
Updating parameters using only gradient signs dates back at least to RProp~\citep{riedmiller1993rprop}, while its modern stochastic formulation was introduced by~\citet{bernstein2018signsgd}, who analyzed SignSGD and its momentum variant Signum for smooth non-convex optimization. Because SignSGD replaces the stochastic gradient with its coordinate-wise sign, its natural stationarity measure is the $\ell_1$-norm rather than the $\ell_2$-norm. This geometry is also closely tied to communication efficiency, as one-bit updates and majority voting make sign-based methods attractive in distributed learning~\citep{bernstein2019signsgd,jin2020stochastic,safaryan2021stochastic,NeurIPS:2024:Jiang}.

When smoothness is absent, deterministic sign updates face an inherent limitation. Even on simple convex non-smooth objectives, the discontinuity of the sign operator can repeatedly select unfavorable subgradient directions and prevent convergence~\citep{karimireddy2019error,xiao2023stochastic,yu2026finite}. Prior work addresses this pathology either through error feedback that accumulates compression residuals~\citep{karimireddy2019error} or by introducing unbiased stochastic sign conversions~\citep{jin2020stochastic,safaryan2021stochastic,yu2026stosignsgd}. In the smooth stochastic regime considered in this paper, another line of research sharpens the convergence guarantees of sign-based methods via momentum analyses~\citep{sun2023momentum,jiang2025improved}, variance reduction~\citep{chzhen2023signsvrg,NeurIPS:2024:Jiang}, and heavy-tailed noise models~\citep{kornilov2025signheavytail,yu2026signheavytails}. Vector sign methods are also closely linked to adaptive algorithms. Several studies interpret Adam~\citep{kingma15adam} through signed or normalized updates~\citep{balles2018dissecting,crawshaw2022robustness,kunstner2023noise,peng2025simple}, while Lion~\citep{chen2023symbolic} has emerged as a prominent momentum-sign optimizer with growing theoretical support~\citep{chenICLR2024lion,dong2024convergence,sfyraki2025lions,jiang2026lion,yu2026signheavytails}. In particular, \citet{yu2026signheavytails} show that SignSGD- and Lion-type methods can provably outperform AdamW under the heavy-tailed gradient noise commonly observed in language model training~\citep{zhang2020whyheavytail,yu2026openproblem}.

\paragraph{Matrix-sign optimizers.}
Matrix-sign optimizers extend the sign-based principle from coordinate-wise vector updates to structured matrix updates. The representative method is Muon~\citep{jordan2024muon}, which moves along the orthogonalized matrix sign direction $\msign{\Gb} = \U_r\V_r^\top$ for a matrix gradient $\Gb = \U_r\SB_r\V_r^\top$ of rank $r$, approximated in practice via Newton--Schulz iterations~\citep{kovarik1970some,bjorck1971iterative}. Subsequent variants incorporating Nesterov momentum, learning-rate alignment, and system-level refinements have demonstrated strong empirical gains in large-scale language model pretraining~\citep{liu2025muon,zeng2025glm,team2025kimi,wen2025fantastic,semenov2025benchmarking,deepseekai2026deepseekv4}, indicating that matrix-level spectral normalization offers advantages beyond coordinate-wise sign updates.

Theoretical analyses of Muon and related matrix optimizers are rapidly expanding along several directions. One line of work establishes convergence guarantees and norm-constrained interpretations under operator and nuclear norm geometry, momentum, and variance reduction~\citep{pethick2025training,bernstein2025modular,li2025note,shen2025convergence,chang2025convergence,chen2025muon,qian2025muon,kim2026convergence,shulgin2026beyond}. A second line studies spectral preconditioning, connections to Adam-type adaptivity, and structural extensions such as memory-efficient or second-order variants~\citep{si2025adamuon,sfyraki2025lions,huang2025limuon,li2025normuon,pan2025unbiased,zhang2025adagrad,crawshaw2025exploration,ma2026preconditioning,du2026newton,yang2026memory,bai2026dram}. A third line investigates implicit bias, generalization behavior, and domain-specific mechanisms behind Muon's empirical performance~\citep{tveit2025muon,page2025muonall,mehta2025muon,vasudeva2025muon,vasudeva2025the,frans2025really,wang2025muon,zhang2025provable,su2025isotropic}.

\paragraph{Lower complexity bounds.}
Complexity lower bounds provide the standard benchmark for determining whether an algorithmic speedup reflects a fundamental advantage or merely an artifact of upper-bound analysis. The oracle-complexity framework originates from~\citet{nemirovskij1983problem} and was extensively developed for stochastic convex optimization~\citep{agarwal2012information} to characterize how dimension, noise, and geometry limit first-order methods.

In smooth non-convex optimization, the classical goal is finding an $\epsilon$-stationary point measured by $\norm{\nabla f(\x)}_2$. In the deterministic setting, \citet{carmon2020lower,carmon2021lower} established sharp lower bounds in high dimensions, while~\citet{chewi2023complexity} characterized the query complexity even in one dimension across deterministic and randomized oracles. In the stochastic setting, \citet{ghadimi2013stochastic} proved the standard $\bigO{\epsilon^{-4}}$ upper bound for SGD under unbiased bounded-variance gradients, and subsequent works established its tightness. Specifically, \citet{drori2020complexity} proved worst-case lower bounds for SGD itself, and~\citet{arjevani2023lower} showed that any stochastic first-order method requires $\Omega(\epsilon^{-4})$ queries under bounded variance (and $\Omega(\epsilon^{-3})$ under mean-squared smoothness). Consequently, under standard Euclidean geometry, SGD is minimax optimal and cannot be improved in the worst case without modifying the geometry, oracle model, or stationarity criterion.

Recent work has therefore investigated whether adaptive or sign-based optimizers can provably improve over SGD under refined structural assumptions. \citet{jiang2024convergence} and~\citet{tao2026when} demonstrate a dimension-factor separation between AdaGrad or SignSGD and vanilla SGD under coordinate-wise smoothness, coordinate-wise noise bounds, and $\ell_1$-stationarity, whereas~\citet{crawshaw2025complexity} establish lower bounds for adaptive methods under $(L_0,L_1)$-smoothness. For matrix-sign methods, concurrent studies derive matching upper and lower bounds for Muon-type algorithms under scalar nuclear-norm moment assumptions~\citep{hubler2026free,zhang2026scale}. Complementing these scalar-noise analyses, our work studies a matrix-valued noise bound that preserves the covariance spectrum, pairing tight Muon bounds with sharp lower bounds for both SGD and SGDM on the same problem class.

\section{Preliminaries}
\label{sec:preliminaries}

This section introduces the notation, smoothness and noise assumptions, and algorithmic framework used throughout our analysis. Although our primary focus is matrix optimization, key lower-bound constructions are first developed in the vector setting and subsequently lifted to matrices. Specifically, the coordinate-wise sign under the $\ell_\infty/\ell_1$ norm pair serves as the vector counterpart of the matrix sign under the operator/nuclear norm pair.

\paragraph{General notation.}
For any positive integer $n$, let $[n]:=\{1,2,\ldots,n\}$. Throughout the paper, $T$ denotes the total number of stochastic-gradient queries, and $\epsilon>0$ denotes the target stationarity accuracy. All algorithms use a fixed step size $\eta>0$ and momentum parameter $\beta\in[0,1)$, which may depend on the horizon $T$. At each step $t\in[T]$, the algorithm queries a stochastic gradient at the current iterate, and we let $\{\F_t\}_{t\ge0}$ denote the natural filtration generated by the algorithm and oracle history up to time $t$ (including any internal oracle state revealed in previous queries), so that the $t$-th query point is $\F_{t-1}$-measurable and all conditional expectations are taken with respect to $\F_{t-1}$.

\subsection{Vector Optimization Geometry}
\label{sec:vector-assumptions}

We first consider vector objectives $\min_{\x\in\R^d} f(\x)$ with stochastic gradient $\g_t\in\R^d$ queried at $\x_t$. For vectors, $\abs{\x}$ denotes the element-wise absolute value, and $\sign{\x}$ is applied coordinate-wise with $\sign{0}=0$. For $\x\in\R^d$, $\Diag{\x}$ denotes the diagonal matrix with diagonal $\x$, and for $\Ab\in\R^{d\times d}$, $\Diagvec{\Ab}:=(A_{11},\ldots,A_{dd})^\top$. Under the sign geometry, the natural stationarity measure is $\norm{\nabla f(\x)}_1$~\citep{bernstein2018signsgd,sun2023momentum,jiang2025improved,yu2026signheavytails}. Below, we state the assumptions for the vector setting.

\AssumptionSeries{a}

\begin{assumption}[Lower-bounded objective]
\label{ass:bounded from below}
The differentiable function $f:\R^d\to\R$ is bounded from below, and for the prescribed initial point $\x_1$, a parameter $\Delta\ge0$ satisfies
$$
\inf_{\x\in\R^d}f(\x)>-\infty, \qquad f(\x_1)-\inf_{\x\in\R^d}f(\x)\le\Delta.
$$
\end{assumption}

\begin{assumption}[Coordinate-wise noise]
\label{ass:variance}
There exists a prescribed vector $\vsigma=(\sigma_1,\ldots,\sigma_d)\in\R_+^d$ such that, for every $t\ge 1$ and every $i\in[d]$,
$$
\expect{\g_t\mid\F_{t-1}}
=
\nabla f(\x_t), \qquad \expect{
    (g_{t,i}-\nabla_i f(\x_t))^2
    \mid\F_{t-1}
}
\le\sigma_i^2.
$$
\end{assumption}

\Cref{ass:bounded from below,ass:variance} are standard in the analysis of sign-based optimizers~\citep{bernstein2018signsgd,crawshaw2022robustness,jiang2024convergence}. In particular, the stochastic oracle is conditionally unbiased, while the noise variance is allowed to vary across coordinates.

\begin{assumption}[$\ell_\infty$-smoothness]
\label{ass:infty smoothness}
There exists $\Linf\ge0$ such that, for all $\x,\x^\prime\in\R^d$,
$$
\norm{\nabla f(\x^\prime)-\nabla f(\x)}_1
\le
\Linf\norm{\x^\prime-\x}_\infty.
$$
\end{assumption}

\Cref{ass:infty smoothness} is the natural smoothness condition paired with the $\ell_\infty/\ell_1$ geometry of sign updates~\citep{balles2020geometry,jiang2025improved,tao2026when}.

\subsection{Matrix Optimization Geometry}
\label{sec:matrix-assumptions}

We next consider differentiable matrix objectives $\min_{\WB\in\R^{m\times n}} F(\WB)$ with $m\le n$ and stochastic gradient $\Gb_t\in\R^{m\times n}$ queried at $\WB_t$. For matrices, $\norm{\cdot}_{\op}$, $\norm{\cdot}_*$, and $\norm{\cdot}_F$ denote the operator (spectral), nuclear (trace), and Frobenius norms, respectively. We denote the matrix trace by $\tr(\cdot)$, and define the matrix inner product as $\inner{\XB}{\YB} := \tr(\XB^\top\YB)$. We write $\SBB_+^m$ for the cone of $m\times m$ positive semidefinite matrices and $\preceq$ for the Loewner partial order. For a nonzero matrix $\XB\in\R^{m\times n}$ of rank $r:=\rank(\XB)$, let $\XB = \U_r\SB_r\V_r^\top$ be a reduced singular value decomposition with $\U_r\in\R^{m\times r}$, $\SB_r\in\R^{r\times r}$, and $\V_r\in\R^{n\times r}$, and define its compact matrix sign by
$$
\msign{\XB} := \U_r\V_r^\top,
$$
with $\msign{\vzero}:=\vzero$. Thus, $\msign{\XB}$ plays the same geometric role under the operator/nuclear norm pair as $\sign{\x}$ under the $\ell_\infty/\ell_1$ norm pair~\citep{pethick2025training,bernstein2025modular}. Throughout our theoretical analysis, we study Muon updates using the exact $\msign{\cdot}$ operator, and measure stationarity via the nuclear norm $\norm{\nabla F(\WB)}_*$~\citep{shen2025convergence,kovalev2025sgd,yu2026signheavytails}.

\AssumptionSeries{b}

\begin{assumption}[Lower-bounded matrix objective]
\label{ass:Matrix-bounded-below}
The differentiable function $F:\R^{m\times n}\to\R$ is bounded from below, and for the prescribed initial matrix $\WB_1$, the parameter $\Delta\ge0$ satisfies
$$
\inf_{\WB\in\R^{m\times n}}F(\WB)>-\infty, \qquad F(\WB_1)-\inf_{\WB\in\R^{m\times n}}F(\WB)
\le\Delta.
$$
\end{assumption}

\begin{assumption}[Matrix noise]
\label{ass:Matrix-variance}
Let $\Nb_t := \Gb_t-\nabla F(\WB_t)$ denote the stochastic-gradient noise. The oracle is conditionally unbiased and there exists a prescribed matrix $\vSigma\in\SBB_+^m$ such that, for every $t\ge 1$,
$$
\expect{\Gb_t\mid\F_{t-1}}
=
\nabla F(\WB_t), \qquad 
\expect{
    \Nb_t\Nb_t^\top
    \mid\F_{t-1}
}
\preceq
\vSigma^2.
$$
\end{assumption}

\Cref{ass:Matrix-variance} extends coordinate-wise variance bounds to the matrix domain by bounding the noise covariance via a positive semidefinite matrix $\vSigma^2$~\citep{xie2025tale,an2025asgo,pan2025unbiased,yu2026signheavytails,tao2026when}; taking the trace recovers the standard finite-variance bound $\E\left[\norm{\Nb_t}_F^2\mid\allowbreak\F_{t-1}\right]\le\norm{\vSigma}_F^2$.

\begin{assumption}[Operator-norm smoothness]
\label{ass:operator norm smooth}
There exists $L_*\ge0$ such that, for all $\WB,\WB'\in\R^{m\times n}$,
$$
\norm{
\nabla F(\WB')-\nabla F(\WB)
}_*
\le
L_*
\norm{\WB'-\WB}_{\op}.
$$
\end{assumption}

\Cref{ass:operator norm smooth} is the matrix analogue of \cref{ass:infty smoothness} and is widely adopted in the convergence analysis of Muon and spectral methods~\citep{pethick2025training,bernstein2025modular,li2025note,shen2025convergence,chang2025convergence,chen2025muon,qian2025muon,kovalev2025sgd,kim2026convergence,shulgin2026beyond,yu2026signheavytails}.

\subsection{Algorithms and Evaluation Criterion}
\label{sec:alg-conventions}

We summarize the matrix optimizers studied in this paper in \cref{alg:matrix-optimizers}. Following standard practical implementations and theoretical analyses~\citep{NEURIPS2019PYTORCH,jordan2024muon,shen2025convergence,yu2026signheavytails}, the momentum buffer is initialized with the first stochastic gradient $\MB_1=\Gb_1$ (equivalently setting $\MB_0=\Gb_1$) and updated via an exponential moving average (EMA) with parameter $\beta\in[0,1)$. \MuonRef{} updates the weight matrix along the matrix-sign direction $\WB_{t+1}=\WB_t-\eta\msign{\MB_t}$, whereas \SGDMRef{} performs the entrywise Euclidean update $\WB_{t+1}=\WB_t-\eta\MB_t$ (and reduces to vanilla \SGDRef{} when $\beta=0$)\footnote{When applied to matrix parameters $\WB\in\R^{m\times n}$, SGD and SGDM perform entrywise Euclidean updates $\WB_{t+1}=\WB_t-\eta\Gb_t$ and $\WB_{t+1}=\WB_t-\eta\MB_t$, which are equivalent to their vector counterparts on the vectorized matrix.}. In the vector setting (\cref{sec:vector-assumptions}), SignSGD with momentum follows the exact vector analogue of \MuonRef{} with $\m_1=\g_1$:
\begin{equation}\label{eq:ssgd-update}
\m_t=\beta\m_{t-1}+(1-\beta)\g_t\quad(t\ge2),
\qquad
\x_{t+1}=\x_t-\eta\sign{\m_t}.
\end{equation}

\begin{algorithm}[t]
\caption{Matrix Optimization via Muon and SGDM}
\label{alg:matrix-optimizers}
\begin{algorithmic}[1]
\STATE \textbf{Input:} Initial matrix $\WB_1\in\R^{m\times n}$, step size $\eta>0$, momentum parameter $\beta\in[0,1)$, horizon $T\ge1$.
\FOR{$t = 1, \ldots, T$}
    \STATE Query stochastic gradient $\Gb_t$ at $\WB_t$.
    \STATE Update momentum buffer: $\MB_t = \beta\MB_{t-1} + (1-\beta)\Gb_t$ \hfill $\triangleright\ \MB_1 = \Gb_1$
    \STATE Update parameters: $\WB_{t+1} = \begin{cases} \phantomsection\label{alg:muon-step}\WB_t - \eta\msign{\MB_t}, & \text{(\textbf{Muon})} \\[2pt] \phantomsection\label{alg:sgdm-step}\WB_t - \eta\MB_t. & \text{(\textbf{SGDM}; \textbf{SGD} if $\beta=0$)} \end{cases}$
\ENDFOR
\end{algorithmic}
\end{algorithm}

To evaluate optimization performance over $T$ queries, we measure the expected minimum gradient norm along the queried trajectory. For a vector algorithm $\mathrm{Alg}$ (such as $\mathrm{Sign}$ for SignSGD) on $f:\R^d\to\R$, and for a matrix algorithm $\mathrm{Alg}\in\{\mathrm{Muon},\mathrm{SGD},\mathrm{SGDM}\}$ on $F:\R^{m\times n}\to\R$, we define
\begin{equation}\label{eq:evaluation-criteria}
\Rt_T^{\mathrm{Alg}}(f) := \expect{\min_{t\in[T]}\norm{\nabla f(\x_t)}_1},
\qquad
\Rt_T^{\mathrm{Alg}}(F) := \expect{\min_{t\in[T]}\norm{\nabla F(\WB_t)}_*}.
\end{equation}
The worst-case iteration complexity to reach $\epsilon$-stationarity is the smallest horizon $T\ge 1$ for which some choice of fixed hyperparameters (with $\eta>0$ and $\beta\in[0,1)$ allowed to depend on $T$, and $\beta=0$ for vanilla \SGDRef{}) achieves $\sup_{(F,\O)}\Rt_T^{\mathrm{Alg}}(F)\le\epsilon$ over all admissible objectives $F$ and stochastic oracles $\O$. Thus an upper complexity bound requires exhibiting a valid horizon-dependent hyperparameter schedule, whereas a lower complexity bound holds uniformly over all fixed hyperparameters.

\section{Tight Upper and Lower Bounds for Muon}\label{sec:matrix-optimizers}

In this section, we establish matching upper and lower complexity bounds for \MuonRef{} under operator-norm smoothness (\cref{ass:operator norm smooth}) and matrix-valued stochastic noise (\cref{ass:Matrix-variance}). We begin in \cref{sec:muon-upper} with a spectrum-aware convergence upper bound for \MuonRef{}. To prove a matching lower bound, we first analyze its vector counterpart, SignSGD with momentum (\cref{sec:signsgd-lower}), where we construct a three-dimensional rotating-gradient hard instance that overcomes momentum averaging at fixed dimension, and we then lift this construction to \MuonRef{} in \cref{sec:muon-lower}. Finally, \cref{sec:muon-tight-characterization} combines our upper and lower bounds to establish the tight complexity characterization of \MuonRef{} and compares it with existing results.

\subsection{Upper Bound for Muon}\label{sec:muon-upper}

We first establish the convergence rate of \MuonRef{} (\cref{alg:matrix-optimizers}). By directly controlling the expected nuclear norm of the momentum noise error under \cref{ass:Matrix-variance}, we obtain an upper bound that depends on the noise spectrum through $\norm{\vSigma}_*$ without any explicit matrix dimension factor.

\begin{theorem}[Muon upper bound]
\label{thm:muon_upper_infty}
Suppose \cref{ass:Matrix-bounded-below,ass:Matrix-variance,ass:operator norm smooth}
hold with $L_*,\Delta>0$. For all sufficiently large integers $T$, by setting
$1-\beta = \min\left\{1, \frac{\sqrt{L_*\Delta}}{\norm{\vSigma}_*\sqrt{T}}\right\}$ (with $\beta=0$ when $\vSigma=\vzero$) and $\eta = \sqrt{\frac{\Delta(1-\beta)}{L_*T}}$ as in \eqref{eq:hyp_muon}, the \MuonRef{} iterates in \cref{alg:matrix-optimizers} satisfy
\begin{equation}\label{eq:matrix-muon-upper-rate}
    \frac1T\SumT\E\norm{\nabla F(\WB_t)}_*
    =\bigO{
        \sqrt{\frac{L_*\Delta}{T}}
        +\brac{\frac{\norm{\vSigma}_*^2{L_*\Delta}}{T}}^{1/4}
    }.
\end{equation}
\end{theorem}

The same bound holds for $\Rt_T^{\mathrm{Muon}}(F)$, and the complete proof is deferred to \cref{sec:proof-muon-upperbound}. Following the non-Euclidean matrix martingale concentration technique of~\citet{yu2026signheavytails}, the proof bounds the expected nuclear norm of the filtered noise sum directly by $\norm{\vSigma}_*$, avoiding Frobenius-norm relaxations that would otherwise incur an extra $\sqrt{m}$ factor. A systematic comparison between \cref{thm:muon_upper_infty} and existing convergence guarantees for \MuonRef{} is deferred to \cref{sec:muon-tight-characterization}.

\subsection{Lower Bound for SignSGD}\label{sec:signsgd-lower}

To establish a matching lower bound for \MuonRef{}, we first analyze its vector counterpart, SignSGD with momentum~\citep{bernstein2018signsgd} as defined in \eqref{eq:ssgd-update}, and prove a lower bound under $\ell_\infty$-smoothness and coordinate-wise noise.

\begin{theorem}[SignSGD lower bound]\label{thm:ssgd_lower_infty}
Fix $d\ge3$, $\Linf,\Delta>0$, $\x_1\in\R^d$, and
$\vsigma\in\R_+^d$ with at least two positive entries.
For all sufficiently large integers $T$ satisfying \eqref{slb-eq-time} in \cref{sec:proof-ssgd-lowerbound} and every fixed $\eta>0$ and $\beta\in[0,1)$,
there exist an objective
and an oracle satisfying
\cref{ass:bounded from below,ass:variance,ass:infty smoothness}, such that the iterates generated by
\eqref{eq:ssgd-update} satisfy
\begin{equation}\label{eq:lower_signsgd}
    \Rt_T^{\mathrm{Sign}}(f)
    =\expect{\min_{t\in[T]}\norm{\nabla f(\x_t)}_1}
    =\Omega\brac{
        \sqrt{\frac{\Linf\Delta}{T}}+
        \brac{\frac{\norm{\vsigma}_1^2{\Linf\Delta}}{T}}^{1/4}}.
\end{equation}
\end{theorem}

\begin{remark}[Dimension-independent lower bound]\label{rem:sign-lower-originality}
\Cref{thm:ssgd_lower_infty} provides the first matching lower bound for SignSGD with momentum under $\ell_\infty$-smoothness (\cref{ass:infty smoothness}), coordinate-wise noise (\cref{ass:variance}), and $\ell_1$-stationarity at fixed dimension. In contrast to the classical stochastic lower-bound construction of~\citet{arjevani2023lower}, which requires the dimension to grow as the target accuracy improves, our bound holds for any fixed $d\ge3$ \emph{without requiring the dimension to grow with the iteration budget}.
\end{remark}

\begin{proof}[Proof sketch of \cref{thm:ssgd_lower_infty}]
We normalize $\Linf=\Delta=1$, write $\alpha:=1-\beta$, and let
$\zeta=\delta(\nicefrac{\norm{\vsigma}_1^2}{T})^{1/4}$ denote the target stochastic
lower-bound scale. Extreme step sizes are handled by simple deterministic
constructions, which also yield the $\Omega(T^{-1/2})$ term. It therefore suffices to
consider the intermediate step-size regime.

The core challenge is overcoming the variance-reduction effect of EMA momentum. In the weak-momentum regime where $\norm{\vsigma}_1\sqrt{\alpha}\ge 8\zeta$, shifting the momentum state by $\Theta(\zeta)$ in a single step requires stochastic perturbations of magnitude $\Theta(\nicefrac{\zeta}{\alpha})$ and variance $\Theta(\nicefrac{\zeta^2}{\alpha})$, which remain within the noise budget $\vsigma$. A one-dimensional two-state construction therefore keeps the iterates in a region with $\norm{\nabla f(\x_t)}_1 = \Theta(\zeta)$ with constant probability.

In the complementary strong-momentum regime, $\norm{\vsigma}_1\sqrt{\alpha}<8\zeta$, the sufficient condition $\norm{\vsigma}_1\sqrt{\alpha}\ge 8\zeta$ for the two-state construction is no longer guaranteed. Instead, we design a three-dimensional rotating-gradient hard instance (\cref{fig:sign_rotating}). Along a longitudinal axis $\q_0$, the true gradient is prescribed as
\begin{equation}\label{eq:grad_along_axis}
\nabla f(z\q_0)
=
b\q_0
+
\zeta\sqbrac{
\cos(kz)\q_1+\sin(kz)\q_2
},
\qquad k\asymp\zeta^{-1},
\end{equation}
where $\q_1,\q_2\perp\q_0$. As the iterate advances along $\q_0$, the transverse gradient component $\v_t=\zeta[\cos(kz_t)\q_1+\sin(kz_t)\q_2]$ maintains a constant norm $\zeta$ while rotating in $\spann\{\q_1,\q_2\}$.

Let $\w_t$ denote the EMA of $\v_t$. At the initial query ($t=1$), the stochastic oracle returns $\g_1=b\q_0$ with probability $1/2$ to cancel the initial transverse gradient $\zeta\q_1$, yielding $\m_1=b\q_0$ and $\w_1=\vzero$; for $t\ge2$, the oracle returns exact gradients so that $\w_t=\beta\w_{t-1}+\alpha\v_t$. Along the candidate trajectory $z_{t+1}=z_t-\sqrt{3}\eta$, consecutive transverse gradients differ by a fixed planar rotation $\mathbf R_\theta$ with angle $\theta=\sqrt{3}k\eta\in(0,1]$ and $\mathbf R_\theta\v_{j+1}=\v_j$. Because the step-size condition of this regime forces $\beta>\nicefrac12$, applying $\mathbf R_\theta$ to the EMA recursion induces destructive interference between successive steps:
$$
\w_t-\beta\mathbf R_\theta\w_t
=\alpha\brac{\v_t-\beta^{t-1}\v_1},
$$
\begin{equation}\label{eq:rotation-cancellation-sketch}
\begin{aligned}
\norm{\w_t-\beta\mathbf R_\theta\w_t}_2
&\le\alpha(\norm{\v_t}_2+\beta^{t-1}\norm{\v_1}_2)
\le2\alpha\zeta.
\end{aligned}
\end{equation}

\begin{figure}[tbp]
\centering
\resizebox{0.46\textwidth}{!}{
\begingroup
\def\JMLRFigureFont{\fontsize{9.6}{11.5}\selectfont}
\begin{tikzpicture}[>=stealth, scale=1.0, every node/.style={font=\JMLRFigureFont,inner sep=1pt}]
    \useasboundingbox (-3.5, -2.1) rectangle (3.5, 2.1);
    \coordinate (O) at (0.4, -0.45);
    \fill[blue!5!gray!8] (-3.1, 0.45) -- (1.9, 0.45) -- (2.9, -1.45) -- (-2.1, -1.45) -- cycle;
    \draw[gray!50, thick] (-3.1, 0.45) -- (1.9, 0.45) -- (2.9, -1.45) -- (-2.1, -1.45) -- cycle;
    \node[text=gray!80!black] at (1.0, -1.76) {$\operatorname{span}\{\mathbf{q}_{\textstyle 1}, \mathbf{q}_{\textstyle 2}\}$};
    \coordinate (Q1_dir) at (-2.1, 0.4);
    \coordinate (Q2_dir) at (-1.5, -0.65);
    \draw[->, thick, gray!70!black] (O) -- ++(Q1_dir);
    \node[black] at (-1.9, 0.12) {$\mathbf{q}_{\textstyle 1}$};
    \draw[->, thick, gray!70!black] (O) -- ++(Q2_dir);
    \node[black] at (-1.3, -1.27) {$\mathbf{q}_{\textstyle 2}$};
    \draw[purple!40, dashed, thick] plot[domain=0:360, samples=60]
        ({0.4 - 1.4*cos(\x) - 0.6*sin(\x)}, {-0.45 + 0.28*cos(\x) - 0.42*sin(\x)});
    \draw[teal!85!black, ->, thick] plot[domain=35:65, samples=20]
        ({0.4 - 1.4*cos(\x) - 0.6*sin(\x)}, {-0.45 + 0.28*cos(\x) - 0.42*sin(\x)});
    \coordinate (Vt) at ({0.4 - 1.4*cos(35) - 0.6*sin(35)}, {-0.45 + 0.28*cos(35) - 0.42*sin(35)});
    \coordinate (CompQ1) at ($(O) + 0.78*(Q1_dir)$);
    \coordinate (CompQ2) at ($(O) + 0.58*(Q2_dir)$);
    \draw[gray!60, dashed, thin] (Vt) -- (CompQ1);
    \draw[gray!60, dashed, thin] (Vt) -- (CompQ2);
    \draw[very thick, teal!85!black, ->] (O) -- (Vt);
    \node[text=teal!85!black] at (-0.65, -0.86) {$\mathbf{v}_{\textstyle t}$};
    \node[anchor=east,text=teal!85!black] at (-1.25, -0.52) {$\|\mathbf{v}_{\textstyle t}\|_{\textstyle 2} = \zeta$};
    \draw[gray!80!black, thin] ($(O) + 0.28*(Q1_dir)$) to[out=-50, in=120] ($(O) + 0.32*(Vt) - 0.32*(O)$);
    \node[text=gray!90!black] at (-0.10, -0.20) {$kz$};
    \coordinate (Wt) at (0.62, -0.4);
    \draw[thick, teal!85!black, ->] (O) -- (Wt);
    \node[anchor=west,text=teal!85!black] at (0.94, -0.48) {$\mathbf{w}_{\textstyle t} \approx \mathbf{0}$};
    \draw[dashed, gray!60] (0.4, -1.6) -- (O);
    \draw[->, thick, gray!70!black] (O) -- (0.4, 1.95);
    \node[anchor=west,black] at (0.58, 1.84) {$b\mathbf{q}_{\textstyle 0}$};
    \draw[line width=1.5pt, red!90!black, ->] (O) -- (0.4, 0.7);
    \node[anchor=east] at (0.17, 0.95) {$\operatorname{sign}(\mathbf{m}_{\textstyle t})$};
    \coordinate (Mt) at ($(0.4, 0.7) + (Wt) - (O)$);
    \draw[thick, blue!60!black, dashed, ->] (O) -- (Mt);
    \node[anchor=west,align=left,text=blue!70!black] at (0.96, 0.53) {$\mathbf{m}_{\textstyle t} =$\\$b\mathbf{q}_{\textstyle 0} + \mathbf{w}_{\textstyle t}$};
\end{tikzpicture}
\endgroup}
\caption{SignSGD: 3D rotating-gradient gadget. The transverse gradient $\v_t$ rotates rapidly in $\spann\{\q_1, \q_2\}$ with frequency $k \asymp \zeta^{-1}$, causing destructive interference in the transverse momentum ($\norm{\w_t}_2 \ll b$). The longitudinal bias $b\q_0$ dominates, forcing $\sign{\m_t} \propto \q_0$ while maintaining $\norm{\nabla f}_1 \ge \zeta$.}
\label{fig:sign_rotating}
\end{figure}
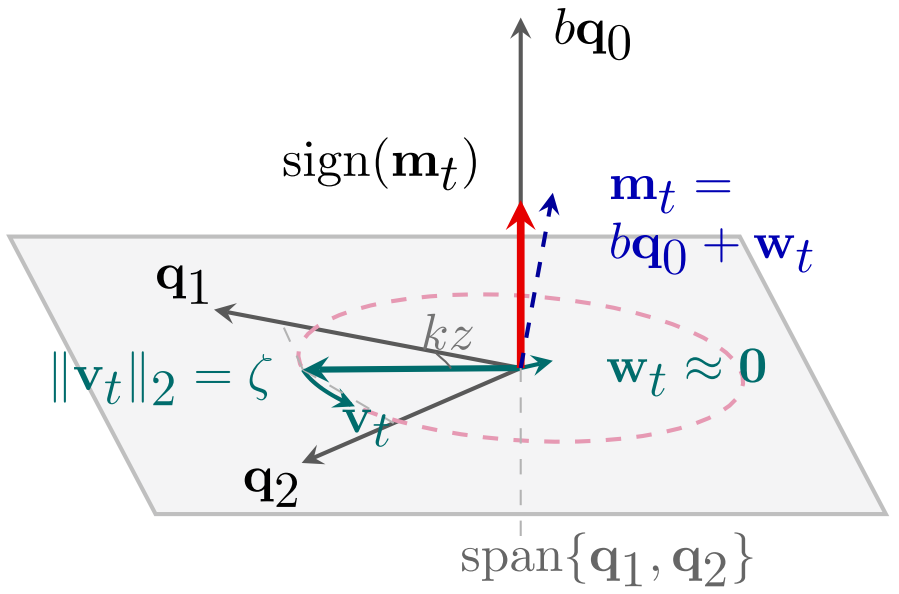

Since $\norm{\w_t-\beta\mathbf R_\theta\w_t}_2 \ge \frac{\theta}{2}\norm{\w_t}_2$, \eqref{eq:rotation-cancellation-sketch} implies $\norm{\w_t}_2\le\nicefrac{4\alpha\zeta}{\theta}=\nicefrac{4\alpha\zeta}{\sqrt3 k\eta}$. Choosing the longitudinal drift $b\asymp\nicefrac{\alpha\zeta}{k\eta}$ slightly larger than this transverse residual ensures that every coordinate of $\m_t = b\q_0 + \w_t$ remains positive, locking the update direction to $\sign{\m_t}=(1,1,1)^\top\propto\q_0$ while \eqref{eq:grad_along_axis} maintains $\norm{\nabla f(\x_t)}_1\ge\zeta$. Combining the weak- and strong-momentum constructions with the deterministic step-size obstructions covers all $(\eta,\beta)$ and completes the proof (\cref{sec:proof-ssgd-lowerbound}).
\end{proof}\hfill\mbox{}\par

\subsection{Lower Bound for Muon}\label{sec:muon-lower}

We now lift the vector hard instance from \cref{thm:ssgd_lower_infty} directly to the matrix domain, obtaining a matching lower bound for \MuonRef{} without additional dimension dependence.

\begin{theorem}[Muon lower bound]
\label{thm:muon_lower_infty}
Fix $3\leq m\leq n$, $L_*>0$, $\Delta>0$, an initial matrix
$\WB_1\in\R^{m\times n}$, and
$\vSigma\in\SBB_+^m$ with $\rank(\vSigma)\geq2$.
For all sufficiently large integers $T$ and every fixed $\eta>0$ and $\beta\in[0,1)$,
there exist a differentiable
objective $F:\R^{m\times n}\to\R$ and an adapted stochastic gradient
oracle satisfying
\cref{ass:Matrix-bounded-below,ass:Matrix-variance,ass:operator norm smooth}
with the prescribed parameters such that \MuonRef{} (\cref{alg:matrix-optimizers}) satisfies
\begin{equation}\label{eq:lower_muon}
    \Rt_T^{\mathrm{Muon}}(F)
    =\expect{\min_{1\leq t\leq T}
        \norm{\nabla F(\WB_t)}_*}
    =\Omega\brac{
        \sqrt{\frac{L_*\Delta}{T}}
        +\brac{\frac{\norm{\vSigma}_*^2L_*\Delta}{T}}^{1/4}
    }.
\end{equation}
\end{theorem}

\begin{proof}[Proof sketch of \cref{thm:muon_lower_infty}]
The proof proceeds via a \emph{direct spectral reduction from the vector SignSGD lower bound} (\cref{thm:ssgd_lower_infty}). The key geometric observation is that on diagonal matrices, the matrix sign reduces to the coordinate-wise sign:
$$
\msign{\Diag{\m}}
=
\Diag{\sign{\m}}.
$$
This allows us to embed a SignSGD trajectory into a \MuonRef{} trajectory by aligning the vector coordinates with orthogonal singular directions. Specifically, we diagonalize the prescribed noise matrix as $\vSigma=\QB^\top\Diag{\vsigma}\QB$ for orthogonal $\QB\in\R^{m\times m}$ and embed the vector hard instance $f:\R^m\to\R$ from \cref{thm:ssgd_lower_infty} via $F(\WB):=f(\A(\WB))$, where $\A(\WB):=\x_1+\Diagvec{\QB(\WB-\WB_1)\PB^\top}$ and $\PB=[\Ib_m,\vzero]\in\R^{m\times n}$. By the chain rule, the matrix gradient and momentum state take the form
$$
\nabla F(\WB) = \QB^\top\Diag{\nabla f(\A(\WB))}\PB,
\qquad
\MB_t=\QB^\top\Diag{\m_t}\PB,
$$
which implies $\msign{\MB_t} = \QB^\top\Diag{\sign{\m_t}}\PB$. Projecting the matrix iterates onto the diagonal coordinates via $\A$ therefore recovers the exact SignSGD recursion $\A(\WB_{t+1})=\A(\WB_t)-\eta\sign{\m_t}$, so the matrix and vector trajectories coincide pathwise.

Finally, orthogonal invariance of singular values ensures that the stationarity measure and noise scale are preserved:
$$
\norm{\nabla F(\WB_t)}_*
=
\norm{\nabla f(\x_t)}_1,
\qquad
\norm{\vSigma}_*=\norm{\vsigma}_1.
$$
The same embedding preserves the initial objective gap, transfers $\ell_\infty$-smoothness (\cref{ass:infty smoothness}) to operator-norm smoothness (\cref{ass:operator norm smooth}), and lifts the coordinate-wise vector noise bound (\cref{ass:variance}) to the PSD matrix noise bound (\cref{ass:Matrix-variance}); see \cref{sec:proof-muon-lowerbound} for the full proof. Consequently, the vector lower bound of \cref{thm:ssgd_lower_infty} transfers directly to \MuonRef{} without incurring any additional dimension factor.
\end{proof}

\subsection{Tight Complexity Characterization of Muon}\label{sec:muon-tight-characterization}

Combining the upper bound in \cref{thm:muon_upper_infty} with the matching lower bound in \cref{thm:muon_lower_infty} (and \cref{cor:matrix-muon-zero-noise} for the noiseless endpoint $\vSigma=\vzero$) directly yields a tight characterization of the worst-case iteration complexity of \MuonRef{}.

\begin{theorem}[Tight complexity of Muon]\label{thm:muon-tight-complexity}
Under \cref{ass:Matrix-bounded-below,ass:Matrix-variance,ass:operator norm smooth} with $3\le m\le n$, $L_*,\Delta>0$, and $\vSigma=\vzero$ or $\rank(\vSigma)\ge2$, for all sufficiently small $\epsilon>0$, the worst-case number of stochastic-gradient queries required by \MuonRef{} (\cref{alg:matrix-optimizers}) to achieve $\Rt_T^{\mathrm{Muon}}(F) = \expect{\min_{t\in[T]}\norm{\nabla F(\WB_t)}_*}\le\epsilon$ is
\begin{equation}\label{eq:muon-tight-ours-thm}
    \Theta\brac{\frac{L_*\Delta}{\epsilon^2}
        +\frac{\norm{\vSigma}_*^2 L_*\Delta}{\epsilon^4}}.
\end{equation}
\end{theorem}

\Cref{tab:muon-complexity-comparison} summarizes \cref{thm:muon-tight-complexity} alongside existing complexity bounds for \MuonRef{} under nuclear-norm stationarity (where $m\le n$), which we compare in detail below by dividing prior results into upper-bound analyses and tight characterizations.

\paragraph{Comparison with upper-bound analyses ($\mathcal{O}$).}
We first compare our upper bound (\cref{thm:muon_upper_infty}) with the three rows of upper-bound guarantees in \cref{tab:muon-complexity-comparison}.
\begin{itemize}
    \item \textbf{Frobenius smoothness and scalar variance~\citep{li2025note,chen2025muon}.} These analyses assume Euclidean smoothness $(F,F,L_F)$ and bounded scalar variance $\sigma^2=\norm{\vSigma}_F^2$. Because converting Frobenius gradient bounds to nuclear-norm stationarity via $\norm{\nabla F}_*\le\sqrt{m}\norm{\nabla F}_F$ incurs a $\sqrt{m}$ factor in the stationarity target (and $L_*\le mL_F$), their guarantees suffer explicit dimension penalties of $m$ in the deterministic term and $m^2\sigma^2$ in the stochastic term.

    \item \textbf{Operator-norm smoothness and scalar variance~\citep{pethick2025training,shen2025convergence}.} Adopting operator-norm smoothness $(*,\op,L_*)$ aligns the descent lemma with the polar update and removes the dimension factor from the deterministic term. However, under the scalar variance bound $\sigma^2=\norm{\vSigma}_F^2$, their noise analysis still relies on the Cauchy--Schwarz relaxation $\norm{\XB}_*\le\sqrt{m}\norm{\XB}_F$. Specifically, for any $\vSigma\ne\vzero$ and sufficiently small $\epsilon>0$ (see \cref{sec:proof-muon-upperbound}), \cref{thm:muon_upper_infty} and~\citet[Corollary~4.8]{shen2025convergence} compare as follows.
    \begin{formulacomparison}
    \begin{comparisonpanel}{comparisonblue}
        \textbf{Our upper bound}
        \begin{equation}\label{eq:muon-complexity-ours}
        \bigO{\frac{L_*\Delta}{\epsilon^2}
            +\frac{\textcolor{red}{\norm{\vSigma}_*^2}L_*\Delta}{\epsilon^4}}
        \end{equation}
    \end{comparisonpanel}\hfill
    \begin{comparisonpanel}{comparisongold}
        \textbf{\citet{shen2025convergence}}
        \begin{equation}\label{eq:muon-complexity-shen}
        \bigO{\frac{L_*\Delta}{\epsilon^2}
            +
            \frac{\textcolor{red}{m\norm{\vSigma}_F^2}L_*\Delta}{\epsilon^4}}
        \end{equation}
    \end{comparisonpanel}
    \end{formulacomparison}
    The ratio between the highlighted stochastic coefficients in \eqref{eq:muon-complexity-shen} and \eqref{eq:muon-complexity-ours} is precisely the spectral concentration $\chi(\vSigma) = \nicefrac{m\norm{\vSigma}_F^2}{\norm{\vSigma}_*^2} \in [1,m]$ defined in \eqref{eq:intro-spectral-concentration}. Thus, our upper bound is never worse than~\eqref{eq:muon-complexity-shen} and is strictly sharper by a factor of $\chi(\vSigma)$ whenever the noise is anisotropic ($\chi(\vSigma)>1$).

    \item \textbf{Operator-norm smoothness and matrix noise~\citep{yu2026signheavytails}.} By developing matrix martingale concentration under structured matrix noise (including heavy-tailed settings), \citet{yu2026signheavytails} obtain the upper bound $\bigO{\nicefrac{L_*\Delta}{\epsilon^2}+\nicefrac{\norm{\vSigma}_*^2 L_*\Delta}{\epsilon^4}}$ without explicit dimension factors. However, their work studies only upper bounds and leaves open whether this rate is minimax optimal for \MuonRef{}.
\end{itemize}

\begin{table}[tbp]
\centering
\caption{Comparison of complexity bounds for \MuonRef{} to achieve $\expect{\min_{t\in[T]}\norm{\nabla F(\WB_t)}_*}\le\epsilon$ (where $m\le n$). Here a smoothness triple $(p,q,L)$ denotes $\norm{\nabla F(\WB')-\nabla F(\WB)}_p \le L\norm{\WB'-\WB}_q$. Our smoothness condition $(*,\op,L_*)$ and matrix noise condition $\expect{\Nb_t\Nb_t^\top}\preceq\vSigma^2$ correspond to \cref{ass:operator norm smooth} and \cref{ass:Matrix-variance}, respectively.}
\label{tab:muon-complexity-comparison}
\small
\renewcommand{\arraystretch}{1.35}
\begin{tabular}{llll}
\toprule
\textbf{Reference} & \textbf{Smoothness} & \textbf{Noise} & \textbf{Complexity} \\
\midrule
\makecell[l]{\citet{li2025note}\\\citet{chen2025muon}} & $(F, F, L_F)$ & $\expect{\norm{\Nb_t}_F^2} \le \sigma^2$ & $\bigO{\dfrac{mL_F\Delta}{\epsilon^2} + \dfrac{m^2\sigma^2 L_F\Delta}{\epsilon^4}}$ \\[14pt]
\makecell[l]{\citet{pethick2025training}\\\citet{shen2025convergence}} & $(*, \op, L_*)$ & $\expect{\norm{\Nb_t}_F^2} \le \sigma^2$ & $\bigO{\dfrac{L_*\Delta}{\epsilon^2} + \dfrac{m\sigma^2 L_*\Delta}{\epsilon^4}}$ \\[14pt]
\citet{yu2026signheavytails} & $(*, \op, L_*)$ & $\expect{\Nb_t\Nb_t^\top} \preceq \vSigma^2$ & $\bigO{\dfrac{L_*\Delta}{\epsilon^2} + \dfrac{\norm{\vSigma}_*^2 L_*\Delta}{\epsilon^4}}$ \\[12pt]
\makecell[l]{\citet{hubler2026free}\\\citet{zhang2026scale}} & $(*, \op, L_*)$ & $\expect{\norm{\Nb_t}_*^2} \le \sigma_{\rm nuc}^2$ & $\Theta\brac{\dfrac{L_*\Delta}{\epsilon^2} + \dfrac{m\sigma_{\rm nuc}^2 L_*\Delta}{\epsilon^4}}$ \\[14pt]
\makecell[l]{\citet{tao2026when}\\($\beta=0$, large batch)} & $(*, \op, L_*)$ & $\expect{\Nb_t\Nb_t^\top} \preceq \vSigma^2$ & $\Theta\brac{\dfrac{L_*\Delta}{\epsilon^2} + \dfrac{\norm{\vSigma}_*^2 L_*\Delta}{\epsilon^4}}$ \\[14pt]
\midrule
\makecell[l]{\textbf{This work}\\(\cref{thm:muon-tight-complexity})} & $(*, \op, L_*)$ & $\expect{\Nb_t\Nb_t^\top} \preceq \vSigma^2$ & $\Theta\brac{\dfrac{L_*\Delta}{\epsilon^2} + \dfrac{\norm{\vSigma}_*^2 L_*\Delta}{\epsilon^4}}$ \\
\bottomrule
\end{tabular}
\end{table}

\paragraph{Comparison with tight characterizations ($\Theta$).}
We next compare \cref{thm:muon-tight-complexity} against existing matching upper and lower bounds in \cref{tab:muon-complexity-comparison}.
\begin{itemize}
    \item \textbf{Momentum-free matrix updates~\citep{tao2026when}.} The preliminary conference version established a matching lower bound only for memoryless updates ($\beta=0$), leaving open whether EMA momentum ($\beta>0$) can reduce the worst-case complexity of \MuonRef{}. \Cref{thm:muon-tight-complexity} resolves this question: our lower bound holds for every fixed $(\eta,\beta)\in(0,\infty)\times[0,1)$ and matches the horizon-tuned upper bound of \cref{thm:muon_upper_infty}, so the tight complexity remains $\Theta\brac{\nicefrac{L_*\Delta}{\epsilon^2}+\nicefrac{\norm{\vSigma}_*^2 L_*\Delta}{\epsilon^4}}$.

    \item \textbf{Scalar nuclear-norm noise bounds~\citep{hubler2026free,zhang2026scale}.} Concurrent works establish matching upper and lower bounds for momentum-based \MuonRef{} under the scalar nuclear-norm variance condition
    \begin{equation}\label{eq:scalar-nuclear-noise-cond}
        \expect{\norm{\Nb_t}_*^2\mid\F_{t-1}} \le \sigma_{\rm nuc}^2,
    \end{equation}
    which corresponds to $p=2$ in their $p$-th moment assumption. Contrasting \cref{thm:muon-tight-complexity} under $\expect{\Nb_t\Nb_t^\top\mid\F_{t-1}}\preceq\vSigma^2$ with the tight complexity of~\citet{hubler2026free} under \eqref{eq:scalar-nuclear-noise-cond} gives the comparison below.
    \begin{formulacomparison}
    \begin{comparisonpanel}{comparisonblue}
        \textbf{Our bound}
        \begin{equation}\label{eq:muon-tight-ours}
        \Theta\brac{\frac{L_*\Delta}{\epsilon^2}
            +\frac{\textcolor{red}{\norm{\vSigma}_*^2}L_*\Delta}{\epsilon^4}}
        \end{equation}
    \end{comparisonpanel}\hfill
    \begin{comparisonpanel}{comparisongold}
        \textbf{\citet{hubler2026free}}
        \begin{equation}\label{eq:muon-tight-scalar}
        \Theta\brac{\frac{L_*\Delta}{\epsilon^2}
            +\frac{\textcolor{red}{m\sigma_{\rm nuc}^2}L_*\Delta}{\epsilon^4}}
        \end{equation}
    \end{comparisonpanel}
    \end{formulacomparison}
    Because \eqref{eq:scalar-nuclear-noise-cond} compresses the matrix noise into a single scalar $\sigma_{\rm nuc}^2$, the complexity in \eqref{eq:muon-tight-scalar} retains an explicit dimension factor $m$. On the common subclass with time-independent covariance $\vSigma^2=\expect{\Nb_t\Nb_t^\top\mid\F_{t-1}}$ and $\sigma_{\rm nuc}^2:=\expect{\norm{\Nb_t}_*^2\mid\F_{t-1}}$, the inequalities $\norm{\vSigma}_F^2=\E\norm{\Nb_t}_F^2\le\sigma_{\rm nuc}^2$ and $\norm{\vSigma}_*^2\le m\norm{\vSigma}_F^2$ imply that the ratio between the highlighted stochastic coefficients in \eqref{eq:muon-tight-scalar} and \eqref{eq:muon-tight-ours} satisfies
    \begin{equation}\label{eq:scalar-matrix-comparison}
        \frac{m\sigma_{\rm nuc}^2}{\norm{\vSigma}_*^2}
        =
        \frac{\sigma_{\rm nuc}^2}{\norm{\vSigma}_F^2}
        \chi(\vSigma)
        \ge \chi(\vSigma)\ge1.
    \end{equation}
    Thus our spectrum-aware complexity \eqref{eq:muon-tight-ours} is never larger and is strictly smaller by a factor of at least $\chi(\vSigma)\in[1,m]$ whenever the noise is anisotropic ($\chi(\vSigma)>1$); in particular, when the noise matrix is rank one almost surely, $\norm{\Nb_t}_*=\norm{\Nb_t}_F$ and the improvement factor in \eqref{eq:scalar-matrix-comparison} is exactly $\chi(\vSigma) = \nicefrac{m\norm{\vSigma}_F^2}{\norm{\vSigma}_*^2}$. This does not contradict the lower bounds in~\citet{hubler2026free,zhang2026scale}, as their scalar noise class \eqref{eq:scalar-nuclear-noise-cond} is broader and must accommodate the worst-case covariance spectra compatible with $\sigma_{\rm nuc}^2$ (for which $\norm{\vSigma}_*^2$ can be as large as $m\sigma_{\rm nuc}^2$). Prescribing the matrix noise envelope $\vSigma$ refines the problem class and yields a sharper, spectrum-adaptive tight complexity. Moreover, whereas their lower bounds rely on high-dimensional constructions where $m$ grows as $\epsilon$ decreases, \cref{thm:muon_lower_infty} holds for any fixed $m \ge 3$.
\end{itemize}

In summary, closing the gap between upper and lower bounds in \cref{thm:muon-tight-complexity} precisely pins down the intrinsic complexity of \MuonRef{} and reveals how it adapts to both matrix geometry and the noise spectrum without explicit dimension penalties. Crucially, this tight characterization provides a rigorous foundation for our comparison with \SGDRef{} and \SGDMRef{} in \cref{sec:compare}, ensuring that the resulting complexity separation reflects a genuine algorithmic advantage rather than an artifact of loose bounds.

\section{Complexity Separation Between Muon and SGD}\label{sec:compare}

\Cref{sec:matrix-optimizers} established that \MuonRef{} achieves a tight complexity of $\Theta\big(\nicefrac{L_*\Delta}{\epsilon^2}+\nicefrac{\norm{\vSigma}_*^2 L_*\Delta}{\epsilon^4}\big)$ without explicit dimension factors. However, this result alone does not yet explain \MuonRef{}'s advantage over classical Euclidean optimizers, as it leaves open how \SGDRef{} and \SGDMRef{} perform when evaluated on the exact same matrix problem class (\cref{ass:Matrix-bounded-below,ass:Matrix-variance,ass:operator norm smooth}) under the same nuclear-norm stationarity criterion. In this section, we resolve this question in three steps.
\begin{enumerate}
    \item \textbf{Tight complexity of vanilla SGD (\cref{sec:sgd-lower-bound}).} We establish matching upper and lower bounds for memoryless \SGDRef{}, showing that Euclidean updates inevitably incur explicit dimension factors in both the deterministic and stochastic terms.
    \item \textbf{Extension to SGDM (\cref{sec:sgdm-results}).} Because \MuonRef{} incorporates momentum whereas vanilla \SGDRef{} does not, we extend our analysis to \SGDMRef{} and prove that adding momentum cannot eliminate these dimensional penalties.
    \item \textbf{When and why Muon wins (\cref{sec:theo_improve}).} Contrasting the tight complexities of \MuonRef{} and \SGDRef{}/\SGDMRef{} reveals a strict complexity separation and characterizes how the speedup depends on matrix dimension and noise anisotropy.
\end{enumerate}

Throughout this section, fix $2\le m\le n$, $L_*,\Delta>0$, the initial matrix $\WB_1\in\R^{m\times n}$, and the noise matrix $\vSigma\in\SBB_+^m$. Consistent with \cref{sec:alg-conventions}, we evaluate any algorithm $\mathrm{Alg}\in\{\mathrm{Muon},\mathrm{SGD},\mathrm{SGDM}\}$ over $T$ queries via $\Rt_T^{\mathrm{Alg}}(F) = \expect{\min_{t\in[T]}\norm{\nabla F(\WB_t)}_*}$. In each lower bound, the algorithm hyperparameters ($\eta$ for \SGDRef{}, or $(\eta,\beta)$ for \SGDMRef{}) are fixed prior to the run, after which a differentiable objective $F$ and an adapted stochastic-gradient oracle satisfying \cref{ass:Matrix-bounded-below,ass:Matrix-variance,ass:operator norm smooth} are constructed.

\subsection{Tight Complexity of Vanilla SGD}
\label{sec:sgd-lower-bound}

Under standard Euclidean smoothness and finite variance, vanilla \SGDRef{} ($\WB_{t+1}=\WB_t-\eta\Gb_t$) is minimax optimal. Under our matrix formulation, however, \cref{ass:operator norm smooth} constrains gradients in the nuclear norm while measuring steps in the operator norm. We first show that converting standard Euclidean descent to nuclear-norm stationarity yields an upper bound with explicit dimension factors (\cref{app:sgd-upper-cap}).

\begin{theorem}[SGD upper bound]\label{thm:matrix-sgd-upper}
Suppose \cref{ass:Matrix-bounded-below,ass:Matrix-variance,ass:operator norm smooth} hold with $L_*,\Delta>0$. For every integer $T\ge1$, by setting $\eta=\min\left\{\frac1{L_*},\sqrt{\frac{2\Delta}{L_*\norm{\vSigma}_F^2 T}}\right\}$ when $\vSigma\ne\vzero$ and $\eta=\frac1{L_*}$ when $\vSigma=\vzero$, the \SGDRef{} iterates in \cref{alg:matrix-optimizers} ($\beta=0$) satisfy
\begin{equation}\label{eq:matrix-sgd-upper-rate}
 \frac1T\sum_{t=1}^T\E\norm{\nabla F(\WB_t)}_*
 =\bigO{\sqrt{\frac{mL_*\Delta}{T}}+
 \left(\frac{m^2\norm{\vSigma}_F^2L_*\Delta}{T}\right)^{1/4}}.
\end{equation}
\end{theorem}

The same upper bound holds for $\Rt_T^{\mathrm{SGD}}(F)$. Next, we prove a matching lower bound showing that these dimension factors are not artifacts of the upper-bound analysis, but are intrinsic to Euclidean updates under operator/nuclear norm geometry.

\begin{theorem}[SGD lower bound]\label{thm:matrix-sgd-all-horizon}
Fix $2\le m\le n$, $L_*,\Delta>0$, an initial matrix
$\WB_1\in\R^{m\times n}$, and any prescribed $\vSigma\in\SBB_+^m$.
For all sufficiently large integers $T$ and every fixed step size $\eta>0$,
there exist a differentiable objective $F:\R^{m\times n}\to\R$
and an adapted stochastic-gradient oracle satisfying
\cref{ass:Matrix-bounded-below,ass:Matrix-variance,ass:operator norm smooth}
with the prescribed parameters such that the \SGDRef{} iterates in \cref{alg:matrix-optimizers} ($\beta=0$) satisfy
\begin{equation}\label{eq:matrix-sgd-all-horizon}
 \Rt_T^{\mathrm{SGD}}(F) = \expect{\min_{t\in[T]}\norm{\nabla F(\WB_t)}_*}
 =\Omega\brac{\sqrt{\frac{mL_*\Delta}{T}}+
 \left(\frac{m^2\norm{\vSigma}_F^2L_*\Delta}{T}\right)^{1/4}}.
\end{equation}
\end{theorem}

\begin{proof}[Proof sketch of \cref{thm:matrix-sgd-all-horizon}]
By rescaling and rotating into the eigenbasis of $\vSigma=\U\Diag{\sigma_1,\ldots,\sigma_m}\U^\top$, we work in the normalized setting $L_*=\Delta=1$ with diagonal noise (\cref{app:matrix-rescaling}).

We first dispose of the extreme and moderate step-size regimes: a diagonal Huber function rules out small $\eta$ (giving the deterministic floor $\Omega(\sqrt{m/T})$), a periodic objective rules out large $\eta$, and a simpler circular noise oracle covers moderate $\eta$, where stochastic perturbations alone prevent convergence.

In the hardest remaining step-size window, we construct a matrix hard instance in eigenbasis coordinates $\XB$ that traps the iterates in a region where $\norm{\nabla F(\WB_t)}_*\ge u$ for all $T$ queries with constant probability via three mechanisms.
\begin{enumerate}
    \item \textbf{Diagonal gradient dilution (origin of the $m$ factor).}
    We let the normalized objective $\bar F(\XB)$ depend on the average $w_1=\frac1{k_1}\sum_{i\in I_1}X_{ii}$ over $k_1=\Theta(m)$ diagonal entries. When $\abs{\partial \bar F/\partial w_1}=u$, each entry $i\in I_1$ receives a partial derivative of size $u/k_1$, so the nuclear norm still satisfies $\norm{\nabla \bar F(\XB)}_*\ge k_1(u/k_1)=u$. Under exact diagonal gradients, however, \SGDRef{} moves each entry by only $\eta u / k_1$, so $w_1$ advances at a diluted speed of $\Theta(\eta u / m)$ per step. (By contrast, on an isolated diagonal block whose momentum is a nonzero multiple of the identity, \MuonRef{} normalizes all $k_1$ nonzero singular values to $1$ and advances $w_1$ at full speed $\eta$.)

    \item \textbf{Sign-switching column pulses (origin of $\norm{\vSigma}_F^2$).}
    Because the initial gap is $\Delta=1$, a slope of size $u$ can be sustained only over an interval $\abs{w_1}\le B\asymp 1/u$. For $m\ge 4$, we prevent $w_1$ from escaping $[-B,B]$ by coupling its slope to the sign product $\zeta := \prod_{i=1}^m\sign{s_i}$ of the first column $s=\XB_{:,1}$, so the active-region objective has the form $\bar F(\XB)=u\,w_1\prod_{i=1}^m\sign{s_i}+H(s)$ (smoothly cut off outside this region for global admissibility). On each active step, the oracle returns zero on the control column with probability $1-p$. With probability $p_i$, it injects a pulse of size $\Theta(1/\eta)$ along entry $i$ that flips $\sign{s_i}$ and reverses the drift sign $\zeta$ of $w_1$; balancing the pulse magnitude against the slope of $H(s)$ keeps the oracle unbiased. In this eigenbasis, a pulse of size $\Theta(1/\eta)$ on entry $i$ fits within the variance budget $\sigma_i^2$ when $p_i \asymp \eta^2\sigma_i^2$, so the total sign-flip probability per step is
    \[
     p = \sum_{i=1}^m p_i \asymp \eta^2\norm{\vSigma}_F^2.
    \]

    \item \textbf{Persistent random walk and step-size cancellation (origin of $m^2\norm{\vSigma}_F^2/T$).}
    Because the drift sign flips with probability $p$ per step, it persists over correlation windows of length $\Theta(1/p)$. The cumulative sign sum therefore behaves as a persistent random walk whose typical maximal excursion over $T$ steps is $\bigO{\sqrt{T/p}}$. Multiplying this excursion by the diluted per-step drift $\Theta(\eta u/m)$ bounds the total displacement of $w_1$ with constant probability:
    \[
     \max_{t\le T}\abs{w_{1,t}}
     \lesssim \frac{\eta u}{m}\sqrt{\frac{T}{p}}
     \asymp \frac{u}{m}\sqrt{\frac{T}{\norm{\vSigma}_F^2}}.
    \]
    Here $\eta$ cancels out: a larger step size accelerates diagonal drift ($\propto \eta$) but raises the sign-flip frequency quadratically ($p \propto \eta^2$). Requiring $\max_{t\le T}\abs{w_{1,t}} \le B \asymp 1/u$ allows us to set $u \asymp (m^2\norm{\vSigma}_F^2 / T)^{1/4}$, which keeps $\norm{\nabla F(\WB_t)}_* \ge u$ for all $t\in[T]$ with constant probability.
\end{enumerate}
The complete proof for all $T\ge 1$ and $m\ge 2$ is given in \cref{app:sgd-lower-bound}.
\end{proof}

Combining \cref{thm:matrix-sgd-upper,thm:matrix-sgd-all-horizon} immediately yields the tight iteration complexity of vanilla \SGDRef{} (\cref{sec:complexity-conversion}).

\begin{theorem}[Tight complexity of SGD]\label{thm:sgd-tight-complexity}
Under \cref{ass:Matrix-bounded-below,ass:Matrix-variance,ass:operator norm smooth} with $2\le m\le n$ and $L_*,\Delta>0$, for all sufficiently small $\epsilon>0$, the worst-case number of stochastic-gradient queries required by \SGDRef{} ($\beta=0$ in \cref{alg:matrix-optimizers}) to achieve $\Rt_T^{\mathrm{SGD}}(F)\le\epsilon$ is
\begin{equation}\label{eq:sgd-tight-complexity}
 \Theta\brac{\frac{mL_*\Delta}{\epsilon^2}
 +\frac{m^2\norm{\vSigma}_F^2 L_*\Delta}{\epsilon^4}}.
\end{equation}
\end{theorem}

\subsection{Extension to SGD with Momentum}\label{sec:sgdm-results}

\Cref{thm:sgd-tight-complexity} establishes a clear separation between \MuonRef{} and vanilla \SGDRef{}. However, since \MuonRef{} incorporates momentum ($\beta > 0$), a comparison against memoryless \SGDRef{} alone cannot distinguish whether \MuonRef{}'s advantage stems from its matrix-sign update or simply from momentum smoothing. We therefore extend our analysis to \SGDMRef{} (\cref{alg:matrix-optimizers}) across all fixed hyperparameter pairs $(\eta,\beta)\in(0,\infty)\times[0,1)$.

\begin{theorem}[SGDM lower bound]\label{thm:direct-matrix-sgdm}
Fix $2\le m\le n$, $L_*,\Delta>0$, $\WB_1\in\R^{m\times n}$, and $\vSigma\in\SBB_+^m$. Set $v:=\norm{\vSigma}_F^2/(L_*\Delta)$, $\ell_m:=\log(e+m)$, and define
\begin{equation}
\begin{aligned}
 b_T:=\min\{1,\sqrt{m/T}\},\quad
 d_T:=\min\{1,(mv/T)^{1/4}\},\\
 s_T:=\min\{\ell_m^{-1/2},(mv/T)^{1/6},(m^2v/T)^{1/4}\},
\label{eq:direct-sgdm-branches}
\end{aligned}
\end{equation}
with $d_T=s_T=0$ when $\vSigma=\vzero$. There is a universal constant $c>0$ such that, for every integer $T\ge1$ and every fixed $(\eta,\beta)\in(0,\infty)\times[0,1)$, there exist a differentiable objective $F:\R^{m\times n}\to\R$ and an adapted stochastic-gradient oracle satisfying \cref{ass:Matrix-bounded-below,ass:Matrix-variance,ass:operator norm smooth} such that the \SGDMRef{} iterates in \cref{alg:matrix-optimizers} satisfy
\begin{equation}\label{eq:direct-sgdm-all-horizon}
 \Rt_T^{\mathrm{SGDM}}(F) = \expect{\min_{t\in[T]}\norm{\nabla F(\WB_t)}_*}
 \ge c\sqrt{L_*\Delta}\max\{b_T,d_T,s_T\}.
\end{equation}
In particular, for all sufficiently large horizons
\begin{equation}\label{eq:direct-sgdm-horizon}
 T\ge T_0:=\left\lceil\max\left\{m,\,\frac{m^4\norm{\vSigma}_F^2}{L_*\Delta}\right\}\right\rceil,
\end{equation}
a universal constant $c_{\mathrm L}>0$ satisfies
\begin{equation}\label{eq:direct-sgdm-large}
 \Rt_T^{\mathrm{SGDM}}(F)
 \ge c_{\mathrm L}\left[
 \sqrt{\frac{mL_*\Delta}{T}}+\left(\frac{m^2\norm{\vSigma}_F^2 L_*\Delta}{T}\right)^{1/4}\right].
\end{equation}
\end{theorem}

\begin{proof}[Proof sketch of \cref{thm:direct-matrix-sgdm}]
Extending the vanilla \SGDRef{} lower bound to \SGDMRef{} is straightforward when momentum is weak, but strong momentum ($\alpha := 1-\beta \ll 1$) introduces two new obstacles: initial momentum persists over $\Theta(1/\alpha)$ updates, and a column pulse leaves residual velocity that can disrupt sign switching. Working in the normalized eigenbasis setting $L_*=\Delta=1$ (\cref{app:matrix-rescaling}), we resolve both obstacles in three steps.
\begin{enumerate}
    \item \textbf{Ruling out strong-momentum inertia.}
    When $\alpha$ is very small relative to $\eta$, the memoryless diagonal Huber instance no longer suffices because initial momentum carries the iterate much farther. We build a deterministic driver-and-guard instance (\cref{app:direct-inertia}) that turns this persistence into a lower bound and rules out small $\alpha$.

    \item \textbf{Taming momentum overshoot with cooldown pulses.}
    In the remaining pulse regime, residual velocity after a column pulse can push the control entry past its target interval. We resolve this (\cref{app:direct-reset-oracle}) by tracking a \emph{momentum-free shadow iterate} and inserting an exact-gradient cooldown window of length $\Theta(1/\alpha)$ after every pulse so residual velocity decays before the next draw. A complementary diagonal signal preserves the nuclear-norm gradient during sign crossings.

    \item \textbf{Covering all $(\eta,\beta)$ and simplifying for large $T$.}
    Combining the deterministic step-size, circular, inertia, and cooldown-pulse constructions covers the entire $(\eta,\beta)$ plane and yields the three all-horizon branches $(b_T,d_T,s_T)$ in \eqref{eq:direct-sgdm-branches}: $b_T$ comes from the deterministic step-size obstructions, $d_T$ balances the circular noise oracle against them, and $s_T$ combines the inertia and cooldown-pulse constructions in the residual regime (\cref{app:direct-sgdm-interpolation}). Once $T\ge T_0$, the $(m^2v/T)^{1/4}$ term dominates $s_T$ and recovers the same large-horizon rate \eqref{eq:direct-sgdm-large} as vanilla \SGDRef{}.
\end{enumerate}
The complete proof is given in \cref{app:sgdm-lower-bound}.
\end{proof}

To confirm that \eqref{eq:direct-sgdm-large} is tight, we also establish a matching upper bound for \SGDMRef{} (\cref{app:direct-sgdm-upper}).

\begin{theorem}[SGDM upper bound]\label{thm:direct-sgdm-upper}
Suppose \cref{ass:Matrix-bounded-below,ass:Matrix-variance,ass:operator norm smooth} hold with $L_*,\Delta>0$. For every integer $T\ge1$, by setting $\beta=1/2$, $\eta=\min\left\{\frac1{4L_*},\sqrt{\frac{\Delta}{5L_*\norm{\vSigma}_F^2 T}}\right\}$ when $\vSigma\ne\vzero$, and $\eta=\frac1{4L_*}$ when $\vSigma=\vzero$, the \SGDMRef{} iterates in \cref{alg:matrix-optimizers} satisfy
\begin{equation}\label{eq:direct-sgdm-upper}
 \frac1T\sum_{t=1}^T\E\norm{\nabla F(\WB_t)}_*
 =\bigO{\sqrt{\frac{mL_*\Delta}{T}}+
 \left(\frac{m^2\norm{\vSigma}_F^2 L_*\Delta}{T}\right)^{1/4}}.
\end{equation}
\end{theorem}

Combining \cref{thm:direct-matrix-sgdm,thm:direct-sgdm-upper} yields the tight complexity of \SGDMRef{} (\cref{app:sgdm-complexity}), proving that momentum provides no asymptotic improvement over vanilla \SGDRef{} on this matrix problem class.

\begin{theorem}[Tight complexity of SGDM]\label{thm:sgdm-tight-complexity}
Under \cref{ass:Matrix-bounded-below,ass:Matrix-variance,ass:operator norm smooth} with $2\le m\le n$ and $L_*,\Delta>0$, for all sufficiently small $\epsilon>0$, the worst-case number of stochastic-gradient queries required by \SGDMRef{} (\cref{alg:matrix-optimizers}) to achieve $\Rt_T^{\mathrm{SGDM}}(F)\le\epsilon$ is
\begin{equation}\label{eq:direct-sgdm-complexity}
 \Theta\brac{\frac{mL_*\Delta}{\epsilon^2}
 +\frac{m^2\norm{\vSigma}_F^2 L_*\Delta}{\epsilon^4}}.
\end{equation}
\end{theorem}

\subsection{Theoretical Separation: When and Why Muon Wins}\label{sec:theo_improve}

With the tight complexity characterizations established for \MuonRef{} (\cref{thm:muon-tight-complexity}), vanilla \SGDRef{} (\cref{thm:sgd-tight-complexity}), and \SGDMRef{} (\cref{thm:sgdm-tight-complexity}), we can now rigorously compare their worst-case complexities on the shared matrix problem class (\cref{ass:Matrix-bounded-below,ass:Matrix-variance,ass:operator norm smooth}). For any sufficiently small target accuracy $\epsilon>0$ (with $m\ge 3$ and $\vSigma=\vzero$ or $\rank(\vSigma)\ge 2$), the complexities to guarantee $\expect{\min_{t\in[T]}\norm{\nabla F(\WB_t)}_*}\le\epsilon$ compare as follows:
\begin{formulacomparison}
\begin{comparisonpanel}{comparisonblue}
 \textbf{Muon}
 \begin{equation}\label{eq:muon-sgd-upper-complexity}
 \Theta\brac{\frac{L_*\Delta}{\epsilon^2}
 +\frac{\textcolor{red}{\norm{\vSigma}_*^2}L_*\Delta}{\epsilon^4}}
 \end{equation}
\end{comparisonpanel}\hfill
\begin{comparisonpanel}{comparisongold}
 \textbf{SGD / SGDM}
 \begin{equation}\label{eq:matrix-sgd-complexity}
 \Theta\brac{\frac{\textcolor{red}{m}L_*\Delta}{\epsilon^2}
 +\frac{\textcolor{red}{m^2\norm{\vSigma}_F^2}L_*\Delta}{\epsilon^4}}
 \end{equation}
\end{comparisonpanel}
\end{formulacomparison}

For $\vSigma\ne\vzero$, recall from \eqref{eq:intro-spectral-concentration} the spectral concentration $\chi(\vSigma)=m\norm{\vSigma}_F^2/\norm{\vSigma}_*^2\in[1,m]$. Comparing the highlighted coefficients in \eqref{eq:muon-sgd-upper-complexity} and \eqref{eq:matrix-sgd-complexity} shows that \MuonRef{} achieves an order-$m$ improvement in the deterministic term ($\epsilon^{-2}$) and an improvement in the stochastic term ($\epsilon^{-4}$) governed by the ratio
\begin{equation}\label{eq:matrix-noise-ratio}
 m\cdot\chi(\vSigma)
 =\frac{m^2}{\erank(\vSigma)}
 =\frac{m^2\norm{\vSigma}_F^2}{\norm{\vSigma}_*^2}\in[m,m^2].
\end{equation}

This comparison directly answers both \emph{why} \MuonRef{} outperforms \SGDRef{} and \SGDMRef{}, and \emph{when} its speedup is most pronounced.
\begin{enumerate}
 \item \textbf{Why Muon wins: eliminating dimensional dilution.}
 Under operator-norm smoothness and nuclear-norm stationarity, Euclidean updates (\SGDRef{} and \SGDMRef{}) distribute progress unevenly across singular values and suffer diagonal gradient dilution (\cref{sec:sgd-lower-bound}), incurring an unavoidable factor of $m$ in the deterministic term and at least a factor of $m$ in the stochastic term regardless of momentum tuning. By contrast, \MuonRef{}'s polar update $\msign{\MB_t}$ normalizes all nonzero singular directions to unit magnitude, matching the operator/nuclear norm duality and completely eliminating the matrix dimension $m$ from its complexity.
 \item \textbf{When Muon wins most: anisotropic, spectrally concentrated gradient noise.}
 In the stochastic regime ($\epsilon^{-4}$), \MuonRef{}'s baseline $\Theta(m)$ advantage is further amplified by the spectral concentration $\chi(\vSigma) = m/\erank(\vSigma) \in [1,m]$. When the gradient noise is isotropic ($\erank(\vSigma)=m$ and $\chi(\vSigma)=1$), \MuonRef{} already improves over \SGDRef{} and \SGDMRef{} by a factor of $\Theta(m)$. When the gradient noise is strongly anisotropic and concentrated along a few singular directions ($\erank(\vSigma)\ll m$, so $\chi(\vSigma)\gg 1$), the stochastic complexity ratio $m\chi(\vSigma)$ grows up to $\Theta(m^2)$.
\end{enumerate}

\begin{remark}[Consistency with structured-noise analyses]
Our complexity-theoretic conclusion that \MuonRef{} benefits most from anisotropic gradient noise aligns with recent analyses in specific learning models~\citep{chen2026muonprecondition,wang2026signsgdfsl}. In associative memory models,~\citet{chen2026muonprecondition} show that \MuonRef{} offers no preconditioning speedup under isotropic noise, whereas under anisotropic mini-batch noise it softens the dependence on task frequency from linear to square-root. Similarly,~\citet{wang2026signsgdfsl} show via functional scaling laws in power-law regression that sign-based updates accelerate convergence when gradient noise is anisotropic and curvature-aligned. While those works study learning dynamics in specialized data models, we establish this phenomenon directly from worst-case optimization complexity, proving that \MuonRef{}'s speedup over \SGDRef{} and \SGDMRef{} scales with $\chi(\vSigma)$.
\end{remark}

In \cref{sec:experiments}, we test this theoretical prediction empirically by measuring the noise spectrum across deep learning benchmarks. We examine whether settings where \MuonRef{} substantially outperforms \SGDMRef{} exhibit strong noise anisotropy ($\chi(\vSigma)\gg 1$), and whether settings where \MuonRef{}'s advantage vanishes correspond to nearly isotropic noise ($\chi(\vSigma)\approx 1$).

\begingroup
\raggedbottom
\section{Experiments}\label{sec:experiments}

We evaluate \MuonRef{} and \SGDMRef{} in two complementary empirical settings to test both \emph{when} \MuonRef{} outperforms \SGDMRef{} and \emph{why} this separation arises. First, we conduct controlled pretraining comparisons across four benchmarks spanning CNN and LLM~\citep{NIPS2017transformer} architectures, namely ResNet-20~\citep{he2016deep} on CIFAR-10~\citep{krizhevsky2009learning} image classification, alongside nanoGPT-124M~\citep{radford2019language,nanogpt}, LLaMA-130M, and LLaMA-1.34B~\citep{touvron2023llama,lialin2023relora,zhao2024galore} on C4~\citep{JMLR:v21:20-074} pretraining. Second, we scale up to the 7B regime to examine the empirical gradient-noise distribution of large-scale foundation models, probing open training checkpoints of Pythia-6.9B~\citep{biderman2023pythia} pretrained on The Pile~\citep{gao2020pile}. We begin in \cref{sec:optimization-performance} by identifying \emph{when} \MuonRef{} empirically outperforms \SGDMRef{} across these benchmarks. To investigate \emph{why} this performance gap emerges, \cref{sec:noise-measurement} introduces our matrix-noise estimation protocol, \cref{sec:noise-concentration} contrasts the empirical spectral concentration $\chi_b$ between CNNs and LLMs up to the 7B scale, and \cref{sec:dimension-spectrum} connects these observations back to our theoretical complexity ratio. Complete training configurations and additional diagnostics are provided in \cref{sec:experimental-setup}.

\begin{figure}[t!]
  \centering
  \includegraphics[width=0.86\textwidth]{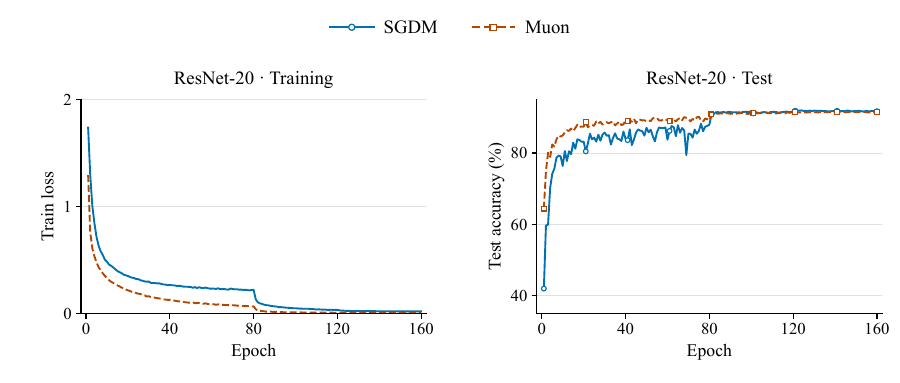}
  \caption{ResNet-20 training loss and test accuracy on CIFAR-10 over 160 epochs under \SGDMRef{} and \MuonRef{}.}
  \label{fig:cnn-performance}
\end{figure}

\subsection{Optimization Performance: When Muon Outperforms SGDM}\label{sec:optimization-performance}

\Cref{fig:cnn-performance,fig:llm-performance} compare the training and validation trajectories of \MuonRef{} and \SGDMRef{} across the four pretraining benchmarks to determine \emph{when} \MuonRef{} delivers a genuine empirical advantage.
\begin{itemize}
    \item \textbf{CNN benchmark (ResNet-20 on CIFAR-10).} As shown in \cref{fig:cnn-performance}, \MuonRef{} lowers training loss and improves test accuracy faster than \SGDMRef{} before the learning-rate decay at epoch $80$. After the decay, however, \SGDMRef{} quickly closes the gap, and both methods reach virtually identical final training loss and test accuracy ($\approx 91.5\%$).
    \item \textbf{LLM benchmarks (nanoGPT and LLaMA on C4).} As shown in \cref{fig:llm-performance}, a stark contrast emerges in LLM pretraining. On both nanoGPT-124M and LLaMA-130M ($2.6$B training tokens), \MuonRef{} converges substantially faster and achieves much lower final training and validation losses than \SGDMRef{}. At the 1.34B LLaMA scale ($1.0$B training tokens), this gap widens further, with \MuonRef{} reducing final validation loss to $3.22$ (perplexity $25.0$) compared with $6.22$ (perplexity $502.8$) for \SGDMRef{}.
\end{itemize}
These results directly answer \emph{when} \MuonRef{} wins in practice: its empirical advantage is not universal across all neural networks, remaining marginal on the CNN benchmark while becoming substantial and consistent across all three LLMs.

\begin{figure}[t!]
  \centering
  \includegraphics[width=0.92\textwidth]{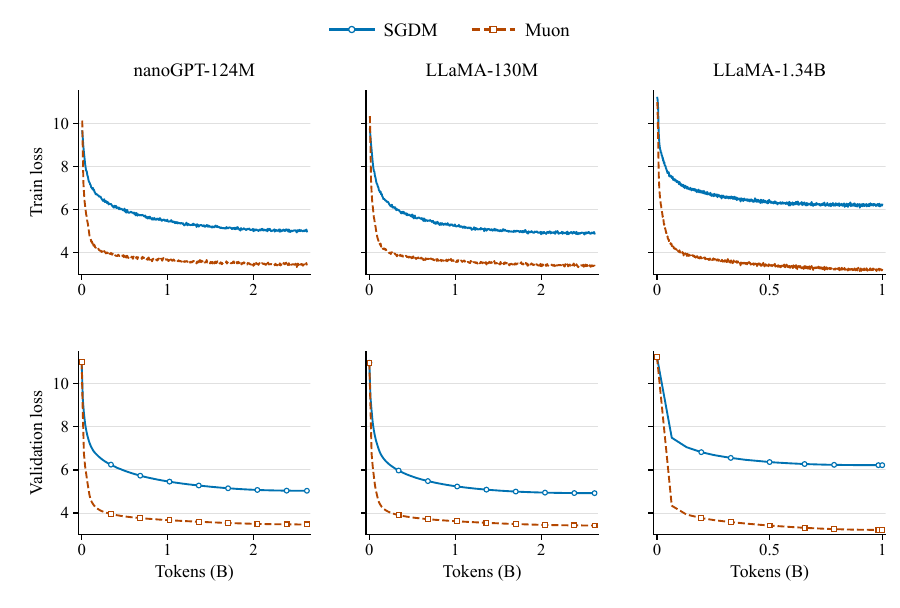}
  \caption{Training (top) and validation (bottom) cross-entropy loss on C4 for nanoGPT-124M, LLaMA-130M, and LLaMA-1.34B under \SGDMRef{} and \MuonRef{} versus training tokens (in billions).}
  \label{fig:llm-performance}
\end{figure}

\subsection{Matrix-Noise Measurement}\label{sec:noise-measurement}
To understand \emph{why} \MuonRef{} excels on LLMs yet offers little final gain on ResNet-20, our theoretical separation in \eqref{eq:matrix-noise-ratio} points to the spectral concentration $\chi(\vSigma)$ of the stochastic gradient noise. We therefore measure the empirical gradient-noise spectrum at frozen training checkpoints. At each checkpoint, we freeze the model parameters, disable dropout, and evaluate $K$ independent minibatch gradients $\{G_b^{(k)}\}_{k=1}^K$ of the data loss prior to momentum, weight decay, or optimizer updates.

For each weight matrix $b$, we orient its gradient $G_b^{(k)}\in\R^{m_b\times n_b}$ with $m_b\le n_b$ and compute the sample covariance
\begin{equation}\label{eq:empirical-matrix-noise}
\bar G_b=\frac1K\sum_{k=1}^K G_b^{(k)},\qquad
\widehat C_b=\frac1{K-1}\sum_{k=1}^K
(G_b^{(k)}-\bar G_b)(G_b^{(k)}-\bar G_b)^\top.
\end{equation}
Setting $\widehat\Sigma_b=\widehat C_b^{1/2}$, we compute the empirical spectral concentration
\[
\chi_b=\frac{m_b}{\erank(\widehat\Sigma_b)}
=\frac{m_b\norm{\widehat\Sigma_b}_F^2}
{\norm{\widehat\Sigma_b}_*^2}\in[1,m_b].
\]
Here $\chi_b\approx 1$ corresponds to an approximately isotropic noise spectrum, whereas $\chi_b\gg 1$ indicates strong spectral concentration along a small number of singular directions. We evaluate $\chi_b$ across all 2D weight kernels in ResNet-20 and all attention and MLP projection matrices in the LLMs, excluding biases, normalization layers, embeddings, and output heads.

\subsection{Noise Concentration across Models and Training Stages}
\label{sec:noise-concentration}

\begin{figure}[t!]
  \centering
  \includegraphics[width=0.86\textwidth]{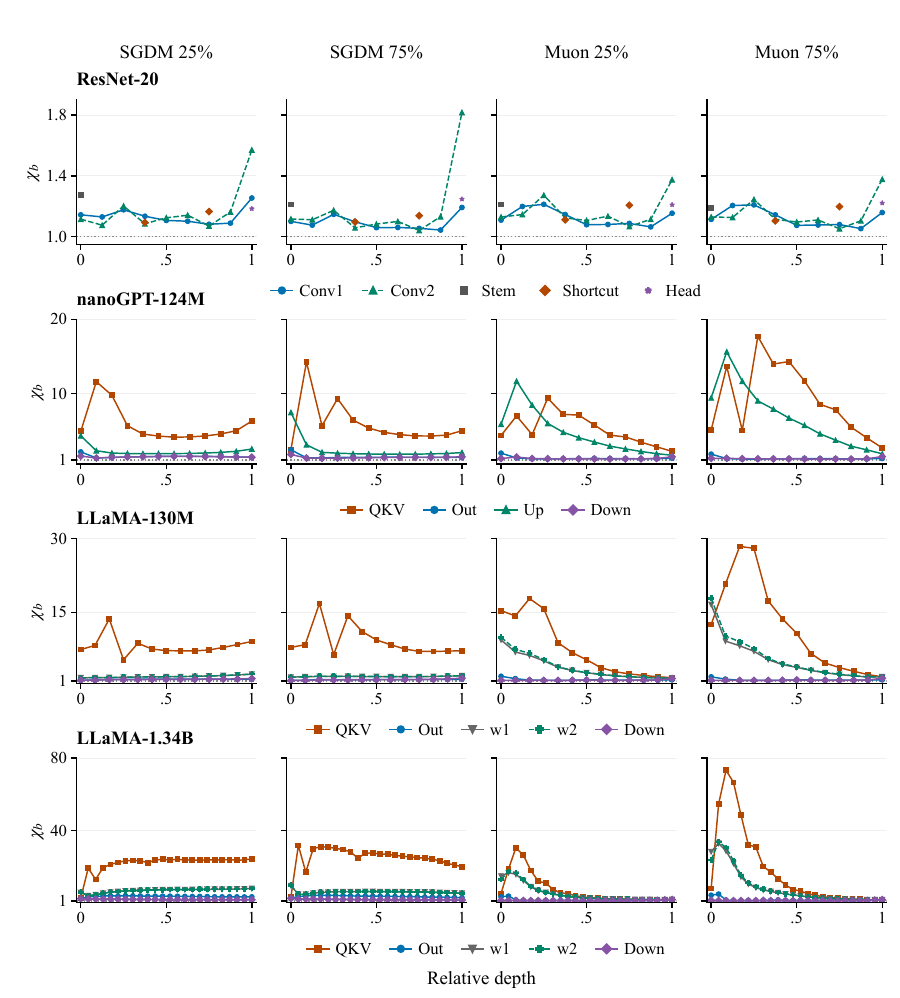}
  \caption{Layerwise spectral concentration $\chi_b$ versus relative depth across ResNet-20, nanoGPT-124M, LLaMA-130M, and LLaMA-1.34B at 25\% and 75\% of training along \SGDMRef{} and \MuonRef{} trajectories. Dotted lines indicate the isotropic baseline $\chi_b=1$.}
  \label{fig:appendix-matrix-chi-depth}
\end{figure}

We now examine the measured coefficients $\chi_b$ to see whether the empirical noise distributions explain the performance contrast in \cref{sec:optimization-performance}.

\paragraph{Distinct noise regimes in CNNs versus LLMs.}
\Cref{fig:appendix-matrix-chi-depth} plots $\chi_b$ across network depth at $25\%$ and $75\%$ of training along both \SGDMRef{} and \MuonRef{} trajectories, revealing a striking separation between the two domains. In the ResNet-20 CNN benchmark (\cref{fig:appendix-matrix-chi-depth}, top row), every weight matrix remains uniformly close to the isotropic baseline ($\chi_b\le 1.82$) across all layers, both training stages, and both optimizers. By contrast, across all three LLMs (\cref{fig:appendix-matrix-chi-depth}, rows 2--4), gradient noise is highly anisotropic ($\chi_b\gg 1$), with spectral concentration reaching over $15$ in nanoGPT-124M, nearly $30$ in LLaMA-130M, and over $70$ in LLaMA-1.34B. This stark contrast between nearly isotropic noise in ResNet-20 and strongly concentrated noise in LLMs provides a direct empirical explanation for \emph{why} \MuonRef{} substantially accelerates LLM pretraining while yielding negligible final improvement on ResNet-20, exactly as predicted by our complexity ratio $m\chi(\vSigma)$ in \eqref{eq:matrix-noise-ratio}.

\begin{figure}[t!]
  \centering
  \includegraphics[width=0.92\textwidth]{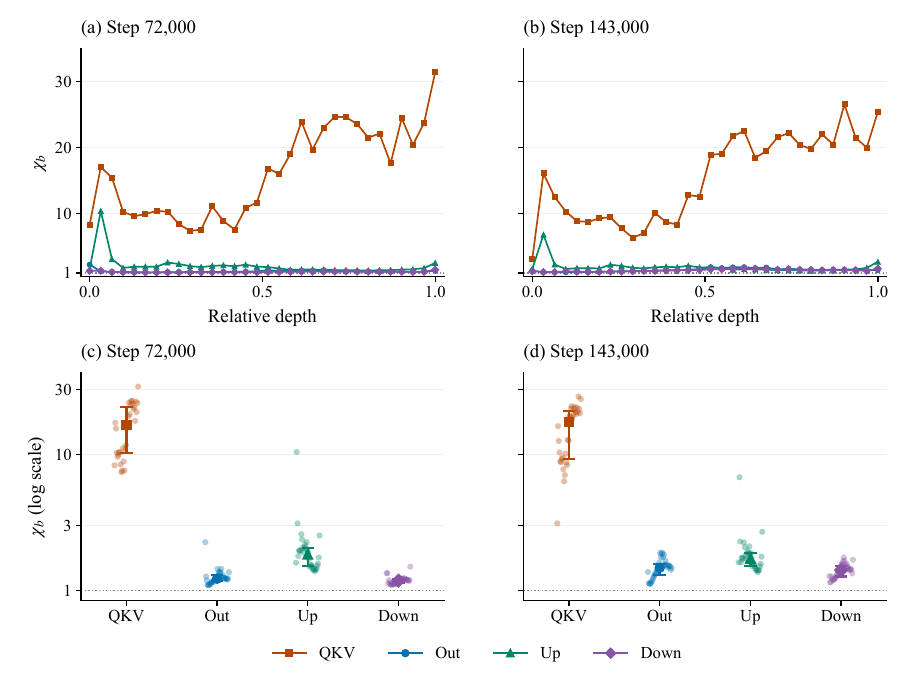}
  \caption{Spectral concentration $\chi_b$ in 7B-scale LLM checkpoints (Pythia-6.9B) at steps 72,000 (left) and 143,000 (right), shown across relative depth (top) and by module family with medians and interquartile ranges on a log scale (bottom). Dotted lines indicate $\chi_b = 1$.}
  \label{fig:pythia-6p9b}
\end{figure}

\paragraph{Scaling up to the 7B regime and module-level heterogeneity.}
To verify that this strong noise anisotropy is a persistent property of large-scale LLMs rather than an artifact of smaller scales, we probe open checkpoints of the 7B-scale model Pythia-6.9B~\citep{biderman2023pythia} (32 layers, $m_b = 4096$, trained on The Pile~\citep{gao2020pile}) at mid-training (\texttt{step72000}, $50\%$) and completion (\texttt{step143000}, $100\%$). As shown in \cref{fig:pythia-6p9b}, strong spectral concentration persists throughout 7B pretraining, with the fused query-key-value (\texttt{QKV}) projections maintaining cross-layer medians of $16.4$ and $17.3$ and deep-layer values reaching $31.4$.

As an interesting byproduct of our block-level diagnostics across \cref{fig:appendix-matrix-chi-depth,fig:pythia-6p9b}, we also observe a consistent hierarchy across LLM module types: \texttt{QKV} projections exhibit the strongest noise anisotropy (followed by MLP expansion projections in the smaller models), whereas attention output (\texttt{Out}) and MLP down-projection (\texttt{Down}) matrices remain much closer to the isotropic baseline ($\chi_b \approx 1\text{--}3$)\footnote{Comparing optimizer trajectories in \cref{fig:appendix-matrix-chi-depth} further shows that while high \texttt{QKV} concentration spans all depths under \SGDMRef{}, \MuonRef{} concentrates large $\chi_b$ in shallower layers and drives deeper layers toward $1\text{--}3$, suggesting that \MuonRef{} actively equalizes singular directions in deeper blocks.}. Two factors explain why this module-level heterogeneity strongly favors \MuonRef{}: first, even on blocks where $\chi_b\approx 1\text{--}2$, \eqref{eq:matrix-noise-ratio} guarantees that \MuonRef{} still enjoys a large baseline dimensional speedup of order $m_b$ (with $m_b=4096$ at the 7B scale); second, prior work has established that attention \texttt{QKV} weights constitute the primary optimization bottleneck in Transformers due to rank collapse, ill-conditioned attention logits, and heavy-tailed gradient noise~\citep{dong2021attention,noci2022signal,tarzanagh2023maxmargin,zhang2024adam_hessian,kunstner2024heavytail,vasudeva2025muon}. Consequently, the massive $\chi_b$-amplified speedup that \MuonRef{} achieves specifically on the bottleneck \texttt{QKV} projections translates directly into end-to-end pretraining acceleration.

\subsection{Dimension and Spectral Concentration}
\label{sec:dimension-spectrum}
By \eqref{eq:matrix-noise-ratio}, the stochastic complexity ratio between \SGDRef{}/\SGDMRef{} and \MuonRef{} decomposes as
$m\chi(\vSigma)=m^2/\erank(\vSigma)$, where the spectral concentration in \eqref{eq:intro-spectral-concentration} satisfies
$\chi(\vSigma)=m/\erank(\vSigma)$.
Here, the dimension factor $m$ provides the baseline separation shared with the deterministic term, while $\chi(\vSigma)$ captures the stochastic amplification from noise anisotropy. An isotropic full-rank spectrum has $\erank(\vSigma)=m$ and $\chi(\vSigma)=1$, yielding the baseline ratio $m$. Thus, a large matrix dimension alone does not yield a large $\chi(\vSigma)$ unless the noise spectrum is also anisotropic.

Our empirical measurements across all five models support this two-factor decomposition explaining \emph{when} and \emph{why} \MuonRef{} wins. In ResNet-20, weight matrices have modest dimensions ($m_b \in [16, 64]$) and nearly isotropic noise spectra ($\chi_b < 2$). In the LLMs, large hidden dimensions ($m_b \in \{768, 2048, 4096\}$) combine with strong spectral concentration ($\chi_b \gg 1$ in bottleneck attention blocks) to produce a large product $m_b\chi_b$, even though $\chi_b$ remains well below the rank-one limit $m_b$. Finally, because our theoretical framework treats $\vSigma$ as a prescribed static noise envelope over the problem class (\cref{ass:Matrix-variance}), theoretically modeling how attention mechanics and optimizer trajectories give rise to the observed module-level spectral heterogeneity is a promising direction for future work.

\begin{remark}[Connection to nuclear rank in local descent analyses]
A related geometric quantity is studied by \citet{davis2025spectral}, who define the squared norm ratio $\norm{\Gb}_*^2/\norm{\Gb}_F^2$ of the gradient matrix as its \emph{nuclear rank} (the reciprocal of $\chi(\Gb)/m$) to analyze single-step local descent. While their analysis compares the nuclear rank of the gradient against the stable rank of incoming activations to predict when a single spectral step decreases the loss more than a Euclidean step, our multistep complexity bound characterizes the noise-dominated stochastic regime: we show that a small effective rank $\erank(\vSigma)=\norm{\vSigma}_*^2/\norm{\vSigma}_F^2\ll m$ of the gradient noise (i.e., $\chi(\vSigma)\gg 1$) severely penalizes Euclidean updates and amplifies \MuonRef{}'s worst-case complexity advantage up to $\Theta(m^2)$.
\end{remark}

\section{Conclusion}\label{sec:conclusion}
In this paper, we answered two fundamental questions about matrix-sign optimization: \emph{why} \MuonRef{} can outperform classical Euclidean gradient methods despite the minimax optimality of \SGDRef{} under standard Euclidean assumptions, and \emph{when} this advantage is most pronounced. By formulating smooth matrix optimization under operator-norm smoothness, nuclear-norm stationarity, and a matrix-valued noise covariance bound $\vSigma^2$, we established matching upper and lower complexity bounds for \MuonRef{} (\cref{thm:muon-tight-complexity}), vanilla \SGDRef{} (\cref{thm:sgd-tight-complexity}), and \SGDMRef{} (\cref{thm:sgdm-tight-complexity}) on a common problem class.

Our tight bounds provide a clear two-part answer. First, \emph{why} \MuonRef{} wins lies in its non-Euclidean update geometry: by normalizing all nonzero singular directions to unit magnitude, \MuonRef{} achieves a dimension-free complexity of $\Theta\big(\nicefrac{L_*\Delta}{\epsilon^2}+\nicefrac{\norm{\vSigma}_*^2 L_*\Delta}{\epsilon^4}\big)$, whereas \SGDRef{} and \SGDMRef{} suffer dimensional dilution across singular values and incur a tight complexity of $\Theta\big(\nicefrac{mL_*\Delta}{\epsilon^2}+\nicefrac{m^2\norm{\vSigma}_F^2 L_*\Delta}{\epsilon^4}\big)$ that momentum alone cannot remove. Second, \emph{when} \MuonRef{} wins most is governed by the spectral concentration $\chi(\vSigma)=m/\erank(\vSigma)\in[1,m]$ of the stochastic gradient noise: while \MuonRef{} enjoys a baseline $\Theta(m)$ speedup even under isotropic noise ($\chi(\vSigma)=1$), strong noise anisotropy ($\chi(\vSigma)\gg 1$) amplifies its stochastic advantage to $m\chi(\vSigma)$, reaching up to $\Theta(m^2)$ when the noise is concentrated along a few singular directions. Our empirical study across CNNs and LLMs up to the 7B scale confirms this mechanism in practice, showing that \MuonRef{}'s large pretraining gains in LLMs coincide with strong spectral noise concentration in bottleneck attention blocks, whereas its marginal gain in CNN training coincides with nearly isotropic noise.

\clearpage
\endgroup
\appendix

\crefalias{section}{appendix}
\crefalias{subsection}{appendix}
\crefalias{subsubsection}{appendix}

\clearpage

\section{Analysis for Muon}
\label{app:muon-analysis}

\Cref{sec:proof-muon-upperbound} proves the \MuonRef{} upper bound. For the lower bound, \cref{sec:proof-ssgd-lowerbound} constructs hard instances for SignSGD with momentum, and \cref{sec:proof-muon-lowerbound} embeds them into \MuonRef{} and derives its tight complexity. The supporting proofs are in \cref{app:muon-supporting-lemmas}.

\subsection{Upper Bound for Muon}
\label{sec:proof-muon-upperbound}

This subsection proves the dimension-free convergence upper bound for \MuonRef{} (\cref{thm:muon_upper_infty}) and carries out the complexity conversion used in \cref{sec:muon-tight-characterization}. Throughout the first two lemmas, $\norm{\cdot}$ denotes an arbitrary norm on a finite-dimensional real inner-product space and $\norm{\cdot}_*$ denotes its dual norm (in particular, $\norm{\cdot}_{\op}$ and $\norm{\cdot}_*$ for matrices, or $\norm{\cdot}_\infty$ and $\norm{\cdot}_1$ for vectors).

To convert each \MuonRef{} update $\WB_{t+1}-\WB_t=-\eta\msign{\MB_t}$ into a one-step descent inequality in Step~1 of the proof of \cref{thm:muon_upper_infty}, we first record the quadratic upper bound induced by dual-norm gradient Lipschitz continuity.

\begin{lemma}[Quadratic Taylor remainder]
\label{lem:gradient-taylor-bound}
Let $\norm{\cdot}$ be any norm with dual norm $\norm{\cdot}_*$. If a differentiable function $h$ satisfies $\norm{\nabla h(\v)-\nabla h(\u)}_* \le L\norm{\v-\u}$ for all $\u,\v$, then
\[
\abs{h(\v)-h(\u)-\inner{\nabla h(\u)}{\v-\u}} \le \frac{L}{2}\norm{\v-\u}^2.
\]
\end{lemma}

Because \MuonRef{} steps along the steepest direction of the noisy momentum $\MB_t = \nabla F(\WB_t) + \EB_t$ rather than the true gradient $\nabla F(\WB_t)$, Step~1 of the proof of \cref{thm:muon_upper_infty} must bound the directional error $\norm{\g}_* - \inner{\g}{\s(\g+\bta)}$. Bounding this gap by $2\min\{\norm{\g}_*,\norm{\bta}_*\}$ instead of the cruder $2\norm{\bta}_*$ lets us truncate large initial noise at $t=1$ against the true gradient norm.

\begin{lemma}[Perturbation bound for a steepest direction]
\label{lem:upper-steepest-truncation}
Let $\norm{\cdot}$ be any norm with dual norm $\norm{\cdot}_*$. For each $\z$, choose $\s(\z) \in \argmax_{\norm{\s}\le 1} \inner{\z}{\s}$, so that $\norm{\s(\z)}\le 1$ and $\inner{\z}{\s(\z)} = \norm{\z}_*$. Then, for any $\g$ and perturbation $\bta$,
\[
0 \le \norm{\g}_* - \inner{\g}{\s(\g+\bta)} \le 2\min\bbrac{\norm{\g}_*, \norm{\bta}_*}.
\]
In particular, if $\bta=\u+\mathbf r$, then
\[
\norm{\g}_* - \inner{\g}{\s(\g+\u+\mathbf r)} \le 2\min\bbrac{\norm{\g}_*, \norm{\u}_*} + 2\norm{\mathbf r}_*.
\]
\end{lemma}

Under the initialization $\MB_1=\Gb_1$, the initial noise $\Nb_1$ enters the momentum error as a geometrically decaying term $\beta^{t-1}\Nb_1$. In Step~4 of the proof of \cref{thm:muon_upper_infty}, we use the next lemma to sum the resulting truncated startup errors over $t\in[T]$, replacing a linear dependence on $R=\norm{\vSigma}_*$ with two logarithmic contributions proportional to $GH/\alpha$ and $qH^2/\alpha^2$.

\begin{lemma}[EMA startup-error truncation]
\label{lem:upper-geometric-truncation}
Let $G>0$, $R,q\geq0$, $0<\alpha<1$, and $\beta=1-\alpha$. Define $H:=\log\brac{e+\nicefrac{R}{G}}$. For every integer $T\geq1$,
\[
\sum_{n=0}^{T-1} \min\bbrac{G+qn,\, R\beta^n} \leq \frac{4GH}{\alpha} + \frac{2qH^2}{\alpha^2}.
\]
\end{lemma}

After the initial noise is separated out, Step~2 of the proof of \cref{thm:muon_upper_infty} must control the EMA accumulation of subsequent stochastic gradient noise $\sum_{k=2}^t \alpha\beta^{t-k}\Nb_k$. The next lemma bounds the expected nuclear norm of any weighted sum of matrix martingale differences directly in terms of $\norm{\vSigma}_*$, without passing through Frobenius-norm relaxations.

\begin{lemma}[Filtered matrix martingale noise]
\label{lem:matrix-filtered-noise}
Let $\{\Nb_k\}_{k=1}^t\subset\R^{m\times n}$ be an adapted matrix martingale difference sequence ($\expect{\Nb_k\mid\F_{k-1}}=\vzero$) satisfying $\expect{\Nb_k\Nb_k^\top \mid\F_{k-1}}\preceq\vSigma^2$ for $\vSigma\in\SBB_+^m$. For any deterministic scalars $a_1,\ldots,a_t\in\R$ and $\mathbf Z:=\sum_{k=1}^ta_k\Nb_k$,
\begin{equation}\label{eq:matrix-filtered-noise}
\expect{\mathbf Z\mathbf Z^\top} \preceq \brac{\sum_{k=1}^ta_k^2}\vSigma^2,
\qquad
\E\norm{\mathbf Z}_* \leq \norm{\vSigma}_* \sqrt{\sum_{k=1}^ta_k^2}.
\end{equation}
\end{lemma}

We are now ready to prove \cref{thm:muon_upper_infty}.

\begin{proof}[Proof of \cref{thm:muon_upper_infty}]
\paragraph{Preliminaries and hyperparameter setup.}
Throughout the proof, write $\alpha:=1-\beta$ and define the shorthands
\[
\nabla f_t := \nabla F(\WB_t),
\quad
\Nb_t := \Gb_t - \nabla f_t,
\quad
\EB_t := \MB_t - \nabla f_t,
\quad
\SB_k := \nabla f_{k-1} - \nabla f_k\ \ (k\ge 2).
\]
If $\Delta=0$, then $\WB_1$ is a global minimizer and $\nabla f_1=\vzero$; if $L_*=0$, then $\nabla F$ is constant and lower boundedness forces $\nabla f_1=\vzero$. In either case $\Rt_T^{\mathrm{Muon}}(F)=0$. Assume now that $L_*,\Delta>0$, and choose the hyperparameters
\begin{equation}\label{eq:hyp_muon}
\alpha=
\begin{cases}
1, & \norm{\vSigma}_*=0,\\[2pt]
\displaystyle\min\bbrac{1,\frac{\sqrt{L_*\Delta}}{\norm{\vSigma}_*\sqrt T}}, & \norm{\vSigma}_*>0,
\end{cases}
\qquad
\eta:=\sqrt{\frac{\Delta\alpha}{L_*T}}.
\end{equation}
Define the initial gradient scale, logarithmic factor, and horizon threshold
\begin{equation}\label{eq:muon-upper-budget}
G_*:=\sqrt{2L_*\Delta},
\qquad
H_*:=\log\brac{e+\frac{\norm{\vSigma}_*}{\sqrt{2L_*\Delta}}},
\qquad
T_{{\rm up},*}:=\max\bbrac{1,\frac{\norm{\vSigma}_*^2}{L_*\Delta}H_*^4}.
\end{equation}
We will prove that for every integer $T\ge T_{{\rm up},*}$,
\begin{equation}\label{eq:muon-upper-explicit}
\frac1T\SumT\E\norm{\nabla F(\WB_t)}_*
\leq 21\sqbrac{\sqrt{\frac{L_*\Delta}{T}} + \brac{\frac{\norm{\vSigma}_*^2{L_*\Delta}}{T}}^{1/4}}.
\end{equation}
First, note a deterministic pathwise bound on the gradient norm along the \MuonRef{} trajectory. By the initial gradient bound~\eqref{eq:direct-initial-cap} (established in \cref{app:sgd-upper-cap}), $\norm{\nabla f_1}_*\le G_*$. Moreover, each \MuonRef{} step satisfies $\norm{\WB_{k+1}-\WB_k}_{\op}=\eta\norm{\msign{\MB_k}}_{\op}\le\eta$, so \cref{ass:operator norm smooth} gives $\norm{\nabla f_{k+1}-\nabla f_k}_*\le L_*\eta$ for every $k\ge 1$. By the triangle inequality,
\begin{equation}\label{eq:muon-pathwise-growth}
\norm{\nabla f_t}_*
\le \norm{\nabla f_1}_* + \sum_{k=1}^{t-1}\norm{\nabla f_{k+1}-\nabla f_k}_*
\le G_* + L_*\eta(t-1)
\qquad (\forall\, t\ge 1).
\end{equation}

\paragraph{Step 1: One-step descent and momentum-error decomposition.}
Recall from \cref{sec:matrix-assumptions} that $\msign{\MB_t}$ is a steepest direction for the operator/nuclear norm pair, satisfying $\norm{\msign{\MB_t}}_{\op}\le 1$ and $\inner{\MB_t}{\msign{\MB_t}}=\norm{\MB_t}_*$. Applying \cref{lem:gradient-taylor-bound} to $\WB_{t+1}=\WB_t-\eta\msign{\MB_t}$ under \cref{ass:operator norm smooth} gives
\begin{align}
F(\WB_{t+1}) - F(\WB_t)
&\le -\eta\inner{\nabla f_t}{\msign{\MB_t}} + \frac{L_*}{2}\norm{\eta\msign{\MB_t}}_{\op}^2 \nonumber\\
&\le -\eta\norm{\nabla f_t}_* + \eta\brac{\norm{\nabla f_t}_* - \inner{\nabla f_t}{\msign{\MB_t}}} + \frac{L_*\eta^2}{2},
\label{eq:muon-onestep-descent}
\end{align}
where the second inequality uses $\norm{\msign{\MB_t}}_{\op}\le 1$ and adds and subtracts $\eta\norm{\nabla f_t}_*$.

To control the directional error $\norm{\nabla f_t}_* - \inner{\nabla f_t}{\msign{\MB_t}}$, we expand the momentum error $\EB_t=\MB_t-\nabla f_t$. For every $t\ge 2$, the EMA recursion $\MB_t=\beta\MB_{t-1}+\alpha\Gb_t$ yields
\begin{align*}
\EB_t
&= \beta\MB_{t-1} + \alpha\Gb_t - \nabla f_t \\
&= \beta(\MB_{t-1}-\nabla f_{t-1}) + \beta(\nabla f_{t-1}-\nabla f_t) + \alpha(\Gb_t-\nabla f_t) \\
&= \beta\EB_{t-1} + \beta\SB_t + \alpha\Nb_t.
\end{align*}
Unrolling this recursion from the initialization $\EB_1=\MB_1-\nabla f_1=\Gb_1-\nabla f_1=\Nb_1$ decomposes $\EB_t$ into an initial-sample residual, a filtered noise sum, and a gradient-drift sum:
\begin{equation}\label{eq:matrix-ema-decomposition}
\EB_t = \MB_t-\nabla f_t
= \underbrace{\beta^{t-1}\Nb_1}_{\EB_t^{\mathrm{init}}}
  + \underbrace{\alpha\sum_{k=2}^t \beta^{t-k}\Nb_k}_{\EB_t^{\mathrm{noise}}}
  + \underbrace{\sum_{k=2}^t \beta^{t-k+1}\SB_k}_{\EB_t^{\mathrm{drift}}},
\end{equation}
where empty sums at $t=1$ vanish and we adopt the convention $\beta^0=1$ throughout (including when $\beta=0$). Applying \cref{lem:upper-steepest-truncation} to \eqref{eq:muon-onestep-descent} with $\g=\nabla f_t$, $\u=\EB_t^{\mathrm{init}}$, and $\mathbf r=\EB_t^{\mathrm{noise}}+\EB_t^{\mathrm{drift}}$ gives
\begin{align}
F(\WB_{t+1}) - F(\WB_t)
&\le -\eta\norm{\nabla f_t}_* + 2\eta\min\bbrac{\norm{\nabla f_t}_*,\norm{\EB_t^{\mathrm{init}}}_*} \nonumber\\
&\quad + 2\eta\norm{\EB_t^{\mathrm{noise}}}_* + 2\eta\norm{\EB_t^{\mathrm{drift}}}_* + \frac{L_*\eta^2}{2}.
\label{eq:muon-onestep-split}
\end{align}

\paragraph{Step 2: Bounding the three error terms.}
We now bound each error term in \eqref{eq:muon-onestep-split} in expectation.

\smallskip\noindent\emph{(a) Initial-sample residual $\EB_t^{\mathrm{init}}$.}
Directly bounding $\E\norm{\EB_t^{\mathrm{init}}}_*=\beta^{t-1}\E\norm{\Nb_1}_*$ without truncation would yield a $1/(\alpha T)$ term proportional to $\norm{\vSigma}_*$, requiring a polynomially large horizon when $\norm{\vSigma}_*$ is large. Retaining the minimum against $\norm{\nabla f_t}_*$ instead gives
\begin{align}
\expect{\min\bbrac{\norm{\nabla f_t}_*,\norm{\EB_t^{\mathrm{init}}}_*}}
&\le \expect{\min\bbrac{G_*+L_*\eta(t-1),\,\beta^{t-1}\norm{\Nb_1}_*}} \nonumber\\
&\le \min\bbrac{G_*+L_*\eta(t-1),\,\beta^{t-1}\E\norm{\Nb_1}_*} \nonumber\\
&\le \min\bbrac{G_*+L_*\eta(t-1),\,\beta^{t-1}\norm{\vSigma}_*} =: c_t^*,
\label{eq:matrix-startup-bound}
\end{align}
where the first inequality uses \eqref{eq:muon-pathwise-growth}, the second uses Jensen's inequality for the concave function $x\mapsto\min\{a,x\}$, and the third applies \cref{lem:matrix-filtered-noise} at $t=1$ ($\E\norm{\Nb_1}_*\le\norm{\vSigma}_*$).

\smallskip\noindent\emph{(b) Filtered martingale noise $\EB_t^{\mathrm{noise}}$.}
Under \cref{ass:Matrix-variance}, $\{\Nb_k\}_{k=2}^t$ is a matrix martingale difference sequence with $\expect{\Nb_k\Nb_k^\top\mid\F_{k-1}}\preceq\vSigma^2$. Applying \cref{lem:matrix-filtered-noise} with weights $a_k=\alpha\beta^{t-k}$ ($2\le k\le t$) gives
\begin{align}
\E\norm{\EB_t^{\mathrm{noise}}}_*
&\le \norm{\vSigma}_*\sqrt{\alpha^2\sum_{k=2}^t\beta^{2(t-k)}}
 = \norm{\vSigma}_*\sqrt{\alpha^2\sum_{j=0}^{t-2}\beta^{2j}} \nonumber\\
&\le \norm{\vSigma}_*\sqrt{\frac{\alpha^2}{1-\beta^2}}
 = \norm{\vSigma}_*\sqrt{\frac{\alpha}{2-\alpha}}
 \le \norm{\vSigma}_*\sqrt{\alpha},
\label{eq:matrix-stable-noise}
\end{align}
where we used $1-\beta^2=(1-\beta)(1+\beta)=\alpha(2-\alpha)$ and $2-\alpha\ge 1$.

\smallskip\noindent\emph{(c) Gradient drift $\EB_t^{\mathrm{drift}}$.}
Because $\norm{\SB_k}_*=\norm{\nabla f_{k-1}-\nabla f_k}_*\le L_*\norm{\WB_{k-1}-\WB_k}_{\op}\le L_*\eta$ by \cref{ass:operator norm smooth}, the triangle inequality gives
\begin{equation}\label{eq:matrix-drift-bound}
\norm{\EB_t^{\mathrm{drift}}}_*
\le \sum_{k=2}^t \beta^{t-k+1}\norm{\SB_k}_*
\le L_*\eta\beta\sum_{j=0}^{t-2}\beta^j
\le \frac{\beta L_*\eta}{\alpha}.
\end{equation}

\paragraph{Step 3: Telescoping and pre-tuned rate.}
Taking expectations in \eqref{eq:muon-onestep-split}, substituting \eqref{eq:matrix-startup-bound}--\eqref{eq:matrix-drift-bound}, rearranging, and averaging over $t=1,\ldots,T$ yields
\begin{align}
\frac1T\SumT\E\norm{\nabla F(\WB_t)}_*
&\le \frac{F(\WB_1)-\E[F(\WB_{T+1})]}{\eta T} + \frac{L_*\eta}{2} + \frac{2\beta L_*\eta}{\alpha}
     + 2\norm{\vSigma}_*\sqrt{\alpha} + \frac2T\SumT c_t^* \nonumber\\
&\le \frac{\Delta}{\eta T} + \frac{2L_*\eta}{\alpha} + 2\norm{\vSigma}_*\sqrt{\alpha} + \frac2T\SumT c_t^*,
\label{eq:matrix-pre-tuned-rate}
\end{align}
where the second inequality applies \cref{ass:Matrix-bounded-below} ($F(\WB_1)-\E[F(\WB_{T+1})]\le\Delta$) and combines the step-size terms via $\frac{1}{2}+\frac{2\beta}{\alpha}=\frac{\alpha+4(1-\alpha)}{2\alpha}=\frac{4-3\alpha}{2\alpha}\le\frac{2}{\alpha}$.

\paragraph{Step 4: Hyperparameter substitution.}
We substitute \eqref{eq:hyp_muon} into \eqref{eq:matrix-pre-tuned-rate} across three exhaustive cases.
\begin{itemize}
    \item \textbf{Case 1 ($\norm{\vSigma}_*=0$).}
    Here $\alpha=1$, $\beta=0$, $c_t^*=0$ for all $t\ge 1$, and $\eta=\sqrt{\Delta/(L_*T)}$. Substituting into \eqref{eq:matrix-pre-tuned-rate} directly yields
    \[
    \frac1T\SumT\E\norm{\nabla F(\WB_t)}_*
    \le \frac{\Delta}{\eta T} + 2L_*\eta
    = 3\sqrt{\frac{L_*\Delta}{T}}.
    \]

    \item \textbf{Case 2 ($\norm{\vSigma}_*>0$ and $\alpha=1$).}
    Here $\beta=0$, $\eta=\sqrt{\Delta/(L_*T)}$, and \eqref{eq:hyp_muon} implies $\norm{\vSigma}_*\le\sqrt{L_*\Delta/T}$. Since $\beta=0$, we have $c_1^*\le\norm{\vSigma}_*$ and $c_t^*=0$ for all $t\ge 2$, so $\frac2T\SumT c_t^*=\frac{2c_1^*}{T}\le 2\norm{\vSigma}_*$. Substituting into \eqref{eq:matrix-pre-tuned-rate} gives
    \[
    \frac1T\SumT\E\norm{\nabla F(\WB_t)}_*
    \le 3\sqrt{\frac{L_*\Delta}{T}} + 2\norm{\vSigma}_* + 2\norm{\vSigma}_*
    \le 7\sqrt{\frac{L_*\Delta}{T}}.
    \]

    \item \textbf{Case 3 ($\norm{\vSigma}_*>0$ and $0<\alpha<1$).}
    Here $\alpha=\frac{\sqrt{L_*\Delta}}{\norm{\vSigma}_*\sqrt{T}}$ and $\eta=\sqrt{\frac{\Delta\alpha}{L_*T}}$. Define $q_*:=\brac{\frac{\norm{\vSigma}_*^2 L_*\Delta}{T}}^{1/4}$. Direct substitution verifies the three balancing identities
    \begin{equation}\label{eq:muon-balancing-identities}
    \frac{\Delta}{\eta T}
    = \sqrt{\frac{L_*\Delta}{\alpha T}}
    = q_*,
    \qquad
    \frac{L_*\eta}{\alpha}
    = \sqrt{\frac{L_*\Delta}{\alpha T}}
    = q_*,
    \qquad
    \norm{\vSigma}_*\sqrt{\alpha}
    = q_*,
    \end{equation}
    so the first three terms in \eqref{eq:matrix-pre-tuned-rate} sum to $q_*+2q_*+2q_*=5q_*$. For the startup average, applying \cref{lem:upper-geometric-truncation} with $G=G_*=\sqrt{2L_*\Delta}$, $q=L_*\eta$, and $R=\norm{\vSigma}_*$ (for which $H=H_*$) yields
    \[
    \frac2T\SumT c_t^*
    \le \frac{8G_*H_*}{\alpha T} + \frac{4L_*\eta H_*^2}{\alpha^2 T}.
    \]
    Because $T\ge T_{{\rm up},*}\ge \frac{\norm{\vSigma}_*^2}{L_*\Delta}H_*^4$ and $\alpha T=\frac{\sqrt{L_*\Delta T}}{\norm{\vSigma}_*}$, we have
    \[
    \frac{H_*}{\sqrt{\alpha T}}
    = H_*\brac{\frac{\norm{\vSigma}_*^2}{L_*\Delta T}}^{1/4}
    \le 1.
    \]
    Using this bound together with \eqref{eq:muon-balancing-identities} controls the two startup terms as
    \begin{align*}
    \frac{8G_*H_*}{\alpha T} + \frac{4L_*\eta H_*^2}{\alpha^2 T}
    &= 8\sqrt{2}\sqrt{\frac{L_*\Delta}{\alpha T}}\cdot\frac{H_*}{\sqrt{\alpha T}}
       + 4\brac{\frac{L_*\eta}{\alpha}}\brac{\frac{H_*}{\sqrt{\alpha T}}}^2 \le \brac{8\sqrt{2}+4}q_*.
    \end{align*}
    Adding the main terms gives $\frac1T\SumT\E\norm{\nabla F(\WB_t)}_*\le (5+8\sqrt{2}+4)q_*\le 21q_*$.
\end{itemize}
Combining all three cases establishes \eqref{eq:muon-upper-explicit} and hence \cref{thm:muon_upper_infty}.
\end{proof}

\paragraph{Complexity conversion and comparison with~\citet{shen2025convergence}.}
In the comparison of \cref{sec:muon-tight-characterization}, taking traces in \cref{ass:Matrix-variance} gives $\expect{\norm{\Nb_t}_F^2\mid\F_{t-1}}\le\tr(\vSigma^2)=\norm{\vSigma}_F^2$, matching the scalar variance parameter $\sigma^2=\norm{\vSigma}_F^2$ in~\citet{shen2025convergence}. Consider any target accuracy $\epsilon>0$ satisfying
\begin{equation}\label{eq:muon-shen-accuracy}
\epsilon \le \frac{\sqrt{L_*\Delta}}{H_*},
\qquad
\epsilon\sqrt{m}\norm{\vSigma}_F \le L_*\Delta,
\end{equation}
where $H_*$ is defined in \eqref{eq:muon-upper-budget}. Set $B:=\norm{\vSigma}_*^2 L_*\Delta$, choose $C_{\mathrm{Muon}}:=42^4$, and define
\[
T := \left\lceil C_{\mathrm{Muon}}\brac{\frac{L_*\Delta}{\epsilon^2} + \frac{B}{\epsilon^4}}\right\rceil.
\]
Since $T\ge 1$ and the first condition in \eqref{eq:muon-shen-accuracy} implies $\frac{L_*\Delta}{\epsilon^2}\ge H_*^2\ge 1$, we have
\[
T \ge \frac{B}{\epsilon^4}
  = \frac{\norm{\vSigma}_*^2}{L_*\Delta}\brac{\frac{L_*\Delta}{\epsilon^2}}^2
  \ge \frac{\norm{\vSigma}_*^2}{L_*\Delta}H_*^4,
\]
and therefore $T\ge\max\{1,\,\frac{\norm{\vSigma}_*^2}{L_*\Delta}H_*^4\}=T_{{\rm up},*}$. Because $T\ge 42^4 L_*\Delta/\epsilon^2$ and $T\ge 42^4 B/\epsilon^4$, we have $\sqrt{L_*\Delta/T}\le\epsilon/42^2$ and $(B/T)^{1/4}\le\epsilon/42$ (including when $B=0$). Substituting into \eqref{eq:muon-upper-explicit} therefore guarantees
\begin{align*}
\Rt_T^{\mathrm{Muon}}(F)
\le \frac1T\SumT\E\norm{\nabla F(\WB_t)}_*
&\le 21\sqrt{\frac{L_*\Delta}{T}} + 21\brac{\frac{B}{T}}^{1/4} \\
&\le \frac{21}{42^2}\epsilon + \frac{21}{42}\epsilon
 \le \frac{\epsilon}{2} + \frac{\epsilon}{2} = \epsilon,
\end{align*}
which establishes \eqref{eq:muon-complexity-ours}. Similarly, substituting $\sigma^2=\norm{\vSigma}_F^2$ and solving the three rate terms in~\citet[Corollary~4.8]{shen2025convergence} for $T$ yields the sufficient iteration budget
\[
\bigO{\frac{L_*\Delta}{\epsilon^2}
      + \frac{m\norm{\vSigma}_F^2 L_*\Delta}{\epsilon^4}
      + \frac{m^2\norm{\vSigma}_F^4}{L_*\Delta\epsilon^2}}.
\]
By the second condition in \eqref{eq:muon-shen-accuracy}, $\epsilon^2 m\norm{\vSigma}_F^2 \le (L_*\Delta)^2$; multiplying both sides by $\frac{m\norm{\vSigma}_F^2}{L_*\Delta\epsilon^4}\ge 0$ gives $\frac{m^2\norm{\vSigma}_F^4}{L_*\Delta\epsilon^2}\le\frac{m\norm{\vSigma}_F^2 L_*\Delta}{\epsilon^4}$. The third term is therefore dominated by the second, establishing \eqref{eq:muon-complexity-shen}.

\subsection{Lower Bound for SignSGD}
\label{sec:proof-ssgd-lowerbound}

For the fixed-dimension lower bound in \cref{thm:ssgd_lower_infty}, write $\alpha:=1-\beta\in(0,1]$. We first set $\Linf=\Delta=1$ and $\x_1=\vzero$, then restore the general parameters by affine rescaling. The table gives the construction for each fixed $(\eta,\beta)\in(0,\infty)\times[0,1)$ at the stochastic gradient scale $\zeta\asymp(\norm{\vsigma}_1^2/T)^{1/4}$.
\begin{center}
\footnotesize
\setlength{\tabcolsep}{5pt}
\renewcommand{\arraystretch}{1.25}
\begin{tabular}{lll}
\toprule
\textbf{Hyperparameter regime} & \textbf{Hard instance} & \textbf{Lower bound} \\
\midrule
Large step size ($\eta \gtrsim \zeta$) & Lattice quadratic (\cref{slb-lem-steps}, Part~1) & $\Omega(\zeta)$ \\
Small step size ($\eta \le (2\zeta T)^{-1}$) & Truncated ramp (\cref{slb-lem-steps}, Part~2) & $\ge \zeta$ \\
Intermediate $\eta$, $\norm{\vsigma}_1\sqrt{\alpha}\ge 8\zeta$ (weak) & Two-state chain (\cref{slb-lem-lift}) & $\ge \zeta/2$ \\
Intermediate $\eta$, $\norm{\vsigma}_1\sqrt{\alpha}< 8\zeta$ (strong) & 3D rotating gradient (\cref{slb-lem-three}) & $\ge \zeta/2$ \\
\bottomrule
\end{tabular}
\end{center}
The proof of \cref{thm:ssgd_lower_infty} gives the exact thresholds and horizon condition. The two deterministic instances also give the noiseless bound in \cref{slb-cor-zero-noise}. Both use the fact that sign updates lie in $\{-\eta,0,\eta\}$ in each coordinate: $\x_t$ stays on the grid $\eta\mathbb Z^d$, with $\norm{\x_t}_\infty\le(t-1)\eta$.

\begin{lemma}[Deterministic step-size obstructions]\label{slb-lem-steps}
Fix $d\ge 1$, $T\ge 1$, $\eta>0$, and $\beta\in[0,1)$ in the normalized setting ($\Linf=\Delta=1$, $\x_1=\vzero$). There exist two one-dimensional deterministic instances in the normalized class for which, respectively,
\begin{equation}\label{slb-eq-step-bounds}
\min_{t\in[T]}\abs{f'(x_t)} \ge \frac12\min\{\eta,1\},
\qquad
\min_{t\in[T]}\abs{f'(x_t)} = \min\bbrac{\frac{1}{2\eta T},\, 1}.
\end{equation}
In particular, embedding the harder of these two instances into the first coordinate of $\R^d$ gives a deterministic instance satisfying $\min_{t\in[T]}\norm{\nabla f(\x_t)}_1\ge\frac{1}{2\sqrt{T}}$ for every $(\eta,\beta)$.
\end{lemma}

Under \eqref{slb-eq-strong-noise}, one noisy sample can reverse the sign of the momentum. We couple all coordinates with $\sigma_i>0$ to the same scalar two-state chain (\cref{slb-lem-scalar}); coordinates with $\sigma_i=0$ need no noise.

\begin{lemma}[Lifting the two-state obstruction to a general noise profile]\label{slb-lem-lift}
In the normalized setting ($\Linf=\Delta=1$, $\x_1=\vzero$), let $0<\zeta\le 1$, $\eta>0$, $\beta\in[0,1)$, $T\ge 1$, and $\alpha:=1-\beta$. Let $\vsigma\in\R_+^d$ have nonempty active set $I_+:=\{i\in[d]:\sigma_i>0\}$. If
\begin{equation}\label{slb-eq-strong-noise}
\norm{\vsigma}_1\sqrt{\alpha} \ge 8\zeta,
\end{equation}
then there exist a normalized $d$-dimensional objective $f:\R^d\to\R$ and an adapted oracle obeying \cref{ass:bounded from below,ass:variance,ass:infty smoothness} with noise profile $\vsigma$ such that
\[
\expect{\min_{t\in[T]}\norm{\nabla f(\x_t)}_1} \ge \frac{\zeta}{2}.
\]
\end{lemma}

When $\norm{\vsigma}_1\sqrt{\alpha}<8\zeta$, the construction in \cref{sec:signsgd-lower} uses a single noisy initial query instead. With probability $1/2$ it sets $\m_1=b\q_0$ along the longitudinal axis. Along this trajectory the transverse component $\v_t$ of the gradient $b\q_0+\v_t$ rotates in $\spann\{\q_1,\q_2\}$, so its contributions to the momentum largely cancel.

\begin{lemma}[Three-dimensional rotating-gradient instance]\label{slb-lem-three}
In the normalized setting ($\Linf=\Delta=1$, $\x_1=\vzero$) with $d\ge 3$, let $\alpha:=1-\beta\in(0,1]$ and suppose $\vsigma\in\R_+^d$ has second-largest entry $\sigma_{(2)}>0$. Set $C_{\mathrm{rot}}:=1152$ and $c:=\frac{1}{512}$. If $\zeta>0$ satisfies
\begin{equation}\label{slb-eq-three-conditions}
\zeta^2\le c,
\qquad \alpha\zeta^2T\le c,
\qquad C_{\mathrm{rot}}\zeta\alpha\le\eta<c\zeta,
\qquad \frac{\zeta^2}{2}\le\sigma_{(2)}^2,
\end{equation}
then there exist a differentiable objective $f:\R^d\to\R$ and an adapted oracle satisfying \cref{ass:bounded from below,ass:variance,ass:infty smoothness} such that
\[
\expect{\min_{t\in[T]}\norm{\nabla f(\x_t)}_1} \ge \frac{\zeta}{2}.
\]
\end{lemma}

\begin{proof}[Proof of \cref{thm:ssgd_lower_infty}]
With $C_{\mathrm{rot}}=1152$ and $c=\frac{1}{512}$ from \cref{slb-lem-three}, set the scaling constants
\begin{equation}\label{slb-eq-constants}
\delta := \frac{1}{32},
\qquad
c_0 := \frac{c\delta}{4}.
\end{equation}
Here $\delta=\frac{1}{32}$ is chosen small enough that $64\delta^4<c$ and $64C_{\mathrm{rot}}\delta^4<\frac12$, which bridges the weak- and strong-momentum regimes in Case~4 below. Let $\sigma_{(2)}>0$ denote the second-largest entry of $\vsigma\in\R_+^d$. Under the explicit horizon condition
\begin{equation}\label{slb-eq-time}
T \ge T_{\mathrm{low}}
:= \max\bbrac{1,\, \frac{\norm{\vsigma}_1^2}{\Linf\Delta},\, \frac{\delta^4\norm{\vsigma}_1^2{\Linf\Delta}}{4\sigma_{(2)}^4}},
\end{equation}
we will establish the explicit lower bound
\begin{equation}\label{eq:sign-lower-explicit}
\expect{\min_{t\in[T]}\norm{\nabla f(\x_t)}_1}
\ge c_0\sqbrac{\sqrt{\frac{\Linf\Delta}{T}} + \brac{\frac{\norm{\vsigma}_1^2{\Linf\Delta}}{T}}^{1/4}}.
\end{equation}

\paragraph{Step 1: Normalized setting ($\Linf=\Delta=1$, $\x_1=\vzero$).}
Set the target stochastic scale
\[
\zeta := \delta\brac{\frac{\norm{\vsigma}_1^2}{T}}^{1/4}.
\]
Under the normalized horizon condition $T\ge\max\{1,\,\norm{\vsigma}_1^2,\,\frac{\delta^4\norm{\vsigma}_1^2}{4\sigma_{(2)}^4}\}$, we have $0<\zeta\le\delta=\frac{1}{32}\le 1$ and $\zeta^2\le\delta^2=\frac{1}{1024}<c$. Fix any $(\eta,\beta)\in(0,\infty)\times[0,1)$ and write $\alpha:=1-\beta$. We partition $(\eta,\beta)$ into four exhaustive cases:
\begin{itemize}
    \item \textbf{Case 1 ($\eta\ge c\zeta$).}
    The lattice quadratic instance of \cref{slb-lem-steps} (Part~1) yields
    \[
    \min_{t\in[T]}\norm{\nabla f(\x_t)}_1
    \ge \frac12\min\{\eta,1\}
    \ge \frac{c\zeta}{2},
    \]
    since $c\zeta\le 1$.

    \item \textbf{Case 2 ($\eta<c\zeta$ and $\eta\le (2\zeta T)^{-1}$).}
    The truncated linear ramp of \cref{slb-lem-steps} (Part~2) yields
    \[
    \min_{t\in[T]}\norm{\nabla f(\x_t)}_1
    = \min\bbrac{\frac{1}{2\eta T},\,1}
    \ge \zeta.
    \]

    \item \textbf{Case 3 ($(2\zeta T)^{-1}<\eta<c\zeta$ and $\norm{\vsigma}_1\sqrt{\alpha}\ge 8\zeta$).}
    Since $0<\zeta\le 1$ and $\sigma_{(2)}>0$ ensures $I_+\ne\varnothing$, \cref{slb-lem-lift} applies directly and yields $\expect{\min_{t\in[T]}\norm{\nabla f(\x_t)}_1}\ge\frac{\zeta}{2}$.

    \item \textbf{Case 4 ($(2\zeta T)^{-1}<\eta<c\zeta$ and $\norm{\vsigma}_1\sqrt{\alpha}<8\zeta$).}
    In this remaining regime, $\alpha<\frac{64\zeta^2}{\norm{\vsigma}_1^2}$. We verify all four conditions of \eqref{slb-eq-three-conditions} in \cref{slb-lem-three}:
    \begin{enumerate}
        \item $\zeta^2\le\delta^2<c$, as noted above;
        \item using $\zeta^4 T = \delta^4\norm{\vsigma}_1^2$ and $\delta=2^{-5}$,
        \[
        \alpha\zeta^2 T < \frac{64\zeta^4 T}{\norm{\vsigma}_1^2} = 64\delta^4 = 2^{-14} < 2^{-9} = c;
        \]
        \item the upper bound $\eta<c\zeta$ holds by the case assumption, while for the lower bound, $64C_{\mathrm{rot}}\delta^4 = 64\cdot 1152\cdot 2^{-20} = \frac{9}{128} < \frac12$ implies
        \[
        C_{\mathrm{rot}}\zeta\alpha
        < \frac{64C_{\mathrm{rot}}\zeta^3}{\norm{\vsigma}_1^2}
        = \frac{64C_{\mathrm{rot}}\delta^4}{\zeta T}
        < \frac{1}{2\zeta T}
        < \eta;
        \]
        \item the third term of the normalized horizon condition gives
        \[
        \frac{\zeta^2}{2}
        = \frac{\delta^2\norm{\vsigma}_1}{2\sqrt{T}}
        \le \sigma_{(2)}^2.
        \]
    \end{enumerate}
    Hence \cref{slb-lem-three} applies and yields $\expect{\min_{t\in[T]}\norm{\nabla f(\x_t)}_1}\ge\frac{\zeta}{2}$.
\end{itemize}
In all four cases, we obtain an admissible normalized instance with expected minimum gradient norm at least $\frac{c\zeta}{2}$. Independently, the final statement of \cref{slb-lem-steps} provides a deterministic normalized instance with minimum gradient norm at least $\frac{1}{2\sqrt{T}}$. Selecting the harder of these two complete instances and using $\max\{a,b\}\ge\frac{a+b}{2}$ together with $c_0=\frac{c\delta}{4}<\frac14$ gives
\begin{equation}\label{slb-eq-normalized-result}
\expect{\min_{t\in[T]}\norm{\nabla f(\x_t)}_1}
\ge \frac{1}{4\sqrt{T}} + \frac{c\delta}{4}\brac{\frac{\norm{\vsigma}_1^2}{T}}^{1/4}
\ge c_0\sqbrac{\frac{1}{\sqrt{T}} + \brac{\frac{\norm{\vsigma}_1^2}{T}}^{1/4}}.
\end{equation}

\paragraph{Step 2: Scale restoration and initial-point shift.}
To restore general parameters $\Linf,\Delta>0$, $\x_1\in\R^d$, and noise profile $\vsigma\in\R_+^d$, apply Step~1 to the normalized parameters
\[
\bar\vsigma := \frac{\vsigma}{\sqrt{\Linf\Delta}},
\qquad
\bar\eta := \eta\sqrt{\frac{\Linf}{\Delta}},
\]
obtaining a normalized objective $\bar f$ and oracle $\bar\O$ initialized at $\bar\x_1=\vzero$. Define the rescaled objective and oracle by
\begin{equation}\label{slb-eq-rescaling}
f(\x) := \Delta\,\bar f\brac{\sqrt{\frac{\Linf}{\Delta}}(\x-\x_1)},
\qquad
\g_t := \sqrt{\Linf\Delta}\,\bar\g_t.
\end{equation}
By the chain rule, $\nabla f(\x)=\sqrt{\Linf\Delta}\,\nabla\bar f(\bar\x)$ where $\bar\x:=\sqrt{\Linf/\Delta}(\x-\x_1)$. Consequently:
\begin{itemize}
    \item \emph{Trajectory coupling.} Because $\g_t$ and hence the momentum state $\m_t$ scale by the positive constant $\sqrt{\Linf\Delta}$, every coordinate sign $\sign{\m_t}=\sign{\bar\m_t}$ is unchanged, so $\bar\x_t=\sqrt{\Linf/\Delta}(\x_t-\x_1)$ holds pathwise for all $t\in[T]$.
    \item \emph{Class admissibility.} The objective $f$ has initial gap $f(\x_1)-\inf f=\Delta(\bar f(\vzero)-\inf\bar f)\le\Delta$ and $\ell_\infty$-smoothness constant $\Linf$, while the rescaled oracle is conditionally unbiased with coordinate variances at most $\Linf\Delta\,\bar\sigma_i^2=\sigma_i^2$.
    \item \emph{Horizon and risk scaling.} Substituting $\bar\vsigma=\vsigma/\sqrt{\Linf\Delta}$ into the normalized horizon requirement recovers \eqref{slb-eq-time} exactly, and multiplying \eqref{slb-eq-normalized-result} by $\sqrt{\Linf\Delta}$ yields \eqref{eq:sign-lower-explicit}, completing the proof of \cref{thm:ssgd_lower_infty}.
\end{itemize}
\end{proof}

\begin{corollary}[Deterministic vector endpoint]\label{slb-cor-zero-noise}
For all integers $d,T\ge 1$, $\Linf,\Delta>0$, $\x_1\in\R^d$, and fixed $\eta>0$, $\beta\in[0,1)$, there exists a deterministic instance satisfying \cref{ass:bounded from below,ass:infty smoothness} such that
\[
\expect{\min_{t\in[T]}\norm{\nabla f(\x_t)}_1} \ge \frac12\sqrt{\frac{\Linf\Delta}{T}}.
\]
\end{corollary}

\begin{proof}
Apply the final statement of \cref{slb-lem-steps} in the normalized setting (obtaining $\min_{t\in[T]}\norm{\nabla\bar f(\bar\x_t)}_1\ge\frac{1}{2\sqrt{T}}$) and restore general $(\Linf,\Delta,\x_1)$ via the rescaling \eqref{slb-eq-rescaling}. Multiplying by $\sqrt{\Linf\Delta}$ yields the stated bound for every integer $T\ge 1$.
\end{proof}

\subsection{Lower Bound and Tight Complexity for Muon}
\label{sec:proof-muon-lowerbound}

On a rectangular diagonal matrix, the compact matrix sign acts coordinatewise on the diagonal. We orient the embedding along the eigenvectors of the prescribed noise matrix $\vSigma$, so its eigenvalues give the noise profile of the vector instance.

Fix the row-isometry $\PB := [\Ib_m,\,\vzero_{m\times(n-m)}]\in\R^{m\times n}$ (with $\PB=\Ib_m$ when $m=n$), so that $\PB\PB^\top=\Ib_m$. Write the eigendecomposition of $\vSigma\in\SBB_+^m$ as
\begin{equation}\label{eq:matrix-noise-eigendecomposition}
\vSigma = \QB^\top\Diag{\vsigma}\QB,
\qquad
\sigma_1 \ge \cdots \ge \sigma_m \ge 0,
\end{equation}
where $\QB\in\R^{m\times m}$ is orthogonal ($\QB^\top\QB=\QB\QB^\top=\Ib_m$) and $\vsigma=(\sigma_1,\ldots,\sigma_m)^\top\in\R_+^m$. Given any differentiable vector objective $f:\R^m\to\R$ with initial point $\x_1\in\R^m$, smoothness $\Linf=L_*$, initial gap $\Delta$, and adapted stochastic gradient oracle with noise profile $\vsigma$, define the affine projection $\A:\R^{m\times n}\to\R^m$ and lifted matrix objective $F:\R^{m\times n}\to\R$ by
\begin{equation}\label{eq:matrix-affine-map}
\A(\WB) := \x_1 + \Diagvec{\QB(\WB-\WB_1)\PB^\top},
\qquad
F(\WB) := f(\A(\WB)).
\end{equation}
\begin{lemma}[Diagonal embedding of SignSGD into Muon]
\label{lem:matrix-diagonal-embedding}
For any $1\le m\le n$, the lift~\eqref{eq:matrix-affine-map} preserves the initial objective gap, maps $\ell_\infty$-smoothness (\cref{ass:infty smoothness}) with constant $\Linf=L_*$ to operator-norm smoothness (\cref{ass:operator norm smooth}) with constant $L_*$, and satisfies
\begin{equation}\label{eq:matrix-lift-norm-identities}
\nabla F(\WB)=\QB^\top\Diag{\nabla f(\A(\WB))}\PB,
\qquad
\norm{\nabla F(\WB)}_*=\norm{\nabla f(\A(\WB))}_1.
\end{equation}
Moreover, lifting the vector oracle via $\Gb_t:=\QB^\top\Diag{\g_t}\PB$ satisfies \cref{ass:Matrix-variance} and couples the trajectories of \MuonRef{} and SignSGD with momentum (under the same $\eta,\beta$ and initializations $\MB_1=\Gb_1$, $\m_1=\g_1$) pathwise:
\begin{equation}\label{eq:matrix-state-identities}
\A(\WB_t)=\x_t,
\qquad
\MB_t=\QB^\top\Diag{\m_t}\PB
\qquad (\forall\, t\ge 1).
\end{equation}
\end{lemma}

\begin{proof}[Proof of \cref{thm:muon_lower_infty}]
Consider the spectral decomposition \eqref{eq:matrix-noise-eigendecomposition} of $\vSigma\in\SBB_+^m$. Because $\rank(\vSigma)\ge 2$, the eigenvalue vector $\vsigma=(\sigma_1,\ldots,\sigma_m)^\top\in\R_+^m$ has at least two strictly positive entries and satisfies
\[
\norm{\vsigma}_1 = \sum_{i=1}^m\sigma_i = \tr(\vSigma) = \norm{\vSigma}_*,
\qquad
\sigma_{(2)} = \sigma_2 = \lambda_{(2)}(\vSigma) > 0,
\]
where $\lambda_{(2)}(\vSigma)$ denotes the second-largest eigenvalue of $\vSigma$. Recall the universal constants $\delta=\frac{1}{32}$ and $c_0=\frac{c\delta}{4}$ (with $c=\frac{1}{512}$) from \eqref{slb-eq-constants}. Substituting $d=m$, $\Linf=L_*$, $\norm{\vsigma}_1=\norm{\vSigma}_*$, and $\sigma_{(2)}=\lambda_{(2)}(\vSigma)$ into the vector horizon condition \eqref{slb-eq-time} gives the explicit matrix horizon requirement
\[
T \ge T_{{\rm low},*}
:= \max\bbrac{1,\, \frac{\norm{\vSigma}_*^2}{L_*\Delta},\, \frac{\delta^4\norm{\vSigma}_*^2 L_*\Delta}{4\lambda_{(2)}(\vSigma)^4}}.
\]
Fix any integer $T\ge T_{{\rm low},*}$ and any $(\eta,\beta)\in(0,\infty)\times[0,1)$. Since $d=m\ge 3$ and $\vsigma\in\R_+^m$ has at least two positive entries, \cref{thm:ssgd_lower_infty} (with the explicit bound \eqref{eq:sign-lower-explicit}) applied with $\Linf=L_*$, initial gap $\Delta$, initial point $\x_1=\vzero\in\R^m$, and noise profile $\vsigma$ supplies a vector objective $f:\R^m\to\R$ and an adapted stochastic gradient oracle satisfying \cref{ass:bounded from below,ass:variance,ass:infty smoothness} such that
\[
\expect{\min_{t\in[T]}\norm{\nabla f(\x_t)}_1}
\ge c_0\sqbrac{\sqrt{\frac{L_*\Delta}{T}} + \brac{\frac{\norm{\vSigma}_*^2 L_*\Delta}{T}}^{1/4}}.
\]
Lift $(f,\x_1)$ and its oracle to $\R^{m\times n}$ via \eqref{eq:matrix-affine-map} and $\Gb_t=\QB^\top\Diag{\g_t}\PB$. By \cref{lem:matrix-diagonal-embedding}, the resulting matrix objective $F$ and oracle satisfy \cref{ass:Matrix-bounded-below,ass:Matrix-variance,ass:operator norm smooth} with parameters $(L_*,\Delta,\WB_1,\vSigma)$, and the \MuonRef{} iterates $\{\WB_t\}_{t=1}^T$ satisfy $\A(\WB_t)=\x_t$ pathwise. Applying \eqref{eq:matrix-lift-norm-identities} at every $t\in[T]$ yields
\[
\min_{1\le t\le T}\norm{\nabla F(\WB_t)}_*
= \min_{1\le t\le T}\norm{\nabla f(\A(\WB_t))}_1
= \min_{1\le t\le T}\norm{\nabla f(\x_t)}_1.
\]
Taking expectations produces the explicit lower bound
\begin{equation}\label{eq:muon-lower-explicit}
\expect{\min_{1\le t\le T} \norm{\nabla F(\WB_t)}_*}
\ge c_0\sqbrac{\sqrt{\frac{L_*\Delta}{T}} + \brac{\frac{\norm{\vSigma}_*^2 L_*\Delta}{T}}^{1/4}},
\end{equation}
which proves \eqref{eq:lower_muon}.
\end{proof}

\begin{corollary}[Deterministic matrix endpoint]
\label{cor:matrix-muon-zero-noise}
For any $1\leq m\leq n$, $L_*,\Delta>0$, $\WB_1\in\R^{m\times n}$, $T\in\N$, and fixed $\eta>0,\beta\in[0,1)$, when $\vSigma=\vzero$, there exists a deterministic instance satisfying \cref{ass:Matrix-bounded-below,ass:operator norm smooth} such that $\expect{\min_{1\leq t\leq T}\norm{\nabla F(\WB_t)}_*} = \Omega\big(\sqrt{\nicefrac{L_*\Delta}{T}}\big)$.
\end{corollary}

\begin{proof}
Choose $\QB=\Ib_m$ so that \eqref{eq:matrix-noise-eigendecomposition} holds with $\vsigma=\vzero\in\R_+^m$ and $\vSigma=\vzero$. Applying \cref{slb-cor-zero-noise} with $d=m\ge 1$, $\Linf=L_*$, initial gap $\Delta$, initial point $\x_1=\vzero\in\R^m$, and the prescribed $(\eta,\beta,T)$ yields a deterministic vector objective $f:\R^m\to\R$ satisfying \cref{ass:bounded from below,ass:infty smoothness} with exact gradients $\g_t=\nabla f(\x_t)$ and $\min_{t\in[T]}\norm{\nabla f(\x_t)}_1\ge\frac12\sqrt{L_*\Delta/T}$.

Lift $f$ and its exact-gradient oracle to $\R^{m\times n}$ via \eqref{eq:matrix-affine-map} and $\Gb_t=\Diag{\g_t}\PB$. Because \cref{lem:matrix-diagonal-embedding} holds for every $1\le m\le n$ without any rank restriction on $\vSigma$, the lifted objective $F$ satisfies \cref{ass:Matrix-bounded-below,ass:operator norm smooth} with parameters $(L_*,\Delta,\WB_1)$, the lifted oracle returns exact matrix gradients $\Gb_t=\Diag{\nabla f(\A(\WB_t))}\PB=\nabla F(\WB_t)$ (so $\Nb_t=\vzero$ and \cref{ass:Matrix-variance} holds with $\vSigma=\vzero$), and the \MuonRef{} trajectory is coupled pathwise via $\A(\WB_t)=\x_t$. By \eqref{eq:matrix-lift-norm-identities},
\[
\expect{\min_{1\le t\le T}\norm{\nabla F(\WB_t)}_*}
= \min_{1\le t\le T}\norm{\nabla f(\x_t)}_1
\ge \frac12\sqrt{\frac{L_*\Delta}{T}},
\]
which proves \cref{cor:matrix-muon-zero-noise}.
\end{proof}

\begin{proof}[Proof of \cref{thm:muon-tight-complexity}]
Fix $3\le m\le n$, $L_*,\Delta>0$, and $\vSigma\in\SBB_+^m$ with either $\vSigma=\vzero$ or $\rank(\vSigma)\ge 2$.
\begin{itemize}
    \item \textbf{Noiseless case ($\vSigma=\vzero$).}
    Assume $0<\epsilon\le\frac12\sqrt{L_*\Delta}$, so $\frac{L_*\Delta}{\epsilon^2}\ge 4$. For the upper bound, Case~1 of the proof of \cref{thm:muon_upper_infty} uses $\beta=0$ and $\eta=\sqrt{\Delta/(L_*T)}$ to guarantee $\Rt_T^{\mathrm{Muon}}(F)\le 3\sqrt{L_*\Delta/T}$ for every $T\ge 1$. Hence $T=\lceil 9L_*\Delta/\epsilon^2\rceil\le 10\frac{L_*\Delta}{\epsilon^2}$ achieves $\sup_{(F,\O)}\Rt_T^{\mathrm{Muon}}(F)\le\epsilon$. For the lower bound, \cref{cor:matrix-muon-zero-noise} gives $\sup_{(F,\O)}\Rt_T^{\mathrm{Muon}}(F)\ge\frac12\sqrt{L_*\Delta/T}$ for every $T\ge 1$ and every fixed $(\eta,\beta)\in(0,\infty)\times[0,1)$, so achieving $\le\epsilon$ requires $T\ge\frac{L_*\Delta}{4\epsilon^2}$.

    \item \textbf{Noisy case ($\rank(\vSigma)\ge 2$).}
    Here $\norm{\vSigma}_*>0$ and $\lambda_{(2)}(\vSigma)>0$. Assume $0<\epsilon\le\min\bbrac{\frac{\sqrt{L_*\Delta}}{H_*},\,\frac12\sqrt{\frac{L_*\Delta}{T_{{\rm low},*}}}}$ (with $H_*$ from \eqref{eq:muon-upper-budget}). For the upper bound, $\frac{L_*\Delta}{\epsilon^2}\ge H_*^2\ge 1$, and as verified in the complexity conversion of \cref{sec:proof-muon-upperbound}, choosing
    \[
    T = \left\lceil 42^4\brac{\frac{L_*\Delta}{\epsilon^2}+\frac{\norm{\vSigma}_*^2 L_*\Delta}{\epsilon^4}}\right\rceil
      \le (42^4+1)\brac{\frac{L_*\Delta}{\epsilon^2}+\frac{\norm{\vSigma}_*^2 L_*\Delta}{\epsilon^4}}
    \]
    satisfies $T\ge T_{{\rm up},*}$ and yields $\sup_{(F,\O)}\Rt_T^{\mathrm{Muon}}(F)\le\epsilon$ under the horizon-dependent tuning \eqref{eq:hyp_muon}.

    For the lower bound, let $T\ge 1$ and $(\eta,\beta)\in(0,\infty)\times[0,1)$ satisfy $\sup_{(F,\O)}\Rt_T^{\mathrm{Muon}}(F)\le\epsilon$. Because deterministic instances satisfy \cref{ass:Matrix-variance} for any $\vSigma\in\SBB_+^m$, \cref{cor:matrix-muon-zero-noise} implies $\frac12\sqrt{L_*\Delta/T}\le\epsilon$, so $T\ge\frac{L_*\Delta}{4\epsilon^2}\ge T_{{\rm low},*}$. Applying \eqref{eq:muon-lower-explicit} at this horizon gives
    \[
    \epsilon
    \ge c_0\sqbrac{\sqrt{\frac{L_*\Delta}{T}} + \brac{\frac{\norm{\vSigma}_*^2 L_*\Delta}{T}}^{1/4}},
    \]
    which forces both $T\ge c_0^2\frac{L_*\Delta}{\epsilon^2}$ and $T\ge c_0^4\frac{\norm{\vSigma}_*^2 L_*\Delta}{\epsilon^4}$. Since $0<c_0<1$, averaging these two bounds gives $T\ge\frac{c_0^4}{2}\brac{\frac{L_*\Delta}{\epsilon^2}+\frac{\norm{\vSigma}_*^2 L_*\Delta}{\epsilon^4}}$ for every fixed $(\eta,\beta)$.
\end{itemize}
Combining the horizon-tuned upper bound with the uniform lower bound proves \eqref{eq:muon-tight-ours-thm}.
\end{proof}

\subsection{Proofs of Supporting Lemmas}
\label{app:muon-supporting-lemmas}

\paragraph{Supporting lemmas for \cref{sec:proof-muon-upperbound}.}

\begin{proof}[Proof of \cref{lem:gradient-taylor-bound}]
Set $\s:=\v-\u$. By the fundamental theorem of calculus along the segment $t\mapsto\u+t\s$ ($t\in[0,1]$),
\begin{align*}
\abs{h(\v)-h(\u)-\inner{\nabla h(\u)}{\s}}
&= \abs{\int_0^1 \inner{\nabla h(\u+t\s)-\nabla h(\u)}{\s}\ud t} \\
&\le \int_0^1 \norm{\nabla h(\u+t\s)-\nabla h(\u)}_*\norm{\s}\ud t \\
&\le \int_0^1 L\norm{t\s}\norm{\s}\ud t
 = L\norm{\s}^2\int_0^1 t\ud t
 = \frac{L}{2}\norm{\s}^2,
\end{align*}
where the first inequality uses norm duality ($\abs{\inner{\a}{\b}}\le\norm{\a}_*\norm{\b}$) and the second uses the Lipschitz condition.
\end{proof}

\begin{proof}[Proof of \cref{lem:upper-steepest-truncation}]
Define $E:=\norm{\g}_*-\inner{\g}{\s(\g+\bta)}$. Because $\norm{\s(\g+\bta)}\le 1$, norm duality yields $\inner{\g}{\s(\g+\bta)}\le \norm{\g}_*\norm{\s(\g+\bta)}\le\norm{\g}_*$, which gives $E\ge 0$.

To obtain the first upper bound, write $\norm{\g}_*=\inner{\g}{\s(\g)}$ and apply norm duality together with the triangle inequality:
\begin{align*}
E
&= \inner{\g}{\s(\g)-\s(\g+\bta)} \\
&\le \norm{\g}_*\norm{\s(\g)-\s(\g+\bta)} \\
&\le \norm{\g}_*\brac{\norm{\s(\g)}+\norm{\s(\g+\bta)}}
 \le 2\norm{\g}_*.
\end{align*}
To obtain the second upper bound, add and subtract $\inner{\bta}{\s(\g)-\s(\g+\bta)}$:
\begin{align*}
E
&= \inner{\g+\bta}{\s(\g)-\s(\g+\bta)} + \inner{\bta}{\s(\g+\bta)-\s(\g)} \\
&\le \inner{\bta}{\s(\g+\bta)-\s(\g)} \\
&\le \norm{\bta}_*\norm{\s(\g+\bta)-\s(\g)}
 \le 2\norm{\bta}_*,
\end{align*}
where the first inequality holds because $\s(\g+\bta)$ maximizes $\inner{\g+\bta}{\cdot}$ over the unit ball and $\norm{\s(\g)}\le 1$, so $\inner{\g+\bta}{\s(\g)-\s(\g+\bta)}\le 0$. Combining both upper bounds gives $E\le 2\min\{\norm{\g}_*,\norm{\bta}_*\}$.

Finally, when $\bta=\u+\mathbf r$, the triangle inequality $\norm{\u+\mathbf r}_*\le\norm{\u}_*+\norm{\mathbf r}_*$ implies both $\min\{\norm{\g}_*,\norm{\u+\mathbf r}_*\}\le \norm{\g}_*+\norm{\mathbf r}_*$ and $\min\{\norm{\g}_*,\norm{\u+\mathbf r}_*\}\le \norm{\u}_*+\norm{\mathbf r}_*$. Taking the minimum of these two bounds gives $\min\{\norm{\g}_*,\norm{\u+\mathbf r}_*\}\le \min\{\norm{\g}_*,\norm{\u}_*\}+\norm{\mathbf r}_*$, which proves the second claim.
\end{proof}

\begin{proof}[Proof of \cref{lem:upper-geometric-truncation}]
Let $K:=\lceil H/\alpha\rceil$. Bound the terms with $n<K$ by $G+qn$ and those with $n\ge K$ by $R\beta^n$. The latter sum is empty if $T\le K$; in all cases the infinite geometric tail gives
\[
 \sum_{n=0}^{T-1}\min\bbrac{G+qn,\,R\beta^n}
 \le KG+\frac{qK(K-1)}{2}+\frac{R\beta^K}{1-\beta}.
\]
Here $1-\beta=\alpha$, and $\beta\le e^{-\alpha}$ gives $R\beta^K\le Re^{-H}=R/(e+R/G)<G$. Since $H\ge 1$ and $K\le 2H/\alpha$, we obtain
\[
 \sum_{n=0}^{T-1}\min\bbrac{G+qn,\,R\beta^n}
 \le \frac{3GH}{\alpha}+\frac{2qH^2}{\alpha^2},
\]
which implies the stated bound.
\end{proof}

\begin{proof}[Proof of \cref{lem:matrix-filtered-noise}]
Expanding $\mathbf Z\mathbf Z^\top$ separates diagonal and off-diagonal terms:
\[
\expect{\mathbf Z\mathbf Z^\top}
= \sum_{k=1}^t a_k^2\,\expect{\Nb_k\Nb_k^\top}
  + \sum_{1\le j<k\le t} a_j a_k\brac{\expect{\Nb_j\Nb_k^\top}+\expect{\Nb_k\Nb_j^\top}}.
\]
For every pair $j<k$, $\Nb_j$ is $\F_{k-1}$-measurable, so the tower property gives $\expect{\Nb_j\Nb_k^\top}=\expect{\Nb_j\,\expect{\Nb_k^\top\mid\F_{k-1}}}=\vzero$ and likewise $\expect{\Nb_k\Nb_j^\top}=\vzero$. Applying the conditional covariance bound to each diagonal term yields the first inequality:
\[
\expect{\mathbf Z\mathbf Z^\top}
= \sum_{k=1}^t a_k^2\,\expect{\expect{\Nb_k\Nb_k^\top\mid\F_{k-1}}}
\preceq \brac{\sum_{k=1}^t a_k^2}\vSigma^2.
\]
For the second inequality, note that $\norm{\mathbf Z}_* = \tr\big(\sqrt{\mathbf Z\mathbf Z^\top}\big)$. Because the matrix square root $\Ab\mapsto\Ab^{1/2}$ is operator concave and operator monotone on $\SBB_+^m$, operator Jensen's inequality gives $\expect{\sqrt{\mathbf Z\mathbf Z^\top}}\preceq\sqrt{\expect{\mathbf Z\mathbf Z^\top}}$, while operator monotonicity together with $\sqrt{\vSigma^2}=\vSigma$ gives $\sqrt{\expect{\mathbf Z\mathbf Z^\top}}\preceq\sqrt{\sum_{k=1}^t a_k^2}\,\vSigma$. Taking traces preserves the Loewner order, so
\begin{align*}
\E\norm{\mathbf Z}_*
= \tr\brac{\expect{\sqrt{\mathbf Z\mathbf Z^\top}}}
\le \tr\brac{\sqrt{\expect{\mathbf Z\mathbf Z^\top}}}
\le \tr\brac{\sqrt{\sum_{k=1}^t a_k^2}\,\vSigma}
= \norm{\vSigma}_*\sqrt{\sum_{k=1}^t a_k^2}.
\end{align*}
\end{proof}

\paragraph{Supporting lemmas for \cref{sec:proof-ssgd-lowerbound}.}

\begin{proof}[Proof of \cref{slb-lem-steps}]
Since $\sign{m_t}\in\{-1,0,1\}$ at every step independently of $\beta$, the one-dimensional iterates starting from $x_1=0$ satisfy
\[
x_t \in \eta\mathbb Z
\qquad\text{and}\qquad
\abs{x_t} \le (t-1)\eta < \eta T
\qquad (\forall\, t\in[T]).
\]
We construct two deterministic one-dimensional objectives (exact gradients satisfy every variance bound with zero noise).

\paragraph{1. Large step size (lattice quadratic).}
Set $x^\star := \min\{\frac{\eta}{2},\sqrt{2}\}$ and define $f(x):=\frac12(x-x^\star)^2$. Then $f'(x)=x-x^\star$ is $1$-Lipschitz everywhere ($f''\equiv 1$), $f$ is bounded below by $0$, and its initial objective gap satisfies
\[
f(0)-\inf f = \frac{(x^\star)^2}{2} \le \frac{(\sqrt{2})^2}{2} = 1.
\]
We claim that every lattice point $k\eta\in\eta\mathbb Z$ lies at distance at least $x^\star$ from $x^\star$:
\begin{itemize}
    \item If $\frac{\eta}{2}\le\sqrt{2}$, then $x^\star=\frac{\eta}{2}$ is the midpoint between $0$ and $\eta$, so $\min_{k\in\mathbb Z}\abs{k\eta-x^\star}=\frac{\eta}{2}=x^\star$.
    \item If $\frac{\eta}{2}>\sqrt{2}$, then $0<x^\star=\sqrt{2}<\frac{\eta}{2}$, so the nearest lattice point to $x^\star$ is $0$ (with distance $x^\star$), while every nonzero lattice point satisfies $\abs{k\eta-x^\star}\ge \eta-\sqrt{2}>\sqrt{2}=x^\star$.
\end{itemize}
Because $x_t\in\eta\mathbb Z$ for all $t\in[T]$, we obtain
\[
\min_{t\in[T]}\abs{f'(x_t)}
= \min_{t\in[T]}\abs{x_t-x^\star}
\ge x^\star
= \min\bbrac{\frac{\eta}{2},\,\sqrt{2}}
\ge \frac12\min\{\eta,1\}.
\]

\paragraph{2. Small step size (truncated linear ramp).}
Set $h:=\min\{\frac{1}{2\eta T},1\}$ and define the piecewise-quadratic ramp
\[
f(x):=
\begin{cases}
-hx, & x \le \eta T,\\[2pt]
-hx + \frac12(x-\eta T)^2, & \eta T < x < \eta T + h,\\[2pt]
-h\eta T - \frac12 h^2, & x \ge \eta T + h.
\end{cases}
\]
Its derivative is
\[
f'(x)=
\begin{cases}
-h, & x \le \eta T,\\[2pt]
-h + (x-\eta T), & \eta T < x < \eta T + h,\\[2pt]
0, & x \ge \eta T + h,
\end{cases}
\]
which is continuous everywhere and $1$-Lipschitz ($0\le f''(x)\le 1$ on each piece). Moreover, $f$ is nonincreasing and achieves its infimum on $[\eta T+h,\infty)$, so
\[
f(0)-\inf f = h\eta T + \frac12 h^2 \le \frac12 + \frac12 = 1.
\]
Since every iterate satisfies $x_t\le (t-1)\eta < \eta T$, we have $f'(x_t)=-h$ and thus $\abs{f'(x_t)}=h=\min\{\frac{1}{2\eta T},1\}$ for all $t\in[T]$.

\paragraph{Combining both instances.}
To obtain the final claim, use the first instance when $\eta\ge T^{-1/2}$ (giving $\frac12\min\{\eta,1\}\ge\frac{1}{2\sqrt{T}}$) and the second instance when $\eta<T^{-1/2}$ (giving $\min\{\frac{1}{2\eta T},1\}\ge\min\{\frac{1}{2\sqrt{T}},1\}=\frac{1}{2\sqrt{T}}$). Embedding either one-dimensional function into the first coordinate of $\R^d$ while keeping the remaining $d-1$ coordinates constant preserves $\ell_\infty$-smoothness and the initial gap, yielding $\min_{t\in[T]}\norm{\nabla f(\x_t)}_1\ge\frac{1}{2\sqrt{T}}$.
\end{proof}

Before proving \cref{slb-lem-lift}, we establish the one-dimensional two-state stochastic obstruction that serves as its shared building block.

\begin{lemma}[One-dimensional two-state stochastic obstruction]\label{slb-lem-scalar}
Let $0<\zeta\le 1$, $\eta>0$, $\beta\in[0,1)$, $T\ge 1$, and $\sigma>0$, and set $\alpha:=1-\beta$. If
\begin{equation}\label{slb-eq-scalar-noise}
\sigma\sqrt{\alpha} \ge 8\zeta,
\end{equation}
then there exist a normalized one-dimensional objective $p:\R\to\R$ ($\Linf=\Delta=1$, $x_1=0$) and a globally defined, conditionally unbiased oracle with conditional variance at most $\sigma^2$ such that
\[
\expect{\min_{t\in[T]}\abs{p'(x_t)}} \ge \frac{\zeta}{2}.
\]
\end{lemma}

\begin{proof}[Proof of \cref{slb-lem-scalar}]
\paragraph{Construction of the objective and two-state oracle.}
Define the globally lower-bounded function
\begin{equation}\label{slb-eq-scalar-function}
p(x):=
\begin{cases}
-\zeta x, & x \le 0,\\[2pt]
-\zeta x + \frac12 x^2, & 0 < x < \zeta,\\[2pt]
-\frac12\zeta^2, & x \ge \zeta.
\end{cases}
\end{equation}
Its derivative $p'(x)$ equals $-\zeta$ on $(-\infty,0]$, $x-\zeta$ on $(0,\zeta)$, and $0$ on $[\zeta,\infty)$, which is continuous and $1$-Lipschitz. Since $p(0)=0$ and $\inf p=-\frac12\zeta^2$, the initial objective gap satisfies $p(0)-\inf p=\frac12\zeta^2\le 1$. Because $\abs{p'(x)}=\zeta$ for all $x\le 0$, it suffices to keep all queried iterates on the half-line $x\le 0$ with probability at least $1/2$.

To drive the iterates leftward into $x\le 0$ even though $p'(x)=-\zeta<0$ pushes exact-gradient updates rightward, we design a conditional oracle that keeps the momentum buffer $m_t$ inside the two-state set $\{+\zeta,-7\zeta\}$ while $x_t\le 0$:
\begin{itemize}
    \item At $t=1$ (with $x_1=0$), the oracle outputs $g_1=+\zeta$ with probability $\frac34$ and $g_1=-7\zeta$ with probability $\frac14$, so that $m_1=g_1\in\{+\zeta,-7\zeta\}$.
    \item At any $t\ge 2$ on a history satisfying $x_t\le 0$ and $m_{t-1}\in\{+\zeta,-7\zeta\}$, define the transition probabilities
    \begin{equation}\label{slb-eq-transition-probabilities}
    q := \frac{\alpha}{4} \in \left(0,\frac14\right],
    \qquad
    r := \frac{3\alpha}{4} \in \left(0,\frac34\right].
    \end{equation}
    The oracle draws the next momentum state $m_t\in\{+\zeta,-7\zeta\}$ according to the Markov transition probabilities $\Pr(m_t=-7\zeta\mid m_{t-1}=+\zeta)=q$ and $\Pr(m_t=+\zeta\mid m_{t-1}=-7\zeta)=r$, and returns the recovered stochastic gradient
    \begin{equation}\label{slb-eq-recovered-sample}
    g_t := \frac{m_t - \beta m_{t-1}}{\alpha},
    \end{equation}
    so that the EMA update $\beta m_{t-1}+\alpha g_t$ reproduces $m_t$ identically.
    \item On every other history (including whenever $x_t>0$ or off-path queries at $t=1$), the oracle defaults to the exact derivative $g_t=p'(x_t)$.
\end{itemize}
This specifies the conditional oracle globally on every history. For clarity, the four transitions on active histories ($t\ge 2$, $x_t\le 0$) produce the following samples:
\begin{center}
\small
\renewcommand{\arraystretch}{1.2}
\begin{tabular}{cccc}
\toprule
$m_{t-1}$ & $m_t$ & Conditional probability & Returned gradient $g_t$ \\
\midrule
$+\zeta$ & $+\zeta$ & $1-q$ & $+\zeta$ \\
$+\zeta$ & $-7\zeta$ & $q$ & $-\dfrac{(7+\beta)\zeta}{\alpha}$ \\[6pt]
$-7\zeta$ & $-7\zeta$ & $1-r$ & $-7\zeta$ \\
$-7\zeta$ & $+\zeta$ & $r$ & $+\dfrac{(1+7\beta)\zeta}{\alpha}$ \\
\bottomrule
\end{tabular}
\end{center}

\paragraph{Verification of unbiasedness and conditional variance.}
At $t=1$, the sample mean and variance are
\[
\expect{g_1} = \frac34(\zeta) + \frac14(-7\zeta) = -\zeta = p'(0),
\qquad
\operatorname{Var}(g_1) = \frac34(2\zeta)^2 + \frac14(-6\zeta)^2 = 12\zeta^2.
\]
For $t\ge 2$ on any active history, direct computation using $q=\alpha/4$, $r=3\alpha/4$, and $\beta=1-\alpha$ gives:
\begin{align*}
\expect{m_t\mid\F_{t-1}, m_{t-1}=+\zeta}
&= \brac{1-\frac{\alpha}{4}}\zeta + \frac{\alpha}{4}(-7\zeta)
 = (1-2\alpha)\zeta
 = \beta(+\zeta) - \alpha\zeta, \\
\expect{m_t\mid\F_{t-1}, m_{t-1}=-7\zeta}
&= \brac{1-\frac{3\alpha}{4}}(-7\zeta) + \frac{3\alpha}{4}(+\zeta)
 = (-7+6\alpha)\zeta
 = \beta(-7\zeta) - \alpha\zeta.
\end{align*}
Hence $\expect{m_t\mid\F_{t-1}}=\beta m_{t-1}-\alpha\zeta$ in both states, and \eqref{slb-eq-recovered-sample} yields $\expect{g_t\mid\F_{t-1}}=-\zeta=p'(x_t)$. Because the two possible values of $m_t$ differ by $8\zeta$, the two possible values of $g_t$ differ by $\frac{8\zeta}{\alpha}$. A two-point distribution with probabilities $(p_0,1-p_0)$ and support gap $\frac{8\zeta}{\alpha}$ has variance $p_0(1-p_0)\frac{64\zeta^2}{\alpha^2}$, so the two active conditional variances are
\begin{align}
\operatorname{Var}(g_t\mid\F_{t-1}, m_{t-1}=+\zeta)
&= q(1-q)\frac{64\zeta^2}{\alpha^2}
 = \brac{\frac{16}{\alpha}-4}\zeta^2, \nonumber\\
\operatorname{Var}(g_t\mid\F_{t-1}, m_{t-1}=-7\zeta)
&= r(1-r)\frac{64\zeta^2}{\alpha^2}
 = \brac{\frac{48}{\alpha}-36}\zeta^2.
\label{slb-eq-scalar-variance}
\end{align}
Since $\alpha\in(0,1]$, all three active variances ($12\zeta^2$, $(\frac{16}{\alpha}-4)\zeta^2$, and $(\frac{48}{\alpha}-36)\zeta^2$) are bounded above by $\frac{64\zeta^2}{\alpha}\le\sigma^2$ by \eqref{slb-eq-scalar-noise} (this also holds when $\beta=0$, i.e., $\alpha=1$). Every history outside the active stochastic branch returns the exact gradient and has zero variance.

\paragraph{Trajectory analysis via a harmonic exit bound.}
If $T=1$, then $x_1=0$ deterministically and $\abs{p'(x_1)}=\zeta\ge\zeta/2$. Assume now $T\ge 2$, and consider the abstract two-state chain with position counter
\[
K_0 := 0,
\qquad
K_n := -\sum_{t=1}^n \sign{m_t} \in \mathbb Z \qquad (n\ge 1).
\]
As long as $K_j\le0$ for all $j<n$, the first $n$ queries stay on the nonpositive half-line. The oracle then uses the chain's transitions and $x_{n+1}=\eta K_n$. Each update changes the counter by one, and only a negative momentum can take it from $0$ to $1$. The first exit state is therefore $(1,-)$.

Define the drift ratio and harmonic candidates
\[
\rho := \frac{1-q}{1-r} = \frac{4-\alpha}{4-3\alpha} = \frac{3+\beta}{1+3\beta} > 1,
\qquad
h_+(k) := \frac{\rho^k}{3},
\qquad
h_-(k) := \rho^{k-1},
\]
where $\rho>1$ holds since $(3+\beta)-(1+3\beta)=2\alpha>0$. Using $r=3q$ and $(1-r)\rho=1-q$, direct substitution verifies that for every integer $k\le 0$,
\begin{align}
(1-q)h_+(k-1) + qh_-(k+1)
&= \rho^k\brac{\frac{1-q}{3\rho} + q}
 = \rho^k\brac{\frac{1-r}{3} + q} \nonumber\\
&= \rho^k\brac{\frac{1-3q}{3} + q}
 = h_+(k), \nonumber\\
rh_+(k-1) + (1-r)h_-(k+1)
&= \rho^{k-1}\brac{\frac{r}{3} + (1-r)\rho}
 = \rho^{k-1}\brac{q + (1-q)}
 = h_-(k),
\label{slb-eq-harmonic}
\end{align}
together with the boundary value $h_-(1)=\rho^0=1$.

On the abstract chain let $S_n\in\{+,-\}$ be the sign of $m_n$, and stop at the first exit:
\[
 \tau:=\inf\{n\ge1:K_n=1\},\qquad
 H_n:=h_{S_{n\wedge\tau}}(K_{n\wedge\tau})\quad(n\ge1).
\]
Before exit, $K_n\le0$ and both harmonic functions take values in $[0,1]$. At exit the state is $(1,-)$, so $H_n$ stays at $1$. By \eqref{slb-eq-harmonic}, $H_n$ is a bounded martingale for the chain's natural filtration. The first positive query is $x_{\tau+1}$ if $\tau<\infty$, since $K_n$ counts completed updates. The first update gives $(-1,+)$ with probability $3/4$ and $(1,-)$ with probability $1/4$, so
\begin{align*}
 \Pr(\tau\le T-1)
 \le \E H_{T-1}=\E H_1=\frac34 h_+(-1)+\frac14 h_-(1)
   =\frac{1}{4\rho}+\frac14<\frac12.
\end{align*}
On $\{\tau>T-1\}$, the coupled queries all lie in $(-\infty,0]$. Thus $\abs{p'(x_t)}=\zeta$ for every $t\in[T]$ on the same event, and $\expect{\min_{t\in[T]}\abs{p'(x_t)}}\ge\zeta/2$.
\end{proof}

\begin{proof}[Proof of \cref{slb-lem-lift}]
Using the scalar objective $p$ in \eqref{slb-eq-scalar-function}, define
\begin{equation}\label{slb-eq-lift-function}
f(\x) := \sum_{i\in I_+} \frac{\sigma_i}{\norm{\vsigma}_1} p(x_i).
\end{equation}
\paragraph{Shared-chain oracle.}
Couple all active coordinates $i\in I_+$ to a single copy of the scalar two-state Markov chain from \cref{slb-lem-scalar} (applied with scalar noise budget $\sigma=\norm{\vsigma}_1$, which satisfies \eqref{slb-eq-scalar-noise} by \eqref{slb-eq-strong-noise}). Specifically, let $g_t^{\mathrm{sc}}$ denote the scalar oracle output from \cref{slb-lem-scalar}:
\begin{itemize}
    \item At $t=1$ (with $\x_1=\vzero$), draw $g_1^{\mathrm{sc}}$ once and return $g_{1,i}=\frac{\sigma_i}{\norm{\vsigma}_1}g_1^{\mathrm{sc}}$ for $i\in I_+$ and $g_{1,i}=0$ for $i\notin I_+$.
    \item At $t\ge 2$, if all active coordinates share a common nonpositive value $x_{t,i}=x_t^{\mathrm{sc}}\le 0$ and a common rescaled momentum $m_{t-1,i}=\frac{\sigma_i}{\norm{\vsigma}_1}s$ for some $s\in\{+\zeta,-7\zeta\}$ (while $x_{t,i}=m_{t-1,i}=0$ for $i\notin I_+$), draw a single scalar transition $g_t^{\mathrm{sc}}$ from state $s$ and return $g_{t,i}=\frac{\sigma_i}{\norm{\vsigma}_1}g_t^{\mathrm{sc}}$ for $i\in I_+$ and $g_{t,i}=0$ for $i\notin I_+$.
    \item On every other history, return the exact gradient $\nabla f(\x_t)$.
\end{itemize}

\paragraph{Verification of smoothness, initial gap, and oracle bounds.}
Because $p'$ is $1$-Lipschitz and $\sum_{i\in I_+}\frac{\sigma_i}{\norm{\vsigma}_1}=1$, for all $\x,\y\in\R^d$ we have
\[
\norm{\nabla f(\x)-\nabla f(\y)}_1
= \sum_{i\in I_+}\frac{\sigma_i}{\norm{\vsigma}_1}\abs{p'(x_i)-p'(y_i)}
\le \sum_{i\in I_+}\frac{\sigma_i}{\norm{\vsigma}_1}\abs{x_i-y_i}
\le \norm{\x-\y}_\infty,
\]
and the initial gap satisfies $f(\vzero)-\inf f = \sum_{i\in I_+}\frac{\sigma_i}{\norm{\vsigma}_1}(p(0)-\inf p) = \frac{\zeta^2}{2}\le 1$. On every active history, the conditional mean and variance in each active coordinate $i\in I_+$ obey
\begin{align*}
\expect{g_{t,i}\mid\F_{t-1}}
&= \frac{\sigma_i}{\norm{\vsigma}_1}\expect{g_t^{\mathrm{sc}}\mid\F_{t-1}}
 = -\frac{\sigma_i}{\norm{\vsigma}_1}\zeta
 = \nabla_i f(\x_t), \\
\operatorname{Var}(g_{t,i}\mid\F_{t-1})
&= \brac{\frac{\sigma_i}{\norm{\vsigma}_1}}^2\operatorname{Var}(g_t^{\mathrm{sc}}\mid\F_{t-1})
 \le \brac{\frac{\sigma_i}{\norm{\vsigma}_1}}^2\frac{64\zeta^2}{\alpha}
 \le \sigma_i^2,
\end{align*}
where the last inequality uses \eqref{slb-eq-strong-noise}. Inactive coordinates ($i\notin I_+$) and histories outside the active stochastic branch return exact partial derivatives with zero variance.

\paragraph{Coupled trajectory and gradient lower bound.}
Because multiplying by the positive scalar $\frac{\sigma_i}{\norm{\vsigma}_1}>0$ preserves signs ($\sign{m_{t,i}}=\sign{m_t^{\mathrm{sc}}}$), induction over $t$ shows that until the first exit of the shared scalar chain from $\{x\le 0\}$, all active coordinates remain synchronized ($x_{t,i}=x_t^{\mathrm{sc}}\le 0$ and $m_{t,i}=\frac{\sigma_i}{\norm{\vsigma}_1}m_t^{\mathrm{sc}}$ for $i\in I_+$), while all inactive coordinates remain fixed at $0$. On the single non-exit event of \cref{slb-lem-scalar} (which has probability at least $\frac12$), every query $t\in[T]$ satisfies
\[
\norm{\nabla f(\x_t)}_1
= \sum_{i\in I_+}\frac{\sigma_i}{\norm{\vsigma}_1}\abs{p'(x_t^{\mathrm{sc}})}
= \sum_{i\in I_+}\frac{\sigma_i}{\norm{\vsigma}_1}\zeta
= \zeta.
\]
Multiplying by the event probability $\ge\frac12$ yields $\expect{\min_{t\in[T]}\norm{\nabla f(\x_t)}_1}\ge\frac{\zeta}{2}$.
\end{proof}

\begin{proof}[Proof of \cref{slb-lem-three}]
Select two coordinates $i_1,i_2\in[d]$ with the largest noise budgets (so $\sigma_{i_1},\sigma_{i_2}\ge\sigma_{(2)}>0$) and any third distinct coordinate $i_3\in[d]\setminus\{i_1,i_2\}$ (possibly with $\sigma_{i_3}=0$). Write $\u:=(x_{i_1},x_{i_2},x_{i_3})^\top\in\R^3$ and introduce the orthonormal basis of $\R^3$
\[
\q_0 := \frac{1}{\sqrt3}(1,1,1)^\top,
\qquad
\q_1 := \frac{1}{\sqrt2}(1,-1,0)^\top,
\qquad
\q_2 := \frac{1}{\sqrt6}(1,1,-2)^\top,
\]
defining longitudinal and transverse coordinates $z:=\inner{\q_0}{\u}$, $y_1:=\inner{\q_1}{\u}$, and $y_2:=\inner{\q_2}{\u}$. (These are analysis coordinates only; SignSGD still applies $\sign{\cdot}$ in the original coordinate basis.)

\paragraph{Construction of the objective and oracle.}
Set the frequency constant $\kappa:=\frac{1}{24}$ (chosen below so that the combined upper bound on the oscillatory and longitudinal Hessian contributions is at most $1$), the spatial frequency $k:=\kappa/\zeta$, and the travel length $D_{\mathrm{rot}}:=\sqrt3\,\eta T$. Define the one-sided smoothed ramp
\begin{equation}\label{slb-eq-ramp}
H_{\mathrm{rot}}(z):=
\begin{cases}
z, & z \ge -D_{\mathrm{rot}},\\[2pt]
z + \dfrac{(z+D_{\mathrm{rot}})^2}{2\zeta}, & -D_{\mathrm{rot}}-\zeta < z < -D_{\mathrm{rot}},\\[4pt]
-D_{\mathrm{rot}} - \dfrac{\zeta}{2}, & z \le -D_{\mathrm{rot}}-\zeta.
\end{cases}
\end{equation}
Direct differentiation confirms three properties of $H_{\mathrm{rot}}$:
\begin{itemize}
    \item $H_{\mathrm{rot}}$ and $H_{\mathrm{rot}}'$ are continuous everywhere, with $H_{\mathrm{rot}}'(z)=1$ for all $z\ge -D_{\mathrm{rot}}$;
    \item $H_{\mathrm{rot}}(z)\ge -D_{\mathrm{rot}}-\frac{\zeta}{2}$ globally, with $H_{\mathrm{rot}}(0)=0$;
    \item $0\le H_{\mathrm{rot}}''(z)\le \frac{1}{\zeta}$ wherever the second derivative exists.
\end{itemize}
For a longitudinal drift parameter $b>0$ chosen below, define $f_3:\R^3\to\R$ by
\begin{equation}\label{slb-eq-three-function}
f_3(\u) := b\,H_{\mathrm{rot}}(z) + \frac{\zeta}{k}\cos(kz)\sin(ky_1) + \frac{\zeta}{k}\sin(kz)\sin(ky_2),
\end{equation}
and embed $f_3$ into $\R^d$ via $f(\x):=f_3(x_{i_1},x_{i_2},x_{i_3})$, leaving the remaining $d-3$ coordinates constant. Along the longitudinal axis $y_1=y_2=0$ with $z\ge -D_{\mathrm{rot}}$, we have $H_{\mathrm{rot}}'(z)=1$, $\sin(ky_1)=\sin(ky_2)=0$, and $\cos(ky_1)=\cos(ky_2)=1$, so the partial derivatives in $(z,y_1,y_2)$ are $(b,\zeta\cos(kz),\zeta\sin(kz))$. Hence the exact gradient along this axis is
\begin{equation}\label{slb-eq-axis-gradient}
\nabla f_3(z\q_0) = b\q_0 + \zeta\sqbrac{\cos(kz)\q_1 + \sin(kz)\q_2}.
\end{equation}

At $t=1$ (where $\u_1=\vzero$), the exact gradient is $\nabla f_3(\vzero)=b\q_0+\zeta\q_1$. The stochastic oracle injects noise only at this initial query, returning
\begin{equation}\label{slb-eq-three-oracle}
\g_1 :=
\begin{cases}
b\q_0, & \text{with probability }\frac12,\\[2pt]
b\q_0 + 2\zeta\q_1, & \text{with probability }\frac12,
\end{cases}
\end{equation}
and returning exact gradients ($\g_t=\nabla f_3(\u_t)$) for all $t\ge 2$ (as well as away from $\u_1=\vzero$ at $t=1$), with zero gradient on all coordinates outside $\{i_1,i_2,i_3\}$.

\paragraph{Verification of unbiasedness and coordinate-wise variance.}
At $t=1$, the noise vector is $\g_1-\nabla f_3(\vzero)=\pm\zeta\q_1=\pm\frac{\zeta}{\sqrt2}(1,-1,0)^\top$ with equal probabilities $\frac12$. It is therefore zero-mean, and its coordinate variances on $(i_1,i_2,i_3)$ are
\[
\brac{\frac{\zeta^2}{2},\,\frac{\zeta^2}{2},\,0}
\le \brac{\sigma_{i_1}^2,\,\sigma_{i_2}^2,\,\sigma_{i_3}^2}
\]
by the last condition in \eqref{slb-eq-three-conditions}. Every query $t\ge 2$ returns exact gradients with zero variance. Thus the oracle satisfies \cref{ass:variance} globally.

\paragraph{Trajectory induction via rotational cancellation.}
Let $I:=(i_1,i_2,i_3)$ denote the active coordinate triple, so $\m_{t,I^c}=\vzero$ for all $t\ge 1$, and let $\mathcal E_{\mathrm{init}}:=\{\g_1=b\q_0\}$, which occurs with probability $\frac12$. On $\mathcal E_{\mathrm{init}}$, the active initial momentum is $\m_{1,I}=\g_{1,I}=b\q_0=\frac{b}{\sqrt3}(1,1,1)^\top$, whose three entries are strictly positive. Define the candidate trajectory along the $\q_0$ axis:
\begin{equation}\label{slb-eq-candidate-path}
\u_j := -(j-1)\eta(1,1,1)^\top = z_j\q_0,
\qquad
z_j := -\sqrt3(j-1)\eta
\qquad (j\in[T]).
\end{equation}
Suppose temporarily that the iterates through time $t\in[T]$ follow \eqref{slb-eq-candidate-path}. Since $z_j\ge -\sqrt3(T-1)\eta > -D_{\mathrm{rot}}$, \eqref{slb-eq-axis-gradient} applies at every $j\le t$, giving $\nabla f_3(\u_j)=b\q_0+\v_j$, where the transverse gradient is
\[
\v_j := \zeta\sqbrac{\cos(kz_j)\q_1 + \sin(kz_j)\q_2},
\qquad
\norm{\v_j}_2 = \zeta.
\]
Because every longitudinal gradient equals $b$ (including $\g_{1,I}=b\q_0$ on $\mathcal E_{\mathrm{init}}$), the longitudinal component of the active EMA $\m_{t,I}$ remains identically $b\q_0$, while the transverse component $\w_t:=\m_{t,I}-b\q_0\in\R^3$ obeys $\w_1=\vzero$ and
\[
\w_t = \alpha\sum_{j=2}^t \beta^{t-j}\v_j \qquad (t\ge 2).
\]
Set the planar rotation angle $\theta:=\sqrt3\,k\eta>0$, and let $\mathbf R_\theta$ denote counterclockwise rotation by $\theta$ in $\spann\{\q_1,\q_2\}$. Because $kz_j = kz_{j-1}-\theta$, we have $\mathbf R_\theta\v_j = \v_{j-1}$ for all $j\ge 2$. Rotating $\w_t$ by $\theta$, multiplying by $\beta$, and subtracting cancels every interior term:
\begin{align*}
\w_t - \beta\mathbf R_\theta\w_t
&= \alpha\sum_{j=2}^t \beta^{t-j}\v_j - \alpha\sum_{j=2}^t \beta^{t-(j-1)}\mathbf R_\theta\v_j \\
&= \alpha\sum_{j=2}^t \beta^{t-j}\v_j - \alpha\sum_{r=1}^{t-1} \beta^{t-r}\v_r
 = \alpha\brac{\v_t - \beta^{t-1}\v_1},
\end{align*}
where $\v_1=\zeta\q_1$ is the true initial transverse gradient. By the triangle inequality,
\[
\norm{\w_t - \beta\mathbf R_\theta\w_t}_2
\le \alpha\brac{\norm{\v_t}_2 + \beta^{t-1}\norm{\v_1}_2}
\le 2\alpha\zeta.
\]
On the other hand, since $\norm{\mathbf R_\theta\w_t}_2=\norm{\w_t}_2$ and $\inner{\w_t}{\mathbf R_\theta\w_t}=(\cos\theta)\norm{\w_t}_2^2$, the law of cosines gives
\[
\norm{\w_t - \beta\mathbf R_\theta\w_t}_2^2
= \brac{1 - 2\beta\cos\theta + \beta^2}\norm{\w_t}_2^2
= \sqbrac{(1-\beta)^2 + 2\beta(1-\cos\theta)}\norm{\w_t}_2^2.
\]
From the step-size condition $C_{\mathrm{rot}}\zeta\alpha\le\eta<c\zeta$ in \eqref{slb-eq-three-conditions}, together with $\kappa=\frac{1}{24}$, $C_{\mathrm{rot}}=1152$, and $c=\frac{1}{512}$, we have $\alpha < \frac{c}{C_{\mathrm{rot}}} < \frac12$ (so $\beta=1-\alpha>\frac12$) and $0<\theta=\sqrt3\,\kappa\frac{\eta}{\zeta}<\sqrt3\,\kappa c < 1$. Because $1-\cos\theta\ge\frac{\theta^2}{4}$ for all $\theta\in[0,1]$, it follows that
\[
\norm{\w_t - \beta\mathbf R_\theta\w_t}_2^2
\ge 2\cdot\frac12\cdot\frac{\theta^2}{4}\norm{\w_t}_2^2
= \frac{\theta^2}{4}\norm{\w_t}_2^2.
\]
Combining the upper and lower bounds on $\norm{\w_t-\beta\mathbf R_\theta\w_t}_2$ yields
\begin{equation}\label{slb-eq-transverse-bound}
\norm{\w_t}_2 \le \frac{4\alpha\zeta}{\theta} = \frac{4\alpha\zeta}{\sqrt3\,k\eta},
\end{equation}
which also holds trivially at $t=1$ since $\w_1=\vzero$.

Now fix the longitudinal drift parameter prior to running the algorithm as
\begin{equation}\label{slb-eq-b-choice}
b := \frac{8\alpha\zeta}{k\eta} = \frac{8\alpha\zeta^2}{\kappa\eta} > 0,
\end{equation}
so that \eqref{slb-eq-transverse-bound} simplifies to $\norm{\w_t}_2\le\frac{b}{2\sqrt3}$. We prove by induction on $t\in[T]$ that on $\mathcal E_{\mathrm{init}}$, the active iterate $\u_t$ obeys \eqref{slb-eq-candidate-path}, $\m_{t,I^c}=\vzero$, and every active momentum entry satisfies $(\m_{t,I})_r\ge\frac{b}{2\sqrt3}>0$ ($r\in\{1,2,3\}$):
\begin{itemize}
    \item \emph{Base case ($t=1$).} At $t=1$, $\u_1=\vzero$ matches \eqref{slb-eq-candidate-path}, $\m_{1,I^c}=\vzero$, and $\m_{1,I}=b\q_0$ has all three entries equal to $\frac{b}{\sqrt3}>\frac{b}{2\sqrt3}>0$.
    \item \emph{Inductive step ($t-1\to t$).} If the claim holds through $t-1$, then $\sign{\m_{j,I}}=(1,1,1)^\top$ for all $j\le t-1$, so $\u_t=\u_{t-1}-\eta(1,1,1)^\top$ satisfies \eqref{slb-eq-candidate-path}. The rotational cancellation bound \eqref{slb-eq-transverse-bound} therefore applies at step $t$, giving $\m_{t,I^c}=\vzero$ and, for every active entry $r\in\{1,2,3\}$,
    \[
    (\m_{t,I})_r = (b\q_0)_r + (\w_t)_r \ge \frac{b}{\sqrt3} - \norm{\w_t}_2 \ge \frac{b}{2\sqrt3} > 0,
    \]
    which closes the induction without assuming any future iterate position.
\end{itemize}

\paragraph{Verification of smoothness, initial gap, and gradient lower bound.}
In the orthonormal coordinates $(z,y_1,y_2)$, differentiating \eqref{slb-eq-three-function} twice wherever $H_{\mathrm{rot}}''$ exists gives
\begin{align*}
\abs{\partial_{zz} f_3}
&= \abs{b H_{\mathrm{rot}}''(z) - \zeta k\cos(kz)\sin(ky_1) - \zeta k\sin(kz)\sin(ky_2)}
 \le \frac{b}{\zeta} + 2\zeta k, \\
\abs{\partial_{y_1y_1} f_3} &\le \zeta k,
\qquad
\abs{\partial_{y_2y_2} f_3} \le \zeta k,
\qquad
\abs{\partial_{zy_1} f_3} \le \zeta k,
\qquad
\abs{\partial_{zy_2} f_3} \le \zeta k,
\qquad
\partial_{y_1y_2} f_3 = 0.
\end{align*}
Using $\zeta k=\kappa$, the maximum row $\ell_1$-norm of the $3\times 3$ symmetric Hessian in $(z,y_1,y_2)$ is at most $(\frac{b}{\zeta}+2\kappa)+\kappa+\kappa=\frac{b}{\zeta}+4\kappa$. By orthogonal invariance of the spectral norm, $\norm{\nabla^2 f_3(\u)}_{\op}\le\frac{b}{\zeta}+4\kappa$ almost everywhere. For any $\a\in\R^3$, Cauchy--Schwarz and $\norm{\a}_2\le\sqrt3\norm{\a}_\infty$ yield
\[
\norm{\nabla^2 f_3(\u)\,\a}_1
\le \sqrt3\norm{\nabla^2 f_3(\u)\,\a}_2
\le \sqrt3\brac{\frac{b}{\zeta}+4\kappa}\norm{\a}_2
\le 3\brac{\frac{b}{\zeta}+4\kappa}\norm{\a}_\infty.
\]
From \eqref{slb-eq-b-choice}, $\eta\ge C_{\mathrm{rot}}\zeta\alpha$, and our choices $\kappa=\frac{1}{24}$ and $C_{\mathrm{rot}}=1152$, we have
\[
\frac{b}{\zeta} = \frac{8\alpha\zeta}{\kappa\eta} \le \frac{8}{\kappa C_{\mathrm{rot}}} = \frac{8}{48} = \frac16,
\qquad
4\kappa = \frac16,
\qquad
3\brac{\frac{b}{\zeta}+4\kappa} \le 3\brac{\frac16 + \frac16} = 1.
\]
Because $H_{\mathrm{rot}}'$ is continuous across its joining planes, integrating this Hessian bound along line segments establishes global $\ell_\infty$-smoothness (\cref{ass:infty smoothness}) with $\Linf\le 1$.

For the initial objective gap, note that $f_3(\vzero)=0$, $H_{\mathrm{rot}}(z)\ge -D_{\mathrm{rot}}-\frac{\zeta}{2}$, and each oscillatory term has magnitude at most $\frac{\zeta}{k}=\frac{\zeta^2}{\kappa}$. Using $D_{\mathrm{rot}}=\sqrt3\,\eta T$, \eqref{slb-eq-b-choice}, and the first three inequalities in \eqref{slb-eq-three-conditions} gives
\begin{align*}
f_3(\vzero) - \inf f_3
&\le b\brac{D_{\mathrm{rot}} + \frac{\zeta}{2}} + \frac{2\zeta}{k} = \frac{8\sqrt3}{\kappa}\alpha\zeta^2 T + \frac{4\alpha\zeta^3}{\kappa\eta} + \frac{2\zeta^2}{\kappa} \\
&\le \frac{8\sqrt3}{\kappa}c + \brac{\frac{4}{\kappa C_{\mathrm{rot}}} + \frac{2}{\kappa}}c
 = c\brac{\frac{8\sqrt3 + 2}{\kappa} + \frac{4}{\kappa C_{\mathrm{rot}}}}
 < \frac{381}{512} < 1,
\end{align*}
verifying \cref{ass:bounded from below} with $\Delta=1$.

Finally, on the event $\mathcal E_{\mathrm{init}}$ (which has probability $\frac12$), every query $t\in[T]$ lies on the candidate axis with $z_t\ge -D_{\mathrm{rot}}$, so \eqref{slb-eq-axis-gradient} gives
\[
\norm{\nabla f(\x_t)}_1
= \norm{\nabla f_3(\u_t)}_1
\ge \norm{\nabla f_3(\u_t)}_2
= \sqrt{b^2 + \zeta^2}
\ge \zeta.
\]
Multiplying by $\Pr(\mathcal E_{\mathrm{init}})=\frac12$ yields $\expect{\min_{t\in[T]}\norm{\nabla f(\x_t)}_1}\ge\frac{\zeta}{2}$.
\end{proof}

\paragraph{Supporting lemma for \cref{sec:proof-muon-lowerbound}.}

\begin{proof}[Proof of \cref{lem:matrix-diagonal-embedding}]
Let $D(\z):=\QB^\top\Diag{\z}\PB$. Write $\q_i=\QB^\top\e_i$ and $\p_i=\PB^\top\e_i$ for $i\in[m]$. Each family is orthonormal because $\QB\QB^\top=\PB\PB^\top=\Ib_m$.

\paragraph{Four identities for the diagonal lift.}
The adjoint of $D$ extracts the coordinates along $\q_i\p_i^\top$: for any $\XB\in\R^{m\times n}$,
\[
 \inner{D(\z)}{\XB}
 =\sum_{i=1}^m z_i\,\q_i^\top\XB\p_i
 =\inner{\z}{D^*(\XB)},\qquad
 D^*(\XB)=\Diagvec{\QB\XB\PB^\top}.
\]
Orthonormality recovers each coordinate of $\z$, and each bilinear form involves two unit vectors. Hence
\[
 D^*D(\z)=\z,\qquad
 \norm{D^*(\XB)}_\infty=\max_{i\in[m]}\abs{\q_i^\top\XB\p_i}
 \le\norm{\XB}_{\op}.
\]
The same vectors give a compact singular value decomposition, up to ordering the nonzero singular values:
\[
 D(\z)=\sum_{i:\,z_i\ne0}\abs{z_i}\,(\sign{z_i}\q_i)\p_i^\top.
\]
Summing the singular values gives the nuclear norm. Replacing each positive singular value by one gives the compact matrix sign; zero coordinates contribute to neither expression. Thus, also for $\z=\vzero$,
\[
 \norm{D(\z)}_*=\norm{\z}_1,\qquad
 \msign{D(\z)}=D(\sign{\z}).
\]

\paragraph{Objective gap and smoothness.}
By \eqref{eq:matrix-affine-map}, $\A(\WB)=\x_1+D^*(\WB-\WB_1)$. Every $\x$ has a preimage because $D^*D=\Ib_m$:
\[
 \A\bigl(\WB_1+D(\x-\x_1)\bigr)=\x.
\]
Thus $\inf F=\inf f$. Since $\A(\WB_1)=\x_1$, the initial gap satisfies
\[
 F(\WB_1)-\inf F=f(\x_1)-\inf f\le\Delta.
\]
The chain rule gives
\[
 \inner{\nabla F(\WB)}{\XB}
 =\inner{\nabla f(\A(\WB))}{D^*(\XB)}
 =\inner{D(\nabla f(\A(\WB)))}{\XB}.
\]
This gradient formula and the nuclear norm identity prove \eqref{eq:matrix-lift-norm-identities}. By \cref{ass:infty smoothness} and the contraction of $D^*$,
\begin{align*}
 \norm{\nabla F(\WB')-\nabla F(\WB)}_*
 &=\norm{\nabla f(\A(\WB'))-\nabla f(\A(\WB))}_1\\
 &\le L_*\norm{D^*(\WB'-\WB)}_\infty
 \le L_*\norm{\WB'-\WB}_{\op}.
\end{align*}

\paragraph{Conditional oracle bounds.}
At $\WB_t$, query the vector oracle at $\A(\WB_t)$ and return $\Gb_t=D(\g_t)$ with the same revealed internal state. The lifted samples determine $\g_s=D^*(\Gb_s)$ for $s<t$, and conversely each $\g_s$ determines $\Gb_s$. The iterates are functions of these past samples, so the two histories generate the same filtration $\{\F_t\}$. Conditional unbiasedness gives
\[
 \expect{\Gb_t\mid\F_{t-1}}
 =D\brac{\expect{\g_t\mid\F_{t-1}}}
 =D\brac{\nabla f(\A(\WB_t))}=\nabla F(\WB_t).
\]
For $\boldsymbol\nu_t=\g_t-\nabla f(\A(\WB_t))$, the matrix noise is $\Nb_t=D(\boldsymbol\nu_t)$. The identity $\PB\PB^\top=\Ib_m$ eliminates all cross terms, without any independence assumption on the coordinates:
\begin{align*}
 \Nb_t\Nb_t^\top
 &=\QB^\top\Diag{\nu_{t,1}^2,\ldots,\nu_{t,m}^2}\QB,\\
 \expect{\Nb_t\Nb_t^\top\mid\F_{t-1}}
 &\preceq \QB^\top\Diag{\sigma_1^2,\ldots,\sigma_m^2}\QB=\vSigma^2.
\end{align*}
The inequality uses the coordinatewise conditional variance bounds in \cref{ass:variance} and the eigendecomposition \eqref{eq:matrix-noise-eigendecomposition}.

\paragraph{Coupled updates.}
At the first query, $\A(\WB_1)=\x_1$ and $\MB_1=\Gb_1=D(\g_1)=D(\m_1)$. Suppose both identities in \eqref{eq:matrix-state-identities} hold at step $t$. Applying $D^*$ to the \MuonRef{} step and using the compact sign identity gives
\[
 \A(\WB_{t+1})=\x_t-\eta D^*\bigl(D(\sign{\m_t})\bigr)
              =\x_t-\eta\sign{\m_t}=\x_{t+1}.
\]
The next vector sample is therefore drawn at $\x_{t+1}$. Linearity of $D$ gives the momentum identity at the same query:
\[
 \MB_{t+1}=\beta D(\m_t)+(1-\beta)D(\g_{t+1})=D(\m_{t+1}).
\]
This proves the pathwise coupling for every $t$.
\end{proof}

\section{Analysis for Vanilla SGD}
\label{app:direct-matrix-sgd}

We analyze vanilla \SGDRef{} (\cref{alg:matrix-optimizers} with $\beta=0$). Its lower bound uses a column-and-diagonal construction that also applies to \SGDMRef{} in \cref{app:direct-matrix-sgdm}. The supporting proofs are in \cref{app:sgd-supporting-lemmas}.

\paragraph{Preliminaries and all-horizon target rate.}
Throughout \cref{app:direct-matrix-sgd,app:direct-matrix-sgdm}, we use the shorthands
\[
A_0 := L_*\Delta,
\qquad
V := \norm{\vSigma}_F^2,
\qquad
v := \frac{V}{A_0}.
\]
To convert finite-time bounds into iteration complexities (\cref{thm:sgd-tight-complexity}), we must also rule out arbitrarily small iteration budgets $T$. We therefore prove a two-sided bound that holds for every integer $T\ge 1$ with respect to the all-horizon target rate
\begin{equation}\label{eq:appendix-sgd-all-horizon}
\mathcal B_T := \min\bbrac{\sqrt{A_0},\,
\sqrt{\frac{mA_0}{T}} + \brac{\frac{m^2A_0V}{T}}^{1/4}},
\end{equation}
where the $\sqrt{A_0}$ cap reflects the initial gradient norm at $t=1$. Specifically, we prove that there exist universal constants $c,C>0$ such that, for every integer $T\ge 1$:
\begin{itemize}
    \item \textbf{Upper bound (\cref{app:sgd-upper-cap}):} An appropriate fixed step size $\eta>0$ guarantees $\expect{\min_{t\in[T]}\norm{\nabla F(\WB_t)}_*} \le C\,\mathcal B_T$ for every admissible objective $F$ and oracle $\O$.
    \item \textbf{Lower bound (\cref{app:sgd-lower-bound}):} For every fixed step size $\eta>0$, there exist an admissible objective $F$ and oracle $\O$ such that $\expect{\min_{t\in[T]}\norm{\nabla F(\WB_t)}_*} \ge c\,\mathcal B_T$.
\end{itemize}
Whenever $T \ge \lceil\max\{4m, 16m^2v\}\rceil$, we have $\sqrt{mA_0/T}\le \frac12\sqrt{A_0}$ and $(m^2A_0V/T)^{1/4}\le \frac12\sqrt{A_0}$, so $\mathcal B_T = \sqrt{mA_0/T} + (m^2A_0V/T)^{1/4}$. This yields \cref{thm:matrix-sgd-all-horizon} with a horizon threshold independent of $\eta$.

\subsection{Upper Bound for Vanilla SGD}
\label{app:sgd-upper-cap}

We prove \cref{thm:matrix-sgd-upper} and the all-horizon upper bound $\expect{\min_{t\in[T]}\norm{\nabla F(\WB_t)}_*} \le C\,\mathcal B_T$ (with $C=2$).

\paragraph{Step 1: Norm ordering and Frobenius reduction.}
For every matrix $U\in\R^{m\times n}$, the singular-value inequalities $\norm{U}_{\op}\le\norm{U}_F\le\norm{U}_*$ imply that operator-norm smoothness (\cref{ass:operator norm smooth}) induces Frobenius smoothness with the same constant $L_*$: for all $U,U'\in\R^{m\times n}$,
\[
\norm{\nabla F(U)-\nabla F(U')}_F
\le \norm{\nabla F(U)-\nabla F(U')}_*
\le L_*\norm{U-U'}_{\op}
\le L_*\norm{U-U'}_F.
\]
Taking traces in \cref{ass:Matrix-variance} yields the conditional Frobenius variance bound
\[
\E\sqbrac{\norm{\Nb_t}_F^2\mid\F_{t-1}}
= \tr\brac{\E\sqbrac{\Nb_t\Nb_t^\top\mid\F_{t-1}}}
\le \tr(\vSigma^2)
= \norm{\vSigma}_F^2
= V.
\]

\paragraph{Step 2: Quadratic descent and telescoping.}
Write $D_t:=\nabla F(\WB_t)$, so that $\Gb_t=D_t+\Nb_t$ and $\WB_{t+1}=\WB_t-\eta\Gb_t$. Applying \cref{lem:gradient-taylor-bound} under Frobenius $L_*$-smoothness gives
\[
F(\WB_{t+1})
\le F(\WB_t) - \eta\inner{D_t}{D_t+\Nb_t} + \frac{L_*\eta^2}{2}\norm{D_t+\Nb_t}_F^2.
\]
Taking conditional expectations given $\F_{t-1}$ and using conditional unbiasedness ($\E[\Nb_t\mid\F_{t-1}]=\vzero$) eliminates the cross term $\inner{D_t}{\Nb_t}$:
\begin{align*}
\E\sqbrac{F(\WB_{t+1})\mid\F_{t-1}}
&\le F(\WB_t) - \eta\norm{D_t}_F^2
     + \frac{L_*\eta^2}{2}\brac{\norm{D_t}_F^2 + \E\sqbrac{\norm{\Nb_t}_F^2\mid\F_{t-1}}} \\
&\le F(\WB_t) - \eta\brac{1-\frac{L_*\eta}{2}}\norm{D_t}_F^2
     + \frac{L_*\eta^2 V}{2}.
\end{align*}
All expectations are finite by induction, because $L_*$-smooth gradients grow at most linearly, $F$ grows at most quadratically, and $\Nb_t$ has bounded conditional second moment. For any $0<\eta\le 1/L_*$, we have $1-\frac{L_*\eta}{2}\ge\frac12$. Taking total expectations, summing over $t=1,\ldots,T$, dividing by $\eta T/2$, and applying \cref{ass:Matrix-bounded-below} ($F(\WB_1)-\E[F(\WB_{T+1})]\le F(\WB_1)-\inf F\le\Delta$) yields
\begin{equation}\label{eq:direct-sgd-descent}
\frac1T\sum_{t=1}^T\E\norm{D_t}_F^2
\le \frac{2\Delta}{\eta T} + L_*\eta V.
\end{equation}

\paragraph{Step 3: Nuclear-norm conversion and step-size selection.}
Because $D_t\in\R^{m\times n}$ with $m\le n$, we have $\rank(D_t)\le m$, so Cauchy--Schwarz on singular values gives $\norm{D_t}_*\le\sqrt{m}\norm{D_t}_F$. Applying Jensen's inequality across the average and expectation together with \eqref{eq:direct-sgd-descent} yields
\[
\frac1T\sum_{t=1}^T\E\norm{D_t}_*
\le \sqrt{m}\brac{\frac1T\sum_{t=1}^T\E\norm{D_t}_F^2}^{1/2}
\le \sqrt{m}\brac{\frac{2\Delta}{\eta T} + L_*\eta V}^{1/2}.
\]
Use the step size of \cref{thm:matrix-sgd-upper}, with $\eta=1/L_*$ when $V=0$. Taking the reciprocal of the minimum when $V>0$ gives
\[
 \frac1\eta=\max\bbrac{L_*,\sqrt{\frac{L_*VT}{2\Delta}}},
 \qquad L_*\eta V\le\sqrt{\frac{2A_0V}{T}}.
\]
Both relations also hold at $V=0$, where the noise term vanishes. Substituting them into the preceding bound gives
\[
 \frac{2\Delta}{\eta T}+L_*\eta V
 \le \max\bbrac{\frac{2A_0}{T},\sqrt{\frac{2A_0V}{T}}}
       +\sqrt{\frac{2A_0V}{T}}
 \le \frac{2A_0}{T}+2\sqrt{\frac{2A_0V}{T}}.
\]
Finally, $\sqrt{x+y}\le\sqrt{x}+\sqrt{y}$ yields
\[
 \frac1T\sum_{t=1}^T\E\norm{D_t}_*
 \le \sqrt{\frac{2mA_0}{T}}+2^{3/4}\brac{\frac{m^2A_0V}{T}}^{1/4}
 \le 2\sqbrac{\sqrt{\frac{mA_0}{T}}+\brac{\frac{m^2A_0V}{T}}^{1/4}},
\]
as claimed in \cref{thm:matrix-sgd-upper}.

\paragraph{Step 4: Short-horizon initial-gradient cap.}
Let $Q:=\msign{\nabla F(\WB_1)}$, so that $\norm{Q}_{\op}\le 1$ and $\inner{\nabla F(\WB_1)}{Q}=\norm{\nabla F(\WB_1)}_*$. Applying \cref{lem:gradient-taylor-bound} under \cref{ass:operator norm smooth} along $\WB_1-sQ$ for any $s\ge 0$ gives
\[
F(\WB_1-sQ)
\le F(\WB_1) - s\inner{\nabla F(\WB_1)}{Q} + \frac{L_*}{2}\norm{sQ}_{\op}^2
\le F(\WB_1) - s\norm{\nabla F(\WB_1)}_* + \frac{L_*s^2}{2}.
\]
Setting $s=\norm{\nabla F(\WB_1)}_*/L_*$ and invoking $F(\WB_1)-F(\WB_1-sQ)\le F(\WB_1)-\inf F\le\Delta$ yields
\begin{equation}\label{eq:direct-initial-cap}
\norm{\nabla F(\WB_1)}_*^2 \le 2L_*\Delta = 2A_0.
\end{equation}
Because $\WB_1$ is queried at $t=1$, we have $\expect{\min_{t\in[T]}\norm{D_t}_*}\le\norm{D_1}_*\le\sqrt{2A_0}$. Combining this bound with Step~3 gives $\expect{\min_{t\in[T]}\norm{D_t}_*}\le 2\,\mathcal B_T$.

\subsection{Lower Bound for Vanilla SGD}
\label{app:sgd-lower-bound}

We first prove the lower bound in \eqref{eq:appendix-sgd-all-horizon} for $L_*=\Delta=1$, initial matrix $\bar X_1$, and diagonal $\vSigma=\Diag{\sigma_1,\ldots,\sigma_m}$, so $V=\sum_{i=1}^m\sigma_i^2$. \Cref{app:matrix-rescaling} transfers the bound to the prescribed $(L_*,\Delta,\vSigma,\WB_1)$.

The Huber and periodic instances cover extreme step sizes. A circular oracle suffices when $0<\eta<4$ and $\eta V\ge u^2$, where $u$ is the target gradient level. In the remaining window for $m\ge4$, each column pulse reverses the drift of a slow diagonal average. Its displacement per step is proportional to $\eta$, while the pulse probability is proportional to $\eta^2$. The step size therefore cancels in the bound on its squared excursion in \cref{app:ordinary-sgd-proof}.

\subsubsection{Deterministic Step-Size Obstructions}
\label{app:direct-deterministic}

The following two noiseless vector instances also apply to \SGDMRef{}. We embed them along the leading diagonal to obtain a matrix lower bound for every $T$.

\begin{lemma}[Huber and periodic instances]\label{lem:direct-step-obstructions}
For every $\eta>0$, $\beta\in[0,1)$, and $T\ge 1$, there exists a normalized deterministic $d$-dimensional instance ($\Linf=\Delta=1$) with initialization $m_1=g_1$ satisfying
\begin{equation}\label{eq:direct-huber-risk}
\min_{t\le T}\norm{\nabla f(x_t)}_1
= \sqrt{\frac{d}{\eta T + d/2}}
\ge \frac12\min\bbrac{1,\, \sqrt{\frac{d}{\eta T}}}.
\end{equation}
Moreover, when $\eta\ge 4$, another normalized deterministic instance maintains $\norm{\nabla f(x_t)}_1=1$ at every query $t\ge 1$. For every $(\eta,\beta)$, embedding the Huber instance when $\eta<4$ and the periodic instance when $\eta\ge 4$ along the leading $m$ diagonal entries of $\R^{m\times n}$ yields a deterministic lower bound of at least $\frac14\min\{1,\sqrt{m/T}\}$.
\end{lemma}

\subsubsection{Column-and-Diagonal Hard Instance Geometry}
\label{app:direct-geometry}

To obtain the stochastic term $(m^2V/T)^{1/4}$ when $m\ge 4$ and $0<\eta<4$, we build a matrix potential $Q$ on $X\in\R^{m\times n}$ that couples a fast-switching control column to slow diagonal averages, matching the proof sketch in \cref{sec:sgd-lower-bound}.

\paragraph{Coordinate extraction and gradient compression.}
Partition $\{2,\ldots,m\}$ into three nonempty subsets $I_0,I_1,I_2$ whose cardinalities differ by at most one, and define
\[
k_j := \abs{I_j} \quad (j=0,1,2),
\qquad
k := \min_{j\in\{0,1,2\}} k_j = \left\lfloor\frac{m-1}{3}\right\rfloor.
\]
Direct inspection for $m\in\{4,5,6\}$ and $\lfloor(m-1)/3\rfloor\ge (m-3)/3\ge m/6$ for $m\ge 6$ verify that $k\ge m/6$ for all $m\ge 4$. For $X\in\R^{m\times n}$, extract the disjoint coordinate blocks
\begin{equation}\label{eq:direct-coordinate-maps}
s := X_{:,1} \in \R^m,
\qquad
z := X_{1,2} \in \R,
\qquad
w_j := \frac{1}{k_j}\sum_{i\in I_j} X_{ii} \in \R \quad (j=0,1,2).
\end{equation}
Here $s$ is the control column that receives sparse sign-flipping pulses, while $w_1,w_2$ are the slow diagonal averages (the remaining coordinates $z$ and $w_0$ are reserved for the strong-momentum construction in \cref{app:direct-matrix-sgdm}, while keeping the partition sizes $k_j$ identical across both appendices). Throughout, subscripts such as $F_{w_j}:=\partial F/\partial w_j$ and $F_z:=\partial F/\partial z$ denote partial derivatives with respect to the extracted coordinates, and $\operatorname{Lip}(\nabla\Phi)$ denotes the Euclidean Lipschitz constant of $\nabla\Phi$.

Each linear extraction map in \eqref{eq:direct-coordinate-maps} contracts the operator norm into its Euclidean or absolute-value norm:
\[
\norm{s}_2 = \norm{Xe_1}_2 \le \norm{X}_{\op},
\quad
\abs{z} = \abs{e_1^\top X e_2} \le \norm{X}_{\op},
\quad
\abs{w_j} \le \frac{1}{k_j}\sum_{i\in I_j}\abs{e_i^\top X e_i} \le \norm{X}_{\op}.
\]
By duality, the adjoint of each map has norm at most $1$ from the Euclidean or absolute-value norm into the nuclear norm. For any differentiable function $F(X)$ depending only on $(s,z,w_0,w_1,w_2)$, its gradient $\nabla_X F$ has $(1,2)$ entry $F_z$, and its submatrix on rows and columns $\{2,\ldots,m\}$ is diagonal with entry $F_{w_j}/k_j$ repeated $k_j$ times for each $j\in\{0,1,2\}$. Because restricting a matrix to a submatrix cannot increase its nuclear norm, restricting $\nabla_X F$ to rows and columns $\{2,\ldots,m\}$ or to entry $(1,2)$ yields
\begin{equation}\label{eq:direct-gradient-compression}
\norm{\nabla_X F}_* \ge \sum_{j=0}^2 \abs{F_{w_j}},
\qquad
\norm{\nabla_X F}_* \ge \abs{F_z}.
\end{equation}
On each diagonal entry $i\in I_j$, the partial derivative is $F_{w_j}/k_j$. When the oracle returns exact partial derivatives on those diagonal entries, an \SGDRef{} update satisfies $w_{j,t+1}-w_{j,t}=-(\eta/k_j)F_{w_j}(X_t)$ (and an \SGDMRef{} update with exact diagonal partials from the first query onward replaces $F_{w_j}(X_t)$ by its EMA initialized at $F_{w_j}(X_1)$), whereas the full magnitude $\abs{F_{w_j}}$ enters the nuclear gradient lower bound \eqref{eq:direct-gradient-compression}.

\paragraph{Building block 1: Smoothed sign-product signals.}
Given a transition radius $R>0$, define the $C^1$ sign transition $\psi:\R\to[-1,1]$ by $\psi(a):=\sin(\pi a/2)$ for $\abs{a}\le 1$ and $\psi(a):=\sign{a}$ for $\abs{a}>1$, so that $\psi'(\pm 1)=0$. Define also the odd $C^1$ ramp $\vartheta:\R\to\R$ given for $a\ge 0$ by
\[
\vartheta(a) :=
\begin{cases}
a, & 0\le a\le 1,\\
a - (a-1)^2/2, & 1\le a\le 2,\\
3/2, & a\ge 2.
\end{cases}
\]
Direct differentiation verifies that $\vartheta'$ is continuous and even with $\abs{\vartheta}\le 3/2$, $0\le\vartheta'\le 1$, and $\operatorname{Lip}(\vartheta')=1$. Using $\psi$, define the two column signals on $s\in\R^m$:
\begin{equation}\label{eq:direct-signals}
\Psi(s) := \prod_{i=1}^m \psi(s_i/R),
\qquad
\Phi_1(s) := \Psi(s),
\qquad
\Phi_2(s) := \frac{1-\Psi(s)^2}{2}.
\end{equation}
The second signal equals $1/2$ when the sign product crosses zero. For $\omega=\abs{\Psi(s)}\in[0,1]$,
\[
 \abs{\Phi_1(s)}+\Phi_2(s)=\omega+\frac{1-\omega^2}{2}
 =1-\frac{(1-\omega)^2}{2}\in[1/2,1],\qquad 0\le\Phi_2(s)\le 1/2.
\]
When all $\abs{s_i}\ge R$, $\Phi_1(s)=\prod_i\sign{s_i}$ and $\Phi_2(s)=0$. Both gradients vanish there because $\psi'(s_i/R)=0$ on every saturated coordinate. The next lemma bounds the derivatives independently of $m$ even within the transition region.

\begin{lemma}[Dimension-free derivatives of smoothed sign products]\label{lem:direct-signal-derivatives}
For any $R>0$, set $\kappa:=\pi/(2R)$. The signals $\Phi_1,\Phi_2$ in \eqref{eq:direct-signals} are $C^1$ on $\R^m$ and satisfy globally
\begin{equation}\label{eq:direct-signal-derivatives}
\norm{\nabla\Phi_j(s)}_2 \le \kappa \quad (j=1,2),
\qquad
\operatorname{Lip}(\nabla\Phi_1) \le \kappa^2,
\qquad
\operatorname{Lip}(\nabla\Phi_2) \le 3\kappa^2.
\end{equation}
\end{lemma}

\paragraph{Building block 2: Auxiliary column potential.}
To make a sparse sign-flipping pulse unbiased on the control column $s$, we add a separable column potential $H(s)$ whose partial derivative matches the conditional mean of the pulse. Set the core-width multiplier $K:=16$ (so the saturated core $R\le\abs{s_i}\le 2KR$ has ample margin around the nominal centers $\pm KR$), and let $\tau:\R\to[0,1]$ be the even continuous cutoff equal to $1$ on $[-2K,2K]$, $3-\abs{a}/K$ on $2K<\abs{a}<3K$, and $0$ for $\abs{a}\ge 3K$. Given pulse probabilities $p_i\ge 0$ with $p:=\sum_{i=1}^m p_i>0$, define
\begin{equation}\label{eq:direct-column-potential}
b(0) := 0,
\qquad
b'(a) := \tau(a)\sqbrac{a + K\psi(a)},
\qquad
H(s) := \sum_{i=1}^m \frac{p_i R^2}{\eta}\,b\!\brac{\frac{s_i}{R}}.
\end{equation}
Because $b'$ is odd and nonnegative on $[0,\infty)$ with support $[-3K,3K]$, $b$ is even and nonnegative on $\R$. For $0\le a\le 3K$, we have $0\le b'(a)\le a+K\le 4K$, and integrating over $[0,3K]$ yields $0\le b(a)\le 3K\cdot 4K = 12K^2$. Since $b'$ is continuous and piecewise smooth, checking $b''(a)$ on each piece for $a>0$ gives:
\begin{itemize}
    \item on $(0,1)$, $b''(a)=1+K\psi'(a)\in[1,1+\frac{\pi}{2}K]\subset[0,10K]$;
    \item on $(1,2K)$, $b''(a)=1\le 10K$;
    \item on $(2K,3K)$, $b'(a)=(3-a/K)(a+K)$, so $b''(a)=2-2a/K\in(-4,-2)$ and $\abs{b''(a)}\le 4\le 10K$;
    \item on $(3K,\infty)$, $b''(a)=0$.
\end{itemize}
Thus $\operatorname{Lip}(b')\le 10K$. Because $H_{s_i}(s)=\frac{p_i R}{\eta}b'(s_i/R)$ and $H$ is separable across coordinates, we obtain
\begin{equation}\label{eq:direct-H-bounds}
0 \le H(s) \le \frac{12K^2 R^2 p}{\eta},
\qquad
\norm{\nabla H(s)}_2 \le \frac{4KRp}{\eta},
\qquad
\operatorname{Lip}(\nabla H) \le \frac{10Kp}{\eta}.
\end{equation}
In the \emph{saturated core} $R\le\abs{s_i}\le 2KR$, we have $\tau(s_i/R)=1$ and $\psi(s_i/R)=\sign{s_i}$, so the partial derivative simplifies to
\begin{equation}\label{eq:direct-H-saturated}
H_{s_i}(s) = \frac{p_i}{\eta}\sqbrac{s_i + KR\sign{s_i}},
\end{equation}
which matches the conditional expectation of a pulse that reflects $s_i$ to $-KR\sign{s_i}$.

\paragraph{Coupled potential and global geometry.}
To keep the Hessian of the coupled potential well below $1$, we set the reference radius $R_0:=100$ and amplitude divisor $C_A:=1024$, collecting the three universal geometric constants as
\begin{equation}\label{eq:direct-constants}
K = 16,
\qquad
R_0 = 100,
\qquad
C_A = 1024.
\end{equation}
Given any target gradient level $u>0$ and any admissible scale pair $(E,R)$ satisfying
\begin{equation}\label{eq:direct-scales}
E \ge 1,
\qquad
R \ge R_0,
\qquad
\frac{E}{R^2} \le \frac{1}{R_0^2},
\qquad
A := \frac{E}{C_A},
\qquad
B := \frac{A}{u},
\end{equation}
we assemble the column-and-diagonal potential
\begin{equation}\label{eq:direct-Q}
Q(s,w_1,w_2)
:= H(s) + A\sqbrac{\vartheta\!\brac{\frac{w_1}{B}}\Phi_1(s) + \vartheta\!\brac{\frac{w_2}{B}}\Phi_2(s)}.
\end{equation}
Whenever $\abs{w_1},\abs{w_2}\le B$, we have $\vartheta'(w_j/B)=1$ and $A/B=u$, so $Q_{w_j}=u\Phi_j(s)$ for $j\in\{1,2\}$. Vanilla \SGDRef{} (\cref{app:ordinary-sgd-proof}) instantiates \eqref{eq:direct-scales} with $(E,R)=(1,R_0)$ and sets $F(X)=Q(s,w_1,w_2)$, whereas \cref{app:direct-matrix-sgdm} scales $(E,R)$ within \eqref{eq:direct-scales} and combines $Q/E$ with a momentum driver on $z$ and guard on $w_0$. To certify global admissibility (initial gap and unit operator-norm smoothness) when instantiating these pulse constructions in \cref{app:ordinary-sgd-proof} and \cref{app:direct-reset-oracle}, the next lemma bounds the values, derivatives, and smoothness of $Q$.

\begin{lemma}[Global geometry of $Q$]\label{lem:direct-global-geometry}
Let $(E,R,A,B)$ satisfy \eqref{eq:direct-scales}, and suppose that
\begin{equation}\label{eq:direct-geometry-condition}
u \le 2^{-12},
\qquad
\frac{12K^2R^2p}{\eta E} \le \frac{1}{128}.
\end{equation}
Write $h_g:=4KRp/\eta$, $h_2:=10Kp/\eta$, and $\kappa:=\pi/(2R)$. Then:
\begin{enumerate}
    \item \emph{Value bounds and initial gap.} Globally on $\R^m\times\R\times\R$,
    \[
    \frac{Q(s,w_1,w_2)}{E} \in \sqbrac{-\frac{3}{C_A},\, \frac{1}{128} + \frac{3}{C_A}}.
    \]
    In particular, when $E=1$ and $\bar X_1$ is initialized at $s_i=KR$ ($i\in[m]$) and $w_1=w_2=0$, $Q$ is bounded below and satisfies $Q(\bar X_1)-\inf Q \le 1$.
    \item \emph{Gradient and Hessian bounds.} Globally on $\R^m\times\R\times\R$,
    \[
    \norm{\nabla_s Q}_2 \le h_g + 3A\kappa,
    \qquad
    \abs{Q_{w_j}} \le u \quad (j=1,2),
    \]
    and both $\operatorname{Lip}(\nabla Q)$ and the operator-to-nuclear smoothness constant of $X\mapsto Q(s(X),w_1(X),w_2(X))$ are at most $h_2+6A\kappa^2+4u\kappa+2u^2/A$. These quantities satisfy
    \begin{equation}\label{eq:direct-hessian-bound}
    h_2 + 6A\kappa^2 + 4u\kappa + \frac{2u^2}{A} + \frac{2(h_g+3A\kappa)+4u}{E}
    < \frac{1}{200}.
    \end{equation}
    In particular, when $E=1$, the matrix objective $F(X)=Q(s(X),w_1(X),w_2(X))$ is globally $1$-smooth with respect to the operator norm.
\end{enumerate}
\end{lemma}

\subsubsection{Circular Oracle for Moderate Step Sizes}
\label{app:direct-circular}

When the step size lies in the moderate regime $0<\eta<4$ with $\eta V \ge u^2$, the rotational-noise construction below already yields a lower bound of at least $u/32$ without relying on slow diagonal drift. In this regime, \cref{app:ordinary-sgd-proof} (and \cref{app:direct-sgdm-interpolation} for \SGDMRef{}) uses this circular oracle, under which each active row stays at fixed radius $R_i$ on its own first-query plus branch (occurring with probability $1/2$).

\begin{lemma}[Circular oracle]\label{lem:direct-circular}
For any $0<\eta<4$, $\beta\in[0,1)$, and $V>0$, there exist a normalized objective $F$ and an adapted oracle $\O$ satisfying \cref{ass:Matrix-bounded-below,ass:Matrix-variance,ass:operator norm smooth} such that, for every $T\ge 1$,
\begin{equation}\label{eq:direct-circular-risk}
\expect{\min_{t\le T}\norm{\nabla F(X_t)}_*}
\ge \frac{1}{32}\min\bbrac{1,\, \sqrt{\eta V}}.
\end{equation}
\end{lemma}

\subsubsection{Proof of the Normalized Lower Bound}
\label{app:ordinary-sgd-proof}

\begin{proof}[Proof of \cref{thm:matrix-sgd-all-horizon} and the lower bound in \eqref{eq:appendix-sgd-all-horizon}]
For each fixed $\eta>0$, \cref{lem:direct-step-obstructions} gives a deterministic instance with $\min_{t\le T}\norm{\nabla F(X_t)}_*\ge\frac14\min\{1,\sqrt{m/T}\}$. This proves the bound when $V=0$.

Assume now that $V>0$. To establish the stochastic term $\Omega(\min\{1,(m^2V/T)^{1/4}\})$ for $m\ge 4$, set the target level
\[
u := c_0\min\bbrac{1,\, \brac{\frac{m^2V}{T}}^{1/4}},
\qquad
c_0 := 2^{-40}.
\]
Here $c_0=2^{-40}$ is small enough that $u\le 2^{-12}$ (satisfying the first inequality in \eqref{eq:direct-geometry-condition}) and that the persistent-walk failure probability in Step~3 below is strictly less than $1/8$. We cover all fixed step sizes $\eta>0$ by four exhaustive regimes:
\begin{center}
\footnotesize
\setlength{\tabcolsep}{5pt}
\renewcommand{\arraystretch}{1.25}
\begin{tabular}{lll}
\toprule
\textbf{Step-size regime} & \textbf{Hard instance} & \textbf{Lower bound} \\
\midrule
Large step ($\eta \ge 4$) & Periodic (\cref{lem:direct-step-obstructions}) & $= 1 \ge u$ \\
Moderate step ($0<\eta<4$, $\eta V \ge u^2$) & Circular oracle (\cref{lem:direct-circular}) & $\ge u/32$ \\
Small step ($0<\eta<4$, $\eta \le m/(u^2T)$) & Diagonal Huber (\cref{lem:direct-step-obstructions}) & $\ge \sqrt{2/3}\,u$ \\
Residual ($\frac{m}{u^2T}<\eta<4$, $\eta V < u^2$) & Column-and-diagonal (\cref{app:direct-geometry}) & $\ge 7u/8$ ($m\ge 4$) \\
\bottomrule
\end{tabular}
\end{center}
The first three rows follow directly from our earlier lemmas:
\begin{itemize}
    \item If $\eta\ge 4$, the periodic instance of \cref{lem:direct-step-obstructions} gives gradient norm $1\ge u$.
    \item If $0<\eta<4$ and $\eta V\ge u^2$, \cref{lem:direct-circular} gives risk at least $\frac{1}{32}\min\{1,\sqrt{\eta V}\}\ge u/32$.
    \item If $0<\eta<4$ and $\eta\le m/(u^2T)$, the Huber instance of \cref{lem:direct-step-obstructions} gives risk
    \[
    \sqrt{\frac{m}{\eta T+m/2}}
    \ge \sqrt{\frac{m}{m/u^2 + m/2}}
    = \frac{u}{\sqrt{1+u^2/2}}
    \ge \sqrt{\frac{2}{3}}\,u,
    \]
    since $u\le 1$.
\end{itemize}
It remains to analyze the residual intermediate regime for $m\ge 4$:
\begin{equation}\label{eq:direct-sgd-residual}
\frac{m}{u^2T} < \eta < 4,
\qquad
\eta V < u^2.
\end{equation}

\paragraph{1. Instantiation and oracle specification ($m\ge 4$).}
Specialize the geometry of \cref{app:direct-geometry} to $(E,R)=(1,R_0)$ with $A=1/C_A$ and $B=A/u$, setting $F(X):=Q(s,w_1,w_2)$ in \eqref{eq:direct-Q}. Define the pulse-calibration constant $c_f:=(64K^2R_0^2)^{-1}$ (chosen so the pulse variance stays below $\sigma_i^2$) and coordinate pulse probabilities
\[
p_i := c_f\eta^2\sigma_i^2 \quad (i\in[m]),
\qquad
p := \sum_{i=1}^m p_i = c_f\eta^2 V.
\]
From \eqref{eq:direct-sgd-residual}, we have $p = c_f\eta(\eta V) < 4c_f u^2 < 1/4$. Initialize $\bar X_1$ at $s_{1,i}=KR$ ($i\in[m]$) and $w_{1,1}=w_{2,1}=0$.

Define the conditional stochastic oracle $\O$ globally on all histories as follows:
\begin{enumerate}
    \item At $t=1$, the oracle returns the exact gradient $\nabla F(X_1)$. Since $s_{1,i}=KR\in[R,2KR]$, both signal column gradients vanish and \eqref{eq:direct-H-saturated} gives $H_{s_i}(s_1)=2p_i KR/\eta$. The first \SGDRef{} update therefore places the control column at
    \[
    s_{2,i} = KR - \eta H_{s_i}(s_1) = KR(1-2p_i) \qquad (i\in[m]).
    \]
    Because $0\le p_i\le p<1/4$, we have $s_{2,i}\in(\frac12 KR, KR]\subset[R,2KR]$ (as $K=16$), so $s_2$ stays inside the saturated core.
    \item At every query $t\ge 2$, as long as the history is consistent with the construction, the slow coordinates satisfy $\abs{w_{1,t}},\abs{w_{2,t}}\le B$, and every column entry lies in the saturated core $R\le\abs{s_{t,i}}\le 2KR$, the oracle draws a sparse column pulse
    \begin{equation}\label{eq:direct-sgd-pulse}
    g_s =
    \begin{cases}
    \sqbrac{s_{t,i} + KR\sign{s_{t,i}}}e_i/\eta, & \text{with probability } p_i \quad (i\in[m]),\\
    0, & \text{with probability } 1-p,
    \end{cases}
    \end{equation}
    on the first column, and returns exact partial derivatives on all other matrix entries.
    \item If the trajectory ever exits the safe region ($\abs{w_{1,t}}>B$ or $\abs{w_{2,t}}>B$) or deviates from a valid history, the oracle permanently defaults to exact gradients.
\end{enumerate}

\paragraph{2. Verification of unbiasedness, PSD covariance, and global geometry.}
At every active query $t\ge 2$, $s_t$ lies in the saturated core, so $\nabla_s\Phi_1(s_t)=\nabla_s\Phi_2(s_t)=0$ and \eqref{eq:direct-H-saturated} yields
\[
\E\sqbrac{g_{s,i}\mid\F_{t-1}}
= p_i\frac{s_{t,i}+KR\sign{s_{t,i}}}{\eta}
= H_{s_i}(s_t)
= Q_{s_i}(X_t).
\]
All other entries receive exact gradients, and fallback steps have zero noise, so $\O$ is conditionally unbiased everywhere.

To verify \cref{ass:Matrix-variance}, note that in the saturated core $\abs{s_{t,i}+KR\sign{s_{t,i}}}=\abs{s_{t,i}}+KR\le 3KR$. Because at most one coordinate is pulsed per query, $g_s g_s^\top$ is diagonal with $(i,i)$ entry at most $(3KR/\eta)^2$ occurring with probability $p_i$. Subtracting the rank-one mean outer product and substituting $p_i=c_f\eta^2\sigma_i^2$ with $c_f=(64K^2R_0^2)^{-1}$ and $R=R_0$ gives
\begin{equation}\label{eq:direct-PSD-calibration}
\E\sqbrac{(g_s-\E_{t-1}g_s)(g_s-\E_{t-1}g_s)^\top\mid\F_{t-1}}
\preceq \operatorname{diag}_{i=1}^m\!\brac{p_i\brac{\frac{3KR}{\eta}}^2}
= \frac{9}{64}\operatorname{diag}_{i=1}^m(\sigma_i^2).
\end{equation}
Because the matrix noise $\Nb_t=(g_s-\E_{t-1}g_s)e_1^\top$ is supported on the first column, we have $\Nb_t\Nb_t^\top=(g_s-\E_{t-1}g_s)(g_s-\E_{t-1}g_s)^\top$, which verifies $\E[\Nb_t\Nb_t^\top\mid\F_{t-1}]\preceq\vSigma^2$.

Finally, $u\le c_0<2^{-12}$ and \eqref{eq:direct-sgd-residual} give
\[
\frac{12K^2R^2p}{\eta E}
= 12K^2R_0^2\cdot\frac{c_f\eta^2 V}{\eta}
= \frac{3}{16}\eta V
< \frac{3}{16}u^2
< \frac{1}{128}.
\]
Condition \eqref{eq:direct-geometry-condition} therefore holds, so \cref{lem:direct-global-geometry} guarantees $F(\bar X_1)-\inf F\le 1$ and operator-norm smoothness with $L_*\le 1$.

\paragraph{3. Persistent random walk and maximal excursion bound.}
At any active query $t\ge 2$, a pulse at coordinate $i$ updates $s_{t+1,i}=s_{t,i}-\eta g_{s,i}=-KR\sign{s_{t,i}}$, flipping $\sign{s_{t,i}}$ while keeping $\abs{s_{t+1,i}}=KR\in[R,2KR]$; a zero sample leaves $s_t$ unchanged. Thus, prior to any slow-coordinate exit, every column coordinate remains in the saturated core throughout the run. Hence $\Phi_2(s_t)=0$, $w_{2,t}=0$, and $Q_{w_1}(X_t)=u\Phi_1(s_t)=u\prod_{i=1}^m\sign{s_{t,i}}$.

Pre-sample i.i.d.\ $\operatorname{Bernoulli}(p)$ pulse indicators $\{\xi_j\}_{j\ge 1}$ on an auxiliary pulse-query clock with filtration $\mathcal H_0$ trivial and $\mathcal H_l:=\sigma(\xi_1,\ldots,\xi_l)$ for $l\ge 1$ (distinct from the actual query filtration $\F_t$), and define the sign-product Markov chain $\zeta_1:=1$, $\zeta_{j+1}:=(-1)^{\xi_j}\zeta_j\in\{\pm 1\}$ (here $\zeta_j$ denotes the column sign-product state, distinct from the target gradient scale in \cref{sec:proof-ssgd-lowerbound}). Since $s_{2,i}=KR(1-2p_i)>0$ for all $i$, we have $\prod_{i=1}^m\sign{s_{t,i}}=\zeta_{t-1}$ for every $t\ge 2$ prior to fallback.

To bound the partial sums $S_0:=0$ and $S_l:=\sum_{j=1}^l\zeta_j$ ($l\ge 1$), introduce the increments $M_{j+1}:=\zeta_{j+1}-(1-2p)\zeta_j$. Because $\zeta_j$ is $\mathcal H_{j-1}$-measurable and $\xi_j$ is $\mathcal H_j$-measurable and independent of $\mathcal H_{j-1}$ with $\E[(-1)^{\xi_j}\mid\mathcal H_{j-1}]=1-2p$, each $M_{j+1}$ is $\mathcal H_j$-measurable and satisfies $\E[M_{j+1}\mid\mathcal H_{j-1}]=0$. Since $M_{j+1}$ equals $+2p\zeta_j$ with probability $1-p$ and $-2(1-p)\zeta_j$ with probability $p$, its conditional variance is
\[
\operatorname{Var}(M_{j+1}\mid\mathcal H_{j-1})
= (1-p)(2p)^2 + p\brac{2(1-p)}^2
= 4p(1-p)
\le 4p.
\]
Rewriting $M_{j+1}=\zeta_{j+1}-\zeta_j+2p\zeta_j$ and summing over $j=1,\ldots,l$ gives the telescoping identity
\[
2pS_l = \zeta_1 - \zeta_{l+1} + Y_l,
\qquad
Y_0 := 0,
\qquad
Y_l := \sum_{j=1}^l M_{j+1}.
\]
Then $\{Y_l\}_{l=0}^T$ is an $\{\mathcal H_l\}_{l=0}^T$-martingale. Applying $(a+b)^2\le 2a^2+2b^2$ and $\abs{\zeta_1-\zeta_{l+1}}\le 2$ gives $4p^2S_l^2\le 8 + 2Y_l^2$. By Doob's $L^2$ maximal inequality for $\{\mathcal H_l\}_{l=0}^T$,
\[
\E\sqbrac{\max_{l\le T} Y_l^2}
\le 4\,\E\sqbrac{Y_T^2}
= 4\sum_{j=1}^T \E\sqbrac{\operatorname{Var}(M_{j+1}\mid\mathcal H_{j-1})}
\le 16pT,
\]
and dividing $4p^2\E\max_{l\le T}\abs{S_l}^2\le 8+32pT$ by $4p^2$ yields
\begin{equation}\label{eq:direct-sign-maximal}
\E\sqbrac{\max_{l\le T}\abs{S_l}^2}
\le \frac{2}{p^2} + \frac{8T}{p}.
\end{equation}

Next, \eqref{eq:direct-sgd-residual} and $u\le c_0(m^2V/T)^{1/4}$ imply
\[
pT = c_f\eta^2 VT > c_f\frac{m^2 V}{u^4 T} \ge c_f c_0^{-4} > 2^{160-28} > 4,
\]
since $c_f^{-1}=64K^2R_0^2=2^{14}\cdot 10^4<2^{28}$. Because $pT>4$, we have $\frac{2}{p^2}<\frac{T}{2p}$, so \eqref{eq:direct-sign-maximal} simplifies to $\E\max_{l\le T}\abs{S_l}^2 < \frac{17T}{2p}$. Set $L_0:=kA/(4\eta u^2)$. Using $k\ge m/6$, $\eta<4$, and $u\le 2^{-40}$, we check $L_0 > \frac{m}{96C_A u^2}\ge 1$. Applying Markov's inequality, substituting $p=c_f\eta^2 V$ (so $\eta^2$ cancels), and using $k\ge m/6$, $A=1/C_A$, and $64C_A^2 K^2 R_0^2 = 2^{34}\cdot 10^4 < 2^{48}$ gives
\begin{align*}
\Pr\brac{\max_{l\le T}\abs{S_l} > L_0}
&\le \frac{\E\max_{l\le T}\abs{S_l}^2}{L_0^2}
\le \frac{16\eta^2 u^4}{k^2 A^2}\cdot\frac{17T}{2p}
= \frac{136\eta^2 u^4 T}{k^2 A^2 p} \\
&\le 136\cdot 36\cdot\brac{64C_A^2 K^2 R_0^2}\frac{u^4 T}{m^2 V}
< 136\cdot 36\cdot 2^{48} c_0^4
< \frac18.
\end{align*}

\paragraph{4. High-probability safety and gradient lower bound.}
On the complementary event $\mathcal E:=\{\max_{l\le T}\abs{S_l}\le L_0\}$ (which has $\Pr(\mathcal E)\ge 7/8$), consider any query $t\in[T]$ up to a putative first exit of $w_1$ from $[-B,B]$. At $t=1$, $w_{1,1}=0$ and $Q_{w_1}(X_1)=u$; at active queries $r\ge 2$, $Q_{w_1}(X_r)=u\zeta_{r-1}$. Since each diagonal entry $i\in I_1$ updates by $-\frac{\eta}{k_1}Q_{w_1}(X_r)$, we have $w_{1,t}=-\frac{\eta u}{k_1}(1+\sum_{j=1}^{t-2}\zeta_j)$ for $2\le t\le T$ (with the empty sum equal to $0$ at $t=2$), and hence
\[
\abs{w_{1,t}}
\le \frac{\eta u}{k}\brac{1 + \max_{l\le T}\abs{S_l}}
\le \frac{\eta u}{k} + \frac{\eta u}{k}L_0
= \frac{\eta u}{k} + \frac{B}{4}
\le \frac{B}{2},
\]
where the last step uses $L_0\ge 1$ (so $\frac{\eta u}{k}\le\frac{\eta u}{k}L_0=\frac{B}{4}$). Thus, on $\mathcal E$, $\abs{w_{1,t}}\le B/2$, $w_{2,t}=0$, and the oracle does not fall back at any query $t\in[T]$. By \eqref{eq:direct-gradient-compression}, $\norm{\nabla F(X_t)}_*\ge\abs{Q_{w_1}(X_t)}=u$ holds at every query $t\in[T]$ on $\mathcal E$, yielding
\[
\expect{\min_{t\le T}\norm{\nabla F(X_t)}_*} \ge \frac{7}{8}u.
\]

\paragraph{5. Small dimensions ($m\in\{2,3\}$) and conclusion.}
For $m\in\{2,3\}$, set $\eta_0:=(m/(VT))^{1/2}$:
\begin{itemize}
    \item If $\eta\ge 4$, the periodic instance of \cref{lem:direct-step-obstructions} gives risk $1$.
    \item If $0<\eta<4$ and $\eta\ge \eta_0$, \cref{lem:direct-circular} gives risk at least $\frac{1}{32}\min\{1,(mV/T)^{1/4}\}$.
    \item If $0<\eta<4$ and $\eta<\eta_0$, the Huber instance of \cref{lem:direct-step-obstructions} gives risk at least
    \[
    \sqrt{\frac{m}{\eta T+m/2}}
    \ge \frac{1}{\sqrt{\sqrt{T/(mV)}+1/2}}
    \ge \frac12\min\bbrac{1,\, \brac{\frac{mV}{T}}^{1/4}}.
    \]
\end{itemize}
Because $m\le 3$, we have $(mV/T)^{1/4}\ge 3^{-1/4}(m^2V/T)^{1/4}$, so in all three cases the risk is at least $\frac{1}{32\cdot 3^{1/4}}\min\{1,(m^2V/T)^{1/4}\}$.

Combining all dimensions $m\ge 2$ and all step sizes $\eta>0$, we have one normalized instance with risk at least $\frac14\min\{1,\sqrt{m/T}\}$ and (when $V>0$) another with risk at least $\frac{c_0}{32}\min\{1,(m^2V/T)^{1/4}\}$. Selecting the harder of these two complete instances and using $\max\{\min\{1,a\},\min\{1,b\}\}\ge\frac12\min\{1,a+b\}$ yields
\[
\expect{\min_{t\le T}\norm{\nabla F(X_t)}_*}
\ge c\,\min\bbrac{1,\, \sqrt{\frac{m}{T}} + \brac{\frac{m^2V}{T}}^{1/4}}
\]
with $c:=c_0/64$. Applying the affine-orthogonal rescaling of \cref{app:matrix-rescaling} restores arbitrary $L_*,\Delta,\vSigma,\WB_1$ and yields the lower bound $c\,\mathcal B_T$, completing the proof of \eqref{eq:appendix-sgd-all-horizon} and \cref{thm:matrix-sgd-all-horizon}.
\end{proof}

\subsubsection{Scale Restoration and Orthogonal Embedding}
\label{app:matrix-rescaling}

We now transfer any normalized lower-bound construction $(\bar F,\bar\O)$ with parameters $\bar L_*=\bar\Delta=1$, initial matrix $\bar X_1\in\R^{m\times n}$, and diagonal noise matrix $\bar\vSigma=\Diag{\sigma_1,\ldots,\sigma_m}/\sqrt{A_0}$ to arbitrary prescribed parameters $(L_*,\Delta,\vSigma,\WB_1)$ and fixed $(\eta,\beta)\in(0,\infty)\times[0,1)$.

Diagonalize the prescribed noise matrix as $\vSigma=U\Diag{\sigma_1,\ldots,\sigma_m}U^\top$ for an orthogonal matrix $U\in\R^{m\times m}$. Run the normalized construction with normalized step size $\bar\eta:=\eta L_*$ (and the same momentum $\beta$) via the affine-orthogonal map
\[
X(\WB) := \bar X_1 + \sqrt{\frac{L_*}{\Delta}}\,U^\top(\WB-\WB_1),
\qquad
F(\WB) := \Delta\,\bar F(X(\WB)).
\]
At each query $\WB_t$, the lifted oracle $\O$ maps the query history to normalized coordinates $X_1,\ldots,X_t$, invokes $\bar\O$ to obtain $\bar G_t$ (preserving any revealed internal state), and returns $\Gb_t:=\sqrt{A_0}\,U\bar G_t$. We verify all four reduction properties:
\begin{itemize}
    \item \emph{Gradient and operator-norm smoothness.} By the chain rule,
    \[
    \nabla F(\WB) = \Delta\sqrt{\frac{L_*}{\Delta}}\,U\nabla\bar F(X(\WB)) = \sqrt{A_0}\,U\nabla\bar F(X(\WB)).
    \]
    By orthogonal invariance of the nuclear and operator norms ($\norm{UM}_*=\norm{M}_*$ and $\norm{U^\top M}_{\op}=\norm{M}_{\op}$), we obtain $\norm{\nabla F(\WB)}_*=\sqrt{A_0}\,\norm{\nabla\bar F(X(\WB))}_*$ and
    \begin{align*}
    \norm{\nabla F(\WB)-\nabla F(\WB')}_*
    &= \sqrt{A_0}\,\norm{\nabla\bar F(X(\WB))-\nabla\bar F(X(\WB'))}_* \\
    &\le \sqrt{A_0}\,\norm{X(\WB)-X(\WB')}_{\op} \\
    &= \sqrt{A_0}\sqrt{\frac{L_*}{\Delta}}\,\norm{\WB-\WB'}_{\op}
     = L_*\norm{\WB-\WB'}_{\op}.
    \end{align*}
    \item \emph{Initial gap.} Since $X(\WB_1)=\bar X_1$, we have $F(\WB_1)-\inf F = \Delta(\bar F(\bar X_1)-\inf\bar F)\le\Delta$.
    \item \emph{Oracle unbiasedness and PSD covariance.} The lifted noise $\Nb_t=\Gb_t-\nabla F(\WB_t)=\sqrt{A_0}\,U\bar N_t$ satisfies $\E[\Nb_t\mid\F_{t-1}]=\sqrt{A_0}\,U\E[\bar N_t\mid\F_{t-1}]=\vzero$ and
    \[
    \E\sqbrac{\Nb_t\Nb_t^\top\mid\F_{t-1}}
    = A_0\,U\,\E\sqbrac{\bar N_t\bar N_t^\top\mid\F_{t-1}}U^\top
    \preceq A_0\,U\bar\vSigma^2 U^\top
    = \vSigma^2,
    \]
    with zero eigenvalues of $\vSigma$ retaining zero noise budget.
    \item \emph{Exact trajectory coupling (for both \SGDRef{} and \SGDMRef{}).} Let $\bar M_1=\bar G_1$ and $\bar M_t=\beta\bar M_{t-1}+(1-\beta)\bar G_t$ be the normalized momentum iterates. Because $\Gb_t=\sqrt{A_0}\,U\bar G_t$ and both the initialization $\MB_1=\Gb_1$ and the EMA recursion are linear, induction gives $\MB_t=\sqrt{A_0}\,U\bar M_t$ for all $t\ge 1$. Applying $X(\cdot)$ to $\WB_{t+1}=\WB_t-\eta\MB_t$ yields
    \[
    X(\WB_{t+1})
    = X(\WB_t) - \eta\sqrt{\frac{L_*}{\Delta}}\,U^\top\brac{\sqrt{A_0}\,U\bar M_t}
    = X(\WB_t) - (\eta L_*)\bar M_t
    = X(\WB_t) - \bar\eta\bar M_t.
    \]
    Since $X(\WB_1)=\bar X_1$, we have $X(\WB_t)=X_t$ pathwise at every query $t\ge 1$.
\end{itemize}
Thus $\min_{t\in[T]}\norm{\nabla F(\WB_t)}_* = \sqrt{A_0}\min_{t\in[T]}\norm{\nabla\bar F(X_t)}_*$, which transfers every normalized lower bound directly to the prescribed parameters $(L_*,\Delta,\vSigma,\WB_1)$.

\subsection{Tight Iteration Complexity of Vanilla SGD}
\label{sec:complexity-conversion}

Write the uncapped rate in \eqref{eq:appendix-sgd-all-horizon} as
\[
 S(T):=\sqrt{\frac{mA_0}{T}}+\brac{\frac{m^2A_0V}{T}}^{1/4}.
\]
Let $0<c\le 1\le C$ be the constants in the all-horizon bounds there. Fix an accuracy in the range
\begin{equation}\label{eq:matrix-complexity-accuracy-range}
 0<\epsilon<\frac{c}{2}\sqrt{A_0}.
\end{equation}
For every fixed step size, the worst-case risk is at least $c\min\{\sqrt{A_0},S(T)\}$. Since $c\sqrt{A_0}>\epsilon$, any horizon achieving risk at most $\epsilon$ must satisfy $S(T)\le\epsilon/c$. Each summand is nonnegative, so
\[
 T\ge \max\bbrac{\frac{c^2mA_0}{\epsilon^2},\,
                    \frac{c^4m^2A_0V}{\epsilon^4}}
 \ge \frac12\brac{\frac{c^2mA_0}{\epsilon^2}+\frac{c^4m^2A_0V}{\epsilon^4}}.
\]
Conversely, choose the integer horizon
\[
 T=\left\lceil\frac{4C^2mA_0}{\epsilon^2}
             +\frac{16C^4m^2A_0V}{\epsilon^4}\right\rceil
\]
and use the fixed step size from the horizon-$T$ upper bound. Then
\[
 C\sqrt{\frac{mA_0}{T}}\le\frac{\epsilon}{2},
 \qquad
 C\brac{\frac{m^2A_0V}{T}}^{1/4}\le\frac{\epsilon}{2}.
\]
The second term is zero when $V=0$. Thus the worst-case risk is at most $CS(T)\le\epsilon$. The ceiling adds less than one, whereas $4C^2mA_0/\epsilon^2>1$ by \eqref{eq:matrix-complexity-accuracy-range}. The two bounds give the complexity in \cref{thm:sgd-tight-complexity}, since $A_0=L_*\Delta$ and $V=\norm{\vSigma}_F^2$.

\subsection{Proofs of Supporting Lemmas}
\label{app:sgd-supporting-lemmas}

\begin{proof}[Proof of \cref{lem:direct-step-obstructions}]
We construct two deterministic instances in $\R^d$ (with exact gradients, so every variance bound holds with zero noise) and then lift them to $\R^{m\times n}$.

\paragraph{1. Huber instance (small $\eta$).}
For parameters $a,\ell>0$, consider the one-dimensional Huber function
\[
h_{a,\ell}(z) :=
\begin{cases}
\dfrac{\ell}{2}z^2, & \abs{z}\le a/\ell,\\[4pt]
a\abs{z} - \dfrac{a^2}{2\ell}, & \abs{z}\ge a/\ell.
\end{cases}
\]
Its derivative $h'_{a,\ell}(z)$ equals $\ell z$ for $\abs{z}\le a/\ell$ and $a\sign{z}$ for $\abs{z}\ge a/\ell$; it is continuous everywhere and $\ell$-Lipschitz ($0\le h''_{a,\ell}\le\ell$ on each piece). Set
\[
\ell := \frac1d,
\qquad
a := \frac{1}{\sqrt{d(\eta T + d/2)}},
\]
and define the separable objective $f(x):=\sum_{i=1}^d h_{a,\ell}(x_i)$, initialized at $x_{1,i}:=a/\ell+\eta a T$ for every $i\in[d]$. We verify admissibility and trace the trajectory:
\begin{itemize}
    \item \emph{Smoothness and initial gap.} For any $x,y\in\R^d$,
    \begin{align*}
    \norm{\nabla f(x)-\nabla f(y)}_1
    &= \sum_{i=1}^d \abs{h'_{a,\ell}(x_i)-h'_{a,\ell}(y_i)} \\
    &\le \sum_{i=1}^d \ell\abs{x_i-y_i}
     \le (d\ell)\norm{x-y}_\infty
     = \norm{x-y}_\infty,
    \end{align*}
    so $f$ is $1$-smooth in the $\ell_\infty/\ell_1$ geometry. Its infimum is $\inf f = f(0) = 0$, and since $x_{1,i}>a/\ell$,
    \[
    f(x_1)
    = d\brac{a\brac{\frac{a}{\ell}+\eta a T} - \frac{a^2}{2\ell}}
    = da^2\brac{\eta T + \frac{1}{2\ell}}
    = da^2\brac{\eta T + \frac{d}{2}}
    = 1.
    \]
    \item \emph{Trajectory induction.} We prove by induction on $t\in[T]$ that $x_{t,i}=a/\ell+\eta a(T-t+1)$ and $m_{t,i}=a$ for every $i\in[d]$. At $t=1$, $x_{1,i}=a/\ell+\eta a T\ge a/\ell$, so $g_{1,i}=h'_{a,\ell}(x_{1,i})=a$ and $m_{1,i}=g_{1,i}=a$. Assuming the claim at step $t\le T-1$, the update gives
    \[
    x_{t+1,i} = x_{t,i} - \eta m_{t,i} = \frac{a}{\ell} + \eta a(T-t) \ge \frac{a}{\ell},
    \]
    whence $g_{t+1,i}=a$ and $m_{t+1,i}=\beta m_{t,i}+(1-\beta)g_{t+1,i}=a$. This completes the induction.
\end{itemize}
Therefore $\norm{\nabla f(x_t)}_1 = da = \sqrt{d/(\eta T+d/2)}$ for every $t\in[T]$. The inequality in \eqref{eq:direct-huber-risk} follows because $\eta T + d/2 \le 2\max\{\eta T, d/2\}$, which gives
\[
\sqrt{\frac{d}{\eta T + d/2}}
\ge \min\bbrac{1,\, \sqrt{\frac{d}{2\eta T}}}
\ge \frac12\min\bbrac{1,\, \sqrt{\frac{d}{\eta T}}}.
\]

\paragraph{2. Periodic instance (large $\eta\ge 4$).}
When $\eta\ge 4$, define $j:=\lfloor\eta/4\rfloor\ge 1$ and $\ell:=4j/\eta$. Since $j\le\eta/4<j+1$, we have $\ell\le 1$ and $\ell > \frac{j}{j+1} \ge \frac12$, so $\ell\in[\frac12,1]$. Let $v_0:\R\to\R$ be the $(4/\ell)$-periodic extension of
\[
v_0(s) :=
\begin{cases}
1 - \ell s, & 0\le s\le 2/\ell,\\
-3 + \ell s, & 2/\ell\le s\le 4/\ell,
\end{cases}
\qquad
f(s) := \int_0^s v_0(z)\ud z,
\qquad
s_1 := 0.
\]
At the joining points, $v_0(0)=1$, $v_0(2/\ell)=-1$ from both sides, and $v_0(4/\ell)=1=v_0(0)$, so $v_0$ is continuous everywhere with slope $\pm\ell$, hence $\ell$-Lipschitz ($\ell\le 1$). Its integral over one period vanishes:
\[
\int_0^{2/\ell}(1-\ell z)\ud z + \int_{2/\ell}^{4/\ell}(-3+\ell z)\ud z
= \brac{\frac{2}{\ell}-\frac{2}{\ell}} + \brac{-\frac{6}{\ell}+\frac{8-2}{\ell}}
= 0.
\]
Thus $f$ is $(4/\ell)$-periodic and $1$-smooth, with $f(0)=0$. On $[0,4/\ell]$, $f$ increases on $[0,1/\ell]$ to $\frac{1}{2\ell}$, decreases on $[1/\ell,3/\ell]$ to
\[
f(3/\ell)
= \int_{2/\ell}^{3/\ell}(-3+\ell z)\ud z
= \sqbrac{-3z + \frac{\ell}{2}z^2}_{2/\ell}^{3/\ell}
= -\frac{1}{2\ell},
\]
and increases back to $0$ on $[3/\ell,4/\ell]$. Hence $\inf f = -\frac{1}{2\ell}$ and $f(0)-\inf f = \frac{1}{2\ell}\le 1$.

Because $\eta = j\cdot(4/\ell)$ is an integer multiple of the period, we verify by induction that $s_t\in(4/\ell)\mathbb Z$ and $g_t=m_t=1$ for all $t\ge 1$. At $t=1$, $s_1=0\in(4/\ell)\mathbb Z$, so $g_1=v_0(0)=1$ and $m_1=g_1=1$. Assuming the claim at step $t$, the update gives $s_{t+1}=s_t-\eta m_t=s_t-j(4/\ell)\in(4/\ell)\mathbb Z$, so periodicity yields $g_{t+1}=v_0(s_{t+1})=v_0(0)=1$ and $m_{t+1}=\beta m_t+(1-\beta)g_{t+1}=1$. Embedding $f$ into the first coordinate of $\R^d$ while keeping the remaining $d-1$ coordinates constant maintains $\norm{\nabla f(x_t)}_1=1$ for all $t\ge 1$.

\paragraph{3. Combining both regimes and lifting to $\R^{m\times n}$.}
When $\eta<4$, \eqref{eq:direct-huber-risk} yields $\min_{t\le T}\norm{\nabla f(x_t)}_1 \ge \frac12\min\{1,\sqrt{d/(4T)}\}\ge \frac14\min\{1,\sqrt{d/T}\}$. When $\eta\ge 4$, the periodic instance yields $1\ge \frac14\min\{1,\sqrt{d/T}\}$.

For either construction take $d=m$, $D(x)=\Diag{x}\PB$, and $X_1=D(x_1)$. This is the diagonal lift of \cref{lem:matrix-diagonal-embedding} with $\QB=\Ib_m$, so $\A(X)=D^*(X)$ extracts the leading diagonal. Its geometry gives the same initial gap, unit operator-norm smoothness, and equality of the nuclear and vector $\ell_1$ gradient norms.

Here the matrix step uses the momentum itself. Exact gradients give $M_1=G_1=D(m_1)$. If $X_t=D(x_t)$ and $M_t=D(m_t)$, then the update and the exact sample at the next query give
\begin{align*}
 X_{t+1}&=D(x_t)-\eta D(m_t)=D(x_{t+1}),\\
 M_{t+1}&=\beta D(m_t)+(1-\beta)D(g_{t+1})=D(m_{t+1}).
\end{align*}
Thus every iterate stays in the diagonal subspace, the oracle has zero noise, and
\[
 \min_{t\le T}\norm{\nabla F(X_t)}_*=\min_{t\le T}\norm{\nabla f(x_t)}_1
 \ge \frac14\min\bbrac{1,\sqrt{\frac{m}{T}}}.
\]
\end{proof}

\begin{proof}[Proof of \cref{lem:direct-signal-derivatives}]
Fix any point $s\in\R^m$ lying off the saturation hyperplanes $\{s_i=\pm R\}$, and let $U:=\{i\in[m]:\abs{s_i}<R\}$ be its set of unsaturated coordinates. On the open cell containing $s$, the sign factor $c_U:=\prod_{i\notin U}\sign{s_i}\in\{\pm 1\}$ is constant and $\Psi(s) = c_U\prod_{i\in U}\sin(\kappa s_i)$. Introducing i.i.d.\ Rademacher random variables $\{\varepsilon_i\}_{i\in U}\subset\{\pm 1\}$ and applying Euler's identity $\sin\theta = \frac{e^{\mathrm{i}\theta}-e^{-\mathrm{i}\theta}}{2\mathrm{i}} = \E_{\varepsilon_i}[-\mathrm{i}\varepsilon_i e^{\mathrm{i}\varepsilon_i\theta}]$ to each factor yields
\[
\Psi(s)
= c_U\prod_{i\in U}\E_{\varepsilon_i}\sqbrac{-\mathrm{i}\varepsilon_i e^{\mathrm{i}\kappa\varepsilon_i s_i}}
= \E_\varepsilon\sqbrac{w(\varepsilon)\exp\brac{\mathrm{i}\kappa\sum_{i\in U}\varepsilon_i s_i}},
\]
where $w(\varepsilon):=c_U\prod_{i\in U}(-\mathrm{i}\varepsilon_i)$ satisfies $\abs{w(\varepsilon)}=1$ and does not depend on $s$. Since this expectation is a finite sum over $2^{\abs{U}}$ outcomes, differentiating along any direction $a\in\R^m$ passes inside the expectation:
\begin{align*}
\inner{\nabla\Psi(s)}{a}
&= \E_\varepsilon\sqbrac{w(\varepsilon)\brac{\mathrm{i}\kappa\sum_{i\in U}\varepsilon_i a_i}\exp\brac{\mathrm{i}\kappa\sum_{i\in U}\varepsilon_i s_i}}, \\
\inner{\nabla^2\Psi(s)a}{a}
&= \E_\varepsilon\sqbrac{w(\varepsilon)\brac{\mathrm{i}\kappa\sum_{i\in U}\varepsilon_i a_i}^2\exp\brac{\mathrm{i}\kappa\sum_{i\in U}\varepsilon_i s_i}}.
\end{align*}
Using $\abs{w(\varepsilon)}=1$, $\abs{\exp(\mathrm{i}\kappa\sum_{i\in U}\varepsilon_i s_i)}=1$, Cauchy--Schwarz, and $\E_\varepsilon[(\sum_{i\in U}\varepsilon_i a_i)^2]=\sum_{i\in U}a_i^2\le\norm{a}_2^2$, we obtain
\begin{align*}
\abs{\inner{\nabla\Phi_1(s)}{a}}
&\le \kappa\,\E_\varepsilon\abs{\sum_{i\in U}\varepsilon_i a_i}
\le \kappa\brac{\E_\varepsilon\brac{\sum_{i\in U}\varepsilon_i a_i}^2}^{1/2}
\le \kappa\norm{a}_2, \\
\abs{\inner{\nabla^2\Phi_1(s)a}{a}}
&\le \kappa^2\,\E_\varepsilon\brac{\sum_{i\in U}\varepsilon_i a_i}^2
\le \kappa^2\norm{a}_2^2.
\end{align*}
Hence $\norm{\nabla\Phi_1(s)}_2\le\kappa$ and $\norm{\nabla^2\Phi_1(s)}_{\op}\le\kappa^2$ on every open cell. Because $\psi\in C^1(\R)$ with $\psi'(\pm 1)=0$, $\nabla\Phi_1$ is continuous across all saturation hyperplanes $\{s_i=\pm R\}$. Integrating the piecewise operator-norm Hessian bound along any line segment $[s,s']$ not contained in a joining hyperplane (which therefore crosses at most $2m$ hyperplanes) yields $\norm{\nabla\Phi_1(s')-\nabla\Phi_1(s)}_2\le\kappa^2\norm{s'-s}_2$; for any segment contained in a joining hyperplane, translating both endpoints by a vanishing generic perturbation and passing to the limit by continuity of $\nabla\Phi_1$ extends $\operatorname{Lip}(\nabla\Phi_1)\le\kappa^2$ globally.

The second signal satisfies $\nabla\Phi_2=-\Psi\nabla\Psi$ on all of $\R^m$, so $\norm{\nabla\Phi_2}_2\le\kappa$. Since $\abs{\Psi}\le 1$ and $\Psi$ is $\kappa$-Lipschitz, the global bound for $\nabla\Psi$ gives
\begin{align*}
 \norm{\nabla\Phi_2(s')-\nabla\Phi_2(s)}_2
 &\le \abs{\Psi(s')}\norm{\nabla\Psi(s')-\nabla\Psi(s)}_2
     +\abs{\Psi(s')-\Psi(s)}\norm{\nabla\Psi(s)}_2\\
 &\le 2\kappa^2\norm{s'-s}_2,
\end{align*}
which suffices for \eqref{eq:direct-signal-derivatives}.
\end{proof}

\begin{proof}[Proof of \cref{lem:direct-global-geometry}]
We verify parts~1 and~2 in turn.

\paragraph{1. Value bounds and initial gap.}
Since $\abs{\vartheta}\le 3/2$ and $\abs{\Phi_1(s)}+\Phi_2(s)\le 1$, the coupling term in \eqref{eq:direct-Q} has magnitude at most
\[
A\abs{\vartheta(w_1/B)\Phi_1(s) + \vartheta(w_2/B)\Phi_2(s)}
\le \frac{3A}{2}\brac{\abs{\Phi_1(s)} + \Phi_2(s)}
\le \frac{3A}{2}
< 3A = \frac{3E}{C_A}.
\]
Combining this with $0\le H(s)/E\le\frac{12K^2R^2p}{\eta E}\le\frac{1}{128}$ from \eqref{eq:direct-H-bounds} and \eqref{eq:direct-geometry-condition} gives
\[
-\frac{3}{C_A} \le \frac{Q(s,w_1,w_2)}{E} \le \frac{1}{128} + \frac{3}{C_A}.
\]
When $E=1$ and $\bar X_1$ has $w_1=w_2=0$, we have $\vartheta(0)=0$, so $Q(\bar X_1)=H(KR\mathbf{1})\le\frac{1}{128}$, while $\inf Q \ge -\frac{3}{C_A}=-\frac{3}{1024}$. Hence $Q(\bar X_1)-\inf Q \le \frac{1}{128}+\frac{3}{1024}=\frac{11}{1024}<1$.

\paragraph{2. Gradient and Hessian bounds.}
Differentiating \eqref{eq:direct-Q} with respect to $s$ and $w_j$ ($j=1,2$) and applying \eqref{eq:direct-signal-derivatives}, \eqref{eq:direct-H-bounds}, $\abs{\vartheta}\le 3/2$, $\abs{\vartheta'}\le 1$, $\abs{\Phi_j}\le 1$, and $A/B=u$ yields
\begin{align*}
\norm{\nabla_s Q}_2
&\le \norm{\nabla H(s)}_2 + A\abs{\vartheta(w_1/B)}\norm{\nabla\Phi_1(s)}_2  + A\abs{\vartheta(w_2/B)}\norm{\nabla\Phi_2(s)}_2
\le h_g + 3A\kappa, \\
\abs{Q_{w_j}}
&= \abs{u\,\vartheta'(w_j/B)\Phi_j(s)}
\le u.
\end{align*}
Next, we bound each Hessian block of $Q$ on any smooth open cell (away from the joining surfaces $\abs{s_i}\in\{R,2KR,3KR\}$ and $\abs{w_j}\in\{B,2B\}$):
\begin{itemize}
    \item \emph{Column block $\nabla^2_{ss} Q$.} Using $\norm{\nabla^2 H}_{\op}\le h_2$, $\abs{\vartheta}\le 3/2$, $\norm{\nabla^2\Phi_1}_{\op}\le\kappa^2$, and $\norm{\nabla^2\Phi_2}_{\op}\le 3\kappa^2$,
    \[
    \norm{\nabla^2_{ss} Q}_{\op}
    \le h_2 + A\brac{\frac{3}{2}\kappa^2 + \frac{3}{2}\cdot 3\kappa^2}
    = h_2 + 6A\kappa^2.
    \]
    \item \emph{Cross blocks $\nabla_s Q_{w_j}$ ($j=1,2$).} Since $\nabla_s Q_{w_j}=u\vartheta'(w_j/B)\nabla\Phi_j(s)$, we have $\norm{\nabla_s Q_{w_j}}_2\le u\kappa$.
    \item \emph{Diagonal blocks $Q_{w_j w_l}$ ($j,l\in\{1,2\}$).} For $j\neq l$, $Q_{w_j w_l}=0$. For $j=l$, $\abs{\vartheta''}\le 1$, $\abs{\Phi_j}\le 1$, and $u/B=u^2/A$ give
    \[
    \abs{Q_{w_j w_j}} = \abs{\frac{u}{B}\vartheta''(w_j/B)\Phi_j(s)} \le \frac{u^2}{A}.
    \]
\end{itemize}
Because every first derivative of $Q$ is continuous across all joining surfaces, integrating the piecewise Hessian bounds along any line segment connecting $(s,w_1,w_2)$ and $(s',w'_1,w'_2)$ that is not contained in a joining hyperplane (and passing to the limit via a vanishing generic translation for segments contained in a joining hyperplane, with $\Delta s:=s'-s$ and $\Delta w_j:=w'_j-w_j$) gives
\begin{align*}
\norm{\nabla_s Q' - \nabla_s Q}_2
&\le (h_2 + 6A\kappa^2)\norm{\Delta s}_2 + u\kappa(\abs{\Delta w_1}+\abs{\Delta w_2}), \\
\abs{Q'_{w_j} - Q_{w_j}}
&\le u\kappa\norm{\Delta s}_2 + \frac{u^2}{A}\abs{\Delta w_j} \qquad (j=1,2).
\end{align*}
Summing these three inequalities bounds $\norm{\nabla_s Q'-\nabla_s Q}_2+\abs{Q'_{w_1}-Q_{w_1}}+\abs{Q'_{w_2}-Q_{w_2}}$ by
\[
\brac{h_2 + 6A\kappa^2 + 4u\kappa + \frac{2u^2}{A}} \max\bbrac{\norm{\Delta s}_2,\, \abs{\Delta w_1},\, \abs{\Delta w_2}}.
\]
On $\R^m\times\R\times\R$, this immediately gives $\operatorname{Lip}(\nabla Q)\le h_2+6A\kappa^2+4u\kappa+2u^2/A$. On $\R^{m\times n}$, because the extraction maps in \eqref{eq:direct-coordinate-maps} are operator-norm contractions and their adjoints have nuclear-norm bound $1$, the same sum bounds $\norm{\nabla_X F(X')-\nabla_X F(X)}_*/\norm{X'-X}_{\op}$.

Finally, we verify the numerical bound \eqref{eq:direct-hessian-bound}. Set $\delta:=R^2p/(\eta E)$, which satisfies $\delta\le(1536K^2)^{-1}$ by \eqref{eq:direct-geometry-condition}. Using $E\ge 1$, $R\ge R_0$, $E/R^2\le R_0^{-2}$, $u\le 2^{-12}$, the constants in \eqref{eq:direct-constants}, and $\pi<22/7$, each term in \eqref{eq:direct-hessian-bound} is bounded explicitly:
\begin{align*}
h_2 &= 10K\delta\frac{E}{R^2} \le \frac{10}{1536 K R_0^2} < 10^{-7}, \\
6A\kappa^2 &= \frac{3\pi^2}{2C_A}\frac{E}{R^2} < \frac{15}{C_A R_0^2} < 2\times 10^{-6}, \\
4u\kappa &= \frac{2\pi u}{R} < \frac{7\cdot 2^{-12}}{R_0} < 2\times 10^{-5}, \\
\frac{2u^2}{A} &= \frac{2C_A u^2}{E} \le 2\cdot 2^{10}\cdot 2^{-24} = 2^{-13} < 2\times 10^{-4}, \\
\frac{2h_g}{E} &= \frac{8K\delta}{R} \le \frac{8}{1536 K R_0} < 4\times 10^{-6}, \\
\frac{6A\kappa}{E} &= \frac{3\pi}{C_A R} < \frac{10}{C_A R_0} < 10^{-4}, \\
\frac{4u}{E} &\le 4\cdot 2^{-12} = 2^{-10} < 10^{-3}.
\end{align*}
Summing all seven bounds yields less than $1.4\times 10^{-3} < \frac{1}{200}$, proving \eqref{eq:direct-hessian-bound}. Since $\frac{1}{200}<1$, the matrix objective $F=Q$ is globally $1$-smooth in the operator norm when $E=1$.
\end{proof}

\begin{proof}[Proof of \cref{lem:direct-circular}]
We construct the objective and oracle on the first two columns of $\R^{m\times n}$ and verify all admissibility and risk conditions in four steps.

\paragraph{1. Construction of objective and oracle.}
Set $\lambda:=1/16$ and define $F(X):=\frac{\lambda}{2}\norm{X_{:,1:2}}_F^2$, with row vectors $x_{t,i}:=(X_{t,i,1},X_{t,i,2})^\top\in\R^2$ for $i\in[m]$. Initialize $X_1$ with $x_{1,i}:=R_i e_1\in\R^2$ (and all entries in columns $3,\ldots,n$ equal to zero), where
\[
R_i^2 := \frac{\sigma_i^2}{8\lambda}\min\{\eta,\, V^{-1}\} \qquad (i\in[m]).
\]
Define the rotation angle $\theta\in(0,\pi/2)$ by
\[
\cos\theta := 1 - \frac{\eta(1-\beta)\lambda}{1+\beta},
\]
which lies in $(3/4,1)$ since $0<\eta<4$, $\lambda=1/16$, and $0<\frac{1-\beta}{1+\beta}\le 1$. Let $\mathsf R_\theta$ denote counterclockwise rotation by $\theta$ in $\R^2$.

Columns $3,\ldots,n$ always receive exact zero gradients, as do inactive rows with $\sigma_i=0$ (for which $R_i=0$ and $x_{1,i}=0$). Independently across active rows $i$ ($\sigma_i>0$), conditioned on the history $\F_{t-1}$, the oracle operates as follows:
\begin{itemize}
    \item At $t=1$, it returns $g_{1,i}^+:=(x_{1,i}-\mathsf R_\theta x_{1,i})/\eta$ with probability $1/2$ (the \emph{plus branch}), and $g_{1,i}^-:=2\lambda x_{1,i}-g_{1,i}^+$ with probability $1/2$ (the \emph{minus branch}). On the minus branch, row $i$ switches to exact gradients for all $t\ge 2$.
    \item On the plus branch, $m_{1,i}=g_{1,i}^+$ gives $x_{2,i}=x_{1,i}-\eta g_{1,i}^+=\mathsf R_\theta x_{1,i}$, which lies on the circle of radius $R_i$. Set $\delta_1:=+1$. At every query $t\ge 2$, as long as the row history matches the prescribed trajectory, the oracle draws $\delta_t\in\{+1,-1\}$ equal to $\delta_{t-1}$ with probability $(1+\beta)/2$ and $-\delta_{t-1}$ with probability $(1-\beta)/2$, sets $x_{t+1,i}:=\mathsf R_{\delta_t\theta}x_{t,i}$, and outputs
    \[
    g_{t,i} := \frac{(1+\beta)x_{t,i} - \beta x_{t-1,i} - x_{t+1,i}}{\eta(1-\beta)}.
    \]
    Because $m_{t-1,i}=(x_{t-1,i}-x_{t,i})/\eta$ and $m_{t,i}=\beta m_{t-1,i}+(1-\beta)g_{t,i}$, the \SGDMRef{} update reproduces the prescribed point exactly:
    \[
    x_{t,i}-\eta m_{t,i}
    = (1+\beta)x_{t,i} - \beta x_{t-1,i} - \eta(1-\beta)g_{t,i}
    = x_{t+1,i}.
    \]
    On any off-path history where $x_{t,i}$ deviates from this recursion, row $i$ defaults permanently to exact gradients.
\end{itemize}

\paragraph{2. Verification of smoothness and initial gap.}
The gradient $\nabla F(X)$ equals $\lambda X_{:,1:2}$ on the first two columns and zero elsewhere, so $\rank(\nabla F(U)-\nabla F(U'))\le 2$. Hence
\begin{align*}
\norm{\nabla F(U)-\nabla F(U')}_*
&\le 2\norm{\nabla F(U)-\nabla F(U')}_{\op}
 = 2\lambda\norm{(U-U')_{:,1:2}}_{\op} \\
&\le 2\lambda\norm{U-U'}_{\op}
 = \frac18\norm{U-U'}_{\op},
\end{align*}
which proves operator-norm smoothness with $L_*\le 1$. The objective is bounded below by $\inf F=0$ and satisfies
\[
F(X_1)
= \frac{\lambda}{2}\sum_{i=1}^m R_i^2
= \frac{1}{16}\min\{\eta,V^{-1}\}\sum_{i=1}^m\sigma_i^2
= \frac{1}{16}\min\{\eta V,1\}
\le \frac{1}{16} < 1.
\]

\paragraph{3. Verification of unbiasedness and PSD covariance.}
At $t=1$, $\E[g_{1,i}]=\frac12 g_{1,i}^+ + \frac12(2\lambda x_{1,i}-g_{1,i}^+)=\lambda x_{1,i}=\nabla_{x_i}F(X_1)$. Because $g_{1,i}-\lambda x_{1,i}=\pm(g_{1,i}^+-\lambda x_{1,i})$, expanding with $\norm{x_{1,i}}_2=R_i$, $\inner{x_{1,i}}{\mathsf R_\theta x_{1,i}}=R_i^2\cos\theta$, and $1-\cos\theta=\frac{\eta(1-\beta)\lambda}{1+\beta}\le\eta\lambda$ gives
\begin{align*}
\E\sqbrac{\norm{g_{1,i}-\lambda x_{1,i}}_2^2}
&= \norm{\frac{x_{1,i}-\mathsf R_\theta x_{1,i}}{\eta}-\lambda x_{1,i}}_2^2 \\
&= R_i^2\sqbrac{\frac{2(1-\cos\theta)}{\eta^2} - \frac{2\lambda(1-\cos\theta)}{\eta} + \lambda^2} \\
&\le R_i^2\brac{\frac{2\lambda}{\eta} + \lambda^2}
\le \frac{3\lambda R_i^2}{\eta}
= \frac{3\sigma_i^2}{8\eta}\min\{\eta,V^{-1}\}
\le \frac{3\sigma_i^2}{8},
\end{align*}
where the second-to-last inequality uses $\eta\lambda<4/16<1$.

At $t\ge 2$ on the plus branch, we have $x_{t-1,i}=\mathsf R_{-\delta_{t-1}\theta}x_{t,i}$, while
\[
\E\sqbrac{x_{t+1,i}\mid\F_{t-1}}
= \frac{1+\beta}{2}\mathsf R_{\delta_{t-1}\theta}x_{t,i}
  + \frac{1-\beta}{2}\mathsf R_{-\delta_{t-1}\theta}x_{t,i}.
\]
Using $\beta+\frac{1-\beta}{2}=\frac{1+\beta}{2}$ and the rotation identity $\mathsf R_\theta+\mathsf R_{-\theta}=(2\cos\theta)I_2$, we obtain
\[
\beta x_{t-1,i} + \E\sqbrac{x_{t+1,i}\mid\F_{t-1}}
= \frac{1+\beta}{2}\brac{\mathsf R_{\delta_{t-1}\theta}+\mathsf R_{-\delta_{t-1}\theta}}x_{t,i}
= (1+\beta)(\cos\theta)x_{t,i},
\]
and therefore
\[
\E\sqbrac{g_{t,i}\mid\F_{t-1}}
= \frac{(1+\beta)(1-\cos\theta)}{\eta(1-\beta)}x_{t,i}
= \lambda x_{t,i}
= \nabla_{x_i}F(X_t).
\]
For the conditional variance at $t\ge 2$, notice that $x_{t+1,i}$ takes two values $u_{\pm}:=\mathsf R_{\pm\delta_{t-1}\theta}x_{t,i}$ with probabilities $q:=\frac{1+\beta}{2}$ and $1-q=\frac{1-\beta}{2}$. Since $\norm{u_+-u_-}_2^2=(2\sin\theta)^2 R_i^2 = 4(1-\cos\theta)(2-(1-\cos\theta))R_i^2$, we obtain
\begin{align*}
\E\sqbrac{\norm{g_{t,i}-\lambda x_{t,i}}_2^2\mid\F_{t-1}}
&= \frac{q(1-q)}{\eta^2(1-\beta)^2}\norm{u_+-u_-}_2^2 \\
&= \frac{1+\beta}{\eta^2(1-\beta)}\cdot(1-\cos\theta)\brac{2-(1-\cos\theta)}R_i^2 \\
&= R_i^2\brac{\frac{2\lambda}{\eta} - \frac{(1-\beta)\lambda^2}{1+\beta}}
\le \frac{2\lambda R_i^2}{\eta}
\le \frac{\sigma_i^2}{4}.
\end{align*}
For $t\ge 2$, the noise is zero on every inactive, minus-branch, or off-path row (at $t=1$, inactive rows have zero noise while active rows were already bounded above). Because the row-wise noise vectors are conditionally independent and zero-mean given $\F_{t-1}$, all off-diagonal entries of $\E[\Nb_t\Nb_t^\top\mid\F_{t-1}]$ vanish and its $i$-th diagonal entry is at most $\sigma_i^2$, so $\E[\Nb_t\Nb_t^\top\mid\F_{t-1}]\preceq\vSigma^2$.

\paragraph{4. Trajectory norm and gradient lower bound.}
Define $a_i:=R_i/\brac{\sum_{j=1}^m R_j^2}^{1/2}$ so that $\sum_{i=1}^m a_i^2=1$. For every $t\in[T]$, applying $\norm{\cdot}_*\ge\norm{\cdot}_F$ and Cauchy--Schwarz gives
\[
\norm{\nabla F(X_t)}_*
\ge \norm{\nabla F(X_t)}_F
= \lambda\brac{\sum_{i=1}^m \norm{x_{t,i}}_2^2}^{1/2}
\ge \lambda\sum_{i=1}^m a_i\norm{x_{t,i}}_2,
\]
and taking the minimum over $t\in[T]$ yields
\[
\min_{t\le T}\norm{\nabla F(X_t)}_*
\ge \lambda\sum_{i=1}^m a_i\min_{t\le T}\norm{x_{t,i}}_2.
\]
For each active row $i$, the plus branch occurs at $t=1$ with probability $1/2$ and keeps $\norm{x_{t,i}}_2=R_i$ for all $t\ge 1$, so $\E[\min_{t\le T}\norm{x_{t,i}}_2]\ge\frac12 R_i$. Taking expectations and substituting $\sum_{i=1}^m R_i^2 = \frac{1}{8\lambda}\min\{\eta V,1\} = 2\min\{\eta V,1\}$ gives
\begin{align*}
\expect{\min_{t\le T}\norm{\nabla F(X_t)}_*}
&\ge \frac{\lambda}{2}\sum_{i=1}^m a_i R_i
 = \frac{\lambda}{2}\brac{\sum_{i=1}^m R_i^2}^{1/2} \\
&= \frac{\sqrt{2}}{32}\min\bbrac{1,\,\sqrt{\eta V}}
 \ge \frac{1}{32}\min\bbrac{1,\,\sqrt{\eta V}}.
\end{align*}
\end{proof}

\section{Analysis for SGD with Momentum}
\label{app:direct-matrix-sgdm}

\Cref{app:direct-sgdm-upper} proves the \SGDMRef{} upper bound by Lyapunov descent on a virtual iterate. \Cref{app:sgdm-lower-bound,app:sgdm-complexity} give the lower bound and its complexity consequence. The supporting proofs are in \cref{app:sgdm-supporting-lemmas}.

We use the \SGDMRef{} updates in \cref{alg:matrix-optimizers}: $\MB_1=\Gb_1$, $\MB_t=\beta\MB_{t-1}+\alpha\Gb_t$ for $t\ge2$, and $\WB_{t+1}=\WB_t-\eta\MB_t$, with $\alpha:=1-\beta\in(0,1]$. Write
\[
 A_0:=L_*\Delta, \qquad V:=\norm{\vSigma}_F^2, \qquad v:=\frac{V}{A_0}, \qquad \ell_m:=\log(e+m),
\]
and recall the lower-bound branches $b_T,d_T,s_T$ in \eqref{eq:direct-sgdm-branches}.

\subsection{Upper Bound for SGDM}
\label{app:direct-sgdm-upper}

\begin{proof}[Proof of \cref{thm:direct-sgdm-upper}]
As established in \cref{app:sgd-upper-cap}, \cref{ass:operator norm smooth} implies Frobenius smoothness with constant $L_*$, and \cref{ass:Matrix-variance} gives $\E[\norm{\Nb_t}_F^2\mid\F_{t-1}]\le\tr(\vSigma^2)=V$, where $\Nb_t:=\Gb_t-\nabla F(\WB_t)$. Fix $\beta=1/2$ and any step size $0<\eta\le 1/(4L_*)$.

\paragraph{Step 1: Virtual-iterate decomposition.}
Define $(Z_1,U_1):=(\WB_1,\vzero)$, and for every $t\ge 2$ set
\[
 U_t:=\eta\MB_{t-1}, \qquad Z_t:=\WB_t-U_t,
\]
so $\WB_t=Z_t+U_t$ for every $t$. Since $U_{t+1}=\eta\MB_t$, the update and the offset each subtract $\eta\MB_t$ from $\WB_t$ in forming $Z_{t+1}$. Using $\MB_1=\Gb_1$ at startup and the EMA recursion thereafter gives
\begin{align}
 (Z_2,U_2)&=(\WB_1-2\eta\Gb_1,\,\eta\Gb_1),\nonumber\\
 U_{t+1}&=\eta\MB_t=\frac12 U_t+\frac{\eta}{2}\Gb_t \qquad(t\ge 2),\nonumber\\
 Z_{t+1}&=\WB_t-2\eta\MB_t=(Z_t+U_t)-(U_t+\eta\Gb_t)=Z_t-\eta\Gb_t
 \qquad(t\ge 2).
 \label{eq:direct-sgdm-first-step}
\end{align}

\paragraph{Step 2: Direct conditional-expectation Lyapunov descent.}
Consider the nonnegative Lyapunov energy
\[
 \H_t := F(Z_t) - \inf F + L_*\norm{U_t}_F^2.
\]
By induction from linear gradient growth ($\norm{\nabla F(\WB)}_F\le\norm{\nabla F(\WB_1)}_F+L_*\norm{\WB-\WB_1}_F$) and the conditional variance bound, $\WB_t,Z_t,U_t,\Gb_t$ have finite second moments for every $t\ge 1$, so $\H_t$ is integrable by quadratic growth of $F$. Set $b:=\nabla F(\WB_t)$. Because $\E[\Nb_t\mid\F_{t-1}]=\vzero$ and $\E[\norm{\Nb_t}_F^2\mid\F_{t-1}]\le V$, we have
\[
 \E[\Gb_t\mid\F_{t-1}] = b,
 \qquad
 \E\sqbrac{\norm{\Gb_t}_F^2\mid\F_{t-1}}
 = \norm{b}_F^2 + \E\sqbrac{\norm{\Nb_t}_F^2\mid\F_{t-1}}
 \le \norm{b}_F^2 + V.
\]
For $t\ge 2$, smoothness at $Z_t$ bounds the change in $F$ along $-\eta\Gb_t$. Expanding the square of $U_{t+1}=(U_t+\eta\Gb_t)/2$ in the same step gives
\begin{align*}
 \E\sqbrac{\H_{t+1}-\H_t\mid\F_{t-1}}
 &\le -\eta\inner{\nabla F(Z_t)}{b}-\frac{3L_*}{4}\norm{U_t}_F^2
       +\frac{L_*\eta}{2}\inner{U_t}{b}
       +\frac{3L_*\eta^2}{4}\brac{\norm{b}_F^2+V}.
\end{align*}
Since $\WB_t-Z_t=U_t$, the gradient difference $\nabla F(Z_t)-b$ has Frobenius norm at most $L_*\norm{U_t}_F$. The two resulting mixed terms are absorbed by Young's inequality:
\[
 \frac{3L_*\eta}{2}\norm{U_t}_F\norm{b}_F
 \le \frac{3L_*}{4}\norm{U_t}_F^2+\frac{3L_*\eta^2}{4}\norm{b}_F^2.
\]
The first term cancels the negative momentum energy, leaving
\[
 \E\sqbrac{\H_{t+1}-\H_t\mid\F_{t-1}}
 \le -\eta\brac{1-\frac{3L_*\eta}{2}}\norm{b}_F^2+\frac{3L_*\eta^2}{4}V
 \qquad(t\ge 2).
\]
At $t=1$, \eqref{eq:direct-sgdm-first-step} gives $Z_2=Z_1-2\eta\Gb_1$ and $U_2=\eta\Gb_1$. Smoothness at $Z_1=\WB_1$ therefore gives
\begin{align*}
 \H_2-\H_1 &\le -2\eta\inner{b}{\Gb_1}+3L_*\eta^2\norm{\Gb_1}_F^2,\\
 \E\sqbrac{\H_2-\H_1\mid\F_0}
 &\le -\eta(2-3L_*\eta)\norm{b}_F^2+3L_*\eta^2 V.
\end{align*}
Since $L_*\eta\le 1/4$, both estimates imply
\[
 \E\sqbrac{\H_{t+1}-\H_t\mid\F_{t-1}}
 \le -\frac{\eta}{4}\norm{\nabla F(\WB_t)}_F^2+3L_*\eta^2 V
 \qquad(t\ge 1).
\]
Taking total expectations, rearranging, and averaging over $t=1,\ldots,T$ gives
\begin{equation}\label{eq:direct-sgdm-energy-rate}
 \frac1T\sum_{t=1}^T \E\norm{\nabla F(\WB_t)}_F^2
 \le \frac{4(\H_1-\E[\H_{T+1}])}{\eta T} + 12L_*\eta V
 \le \frac{4\Delta}{\eta T} + 12L_*\eta V,
\end{equation}
using $\H_1=F(\WB_1)-\inf F\le\Delta$ and $\H_{T+1}\ge 0$.

\paragraph{Step 3: Nuclear-norm conversion and step-size selection.}
Because $\rank(\nabla F(\WB_t))\le m$, Cauchy--Schwarz gives $\norm{\nabla F(\WB_t)}_*\le\sqrt{m}\norm{\nabla F(\WB_t)}_F$. Applying Jensen's inequality to \eqref{eq:direct-sgdm-energy-rate} yields
\[
 \frac1T\sum_{t=1}^T \E\norm{\nabla F(\WB_t)}_*
 \le \sqrt{m}\brac{\frac1T\sum_{t=1}^T \E\norm{\nabla F(\WB_t)}_F^2}^{1/2}
 \le \sqrt{m}\brac{\frac{4\Delta}{\eta T}+12L_*\eta V}^{1/2}.
\]
For the prescribed step size $\eta=\min\{\frac{1}{4L_*},\sqrt{\frac{\Delta}{5L_*VT}}\}$ when $V>0$ (and $\eta=\frac{1}{4L_*}$ when $V=0$), we have $0<\eta\le\frac{1}{4L_*}$ together with
\[
 \frac{1}{\eta} \le 4L_* + \sqrt{\frac{5L_*VT}{\Delta}},
 \qquad
 \eta V \le \sqrt{\frac{\Delta V}{5L_*T}},
\]
both of which also hold trivially when $V=0$. Substituting into the right-hand side of \eqref{eq:direct-sgdm-energy-rate} yields
\begin{align*}
 \frac{4\Delta}{\eta T} + 12L_*\eta V
 &\le \frac{16L_*\Delta}{T} + 4\sqrt{\frac{5L_*\Delta V}{T}} + \frac{12}{\sqrt{5}}\sqrt{\frac{L_*\Delta V}{T}} \\
 &= \frac{16A_0}{T} + \frac{32}{\sqrt{5}}\sqrt{\frac{A_0V}{T}}
  \le \frac{16A_0}{T} + 16\sqrt{\frac{A_0V}{T}},
\end{align*}
since $\sqrt{5}>2$ implies $32/\sqrt{5}<16$. Applying $\sqrt{x+y}\le\sqrt{x}+\sqrt{y}$ therefore gives
\begin{align*}
 \frac1T\sum_{t=1}^T \E\norm{\nabla F(\WB_t)}_*
 \le \sqrt{m}\brac{\frac{16A_0}{T} + 16\sqrt{\frac{A_0V}{T}}}^{1/2} \le 4\sqbrac{\sqrt{\frac{mA_0}{T}} + \brac{\frac{m^2A_0V}{T}}^{1/4}},
\end{align*}
proving \eqref{eq:direct-sgdm-upper} with explicit constant $C=4$.
\end{proof}

\subsection{Lower Bound for SGDM}
\label{app:sgdm-lower-bound}

Fix $\eta>0$ and $\beta\in[0,1)$. We first prove \cref{thm:direct-matrix-sgdm} for $L_*=\Delta=1$ and diagonal $\vSigma=\Diag{\sigma_1,\ldots,\sigma_m}$, so $V=v=\sum_{i=1}^m\sigma_i^2$. Momentum changes the vanilla \SGDRef{} constructions in two places. Initial momentum persists for order $1/\alpha$ steps; \cref{app:direct-inertia} uses this time to obtain a lower bound that depends on $\alpha$. After a column pulse, residual velocity can delay the sign change; \cref{app:direct-reset-oracle} tracks a shadow iterate and allows the physical column to catch up between pulses. \Cref{app:direct-sgdm-interpolation} combines these constructions with the vanilla instances and restores the general scales.

\subsubsection{Deterministic Inertia Obstruction under Strong Momentum}
\label{app:direct-inertia}

For $0<\eta<4$ and $\alpha<\eta/128$, the Huber bound in \cref{lem:direct-step-obstructions} does not depend on $\alpha$. We use a scalar driver $\mathsf V$ that carries an auxiliary coordinate onto a flat plateau. A guard $J$ keeps the gradient large during the first few steps. The same driver and guard will be used in the matrix pulse construction.

Fix $\ell:=7/16$ and $a:=1/\ell=16/7$. For any plateau location $D>4a$ with plateau height $E:=D-2a$, define the piecewise-quadratic driver
\begin{equation}\label{eq:direct-driver}
 \mathsf V(z) := \begin{cases}
  a, & z \le -a, \\
  a - \frac{\ell}{2}(z+a)^2, & -a \le z \le 0, \\
  \frac{\ell}{2}(z-a)^2, & 0 \le z \le 2a, \\
  z - \frac{3a}{2}, & 2a \le z \le D-a, \\
  E - \frac{\ell}{2}(D-z)^2, & D-a \le z \le D, \\
  E, & z \ge D.
 \end{cases}
\end{equation}
For any target slope $u>0$, define the startup guard potential
\begin{equation}\label{eq:direct-guard-potential}
 J(\varrho) := \begin{cases}
  0, & v \le 0, \\
  \frac{\ell}{2}v^2, & 0 \le v \le u/\ell, \\
  uv - \frac{u^2}{2\ell}, & v \ge u/\ell.
 \end{cases}
\end{equation}
Direct inspection at each joining point (using $\ell a=1$ and $E=D-2a>2a>a$) verifies three properties:
\begin{itemize}
    \item \textbf{Value range:} $0\le\mathsf V(z)\le E$ for all $z\in\R$, with $\mathsf V(0)=a/2$, $\mathsf V(a)=0$, and $\mathsf V(z)=E$ on the plateau $z\ge D$; similarly $J(\varrho)\ge 0$ with $J(\varrho)=0$ for $\varrho\le 0$.
    \item \textbf{First derivatives:} Both $\mathsf V$ and $J$ are $C^1$ globally, with $\abs{\mathsf V'(z)}\le 1$ everywhere, $\mathsf V'(0)=-1$, $\mathsf V'(z)=1$ on $[2a,D-a]$, and $\mathsf V'(z)=0$ on $(-\infty,-a]\cup[D,\infty)$; for the guard, $0\le J'(v)\le u$ everywhere and $J'(v)=u$ for $v\ge u/\ell$.
    \item \textbf{Smoothness:} On each smooth piece, $\abs{\mathsf V''(z)}\le\ell$ and $0\le J''(v)\le\ell$, so $\operatorname{Lip}(\mathsf V')\le\ell$ and $\operatorname{Lip}(J')\le\ell$.
\end{itemize}

\paragraph{Time to the plateau.}
Both constructions use a driver coordinate $z_1=0$ with update $z_{t+1}=z_t-\eta m_{t,z}$ and plateau at $D=\eta/(14\alpha)>2a$. Suppose its samples satisfy $m_{1,z}=g_{1,z}\le-3/4$, $g_{t,z}\le 1$ at every step, and $g_{t,z}=0$ whenever $z_t\ge D$. Expanding the momentum recursion gives the pathwise bound
\[
 m_{t,z}
 =\beta^{t-1}m_{1,z}+\alpha\sum_{j=2}^t\beta^{t-j}g_{j,z}
 \le -\frac34\beta^{t-1}+1-\beta^{t-1}
 =1-\frac74\beta^{t-1}.
\]
Let $K_{\mathrm{acc}}=\lceil1/(7\alpha)\rceil$. For $1\le t\le K_{\mathrm{acc}}$, Bernoulli's inequality gives $\beta^{t-1}\ge 1-\alpha(t-1)>6/7$. Hence
\[
 m_{t,z}<-1/2\quad(1\le t\le K_{\mathrm{acc}}),\qquad
 z_t>\frac{\eta}{2}(t-1)\quad(2\le t\le K_{\mathrm{acc}}+1).
\]
Since $\eta K_{\mathrm{acc}}/2\ge D$, the first plateau query $\tau_D$ lies between $2$ and $K_{\mathrm{acc}}+1$. At entry $m_{\tau_D-1,z}<0$. The zero driver sample leaves $m_{\tau_D,z}=\beta m_{\tau_D-1,z}\le0$, so $z_{\tau_D+1}\ge z_{\tau_D}$; the same argument keeps the driver on the plateau thereafter. In particular, a query below $2a$ must occur before entry, and the displacement bound gives
\[
 z_t<2a\quad\Longrightarrow\quad t-1<\frac{4a}{\eta}=\frac{4}{\eta\ell}.
\]

\begin{lemma}[Inertia obstruction]\label{lem:direct-inertia}
If $0<\eta<4$ and $\alpha=1-\beta<\eta/128$, there exists a normalized deterministic $d$-dimensional instance ($d\ge 2$) with initialization $m_1=g_1$ satisfying
\begin{equation}\label{eq:direct-inertia-risk}
 \min_{t\le T}\norm{\nabla f(x_t)}_1
 \ge \frac{1}{32}\min\bbrac{1,\,\sqrt{\frac{d}{\alpha T}}}.
\end{equation}
\end{lemma}

\subsubsection{Shadow Recursion and Guarded Cooldown Pulses}
\label{app:direct-reset-oracle}

A vanilla \SGDRef{} pulse sends the selected coordinate directly to $-KR\sign{s_i}$ by \eqref{eq:direct-sgd-pulse}. For \SGDMRef{}, the pulse acts on the shadow coordinate defined below; the physical coordinate retains velocity and may still have its old sign.

\paragraph{Shadow-iterate recursion.}
For $t\ge 2$, define the momentum-free \emph{shadow iterate} $q_t\in\R^m$ from the control column $s_t=X_{t,:,1}$ and its momentum $m_{t-1,s}$ by $q_t:=s_t-\frac{\eta\beta}{\alpha}m_{t-1,s}$. Using $s_{t+1}=s_t-\eta m_{t,s}$ and $m_{t,s}=\beta m_{t-1,s}+\alpha g_{t,s}$ for $t\ge 2$, direct algebra gives
\begin{align*}
 q_{t+1}
 &= s_{t+1}-\frac{\eta\beta}{\alpha}m_{t,s}
  = s_t-\eta\brac{1+\frac{\beta}{\alpha}}m_{t,s}
  = s_t-\frac{\eta}{\alpha}m_{t,s} \\
 &= s_t-\frac{\eta}{\alpha}(\beta m_{t-1,s}+\alpha g_{t,s})
  = \brac{s_t-\frac{\eta\beta}{\alpha}m_{t-1,s}}-\eta g_{t,s}
  = q_t-\eta g_{t,s}.
\end{align*}
Solving $q_{t+1}=s_t-\frac{\eta}{\alpha}m_{t,s}$ for $-\eta m_{t,s}=\alpha(q_{t+1}-s_t)$ then yields $s_{t+1}=s_t-\eta m_{t,s}=\beta s_t+\alpha q_{t+1}$. Summarizing,
\begin{equation}\label{eq:direct-shadow}
 q_t:=s_t-\frac{\eta\beta}{\alpha}m_{t-1,s}, \qquad
 q_{t+1}=q_t-\eta g_{t,s}, \qquad
 s_{t+1}=\beta s_t+\alpha q_{t+1} \qquad (t\ge 2).
\end{equation}
Thus $q_t$ evolves by a memoryless gradient step while $s_t$ tracks $q_t$ via exponential smoothing. The tracking error $e_t:=q_t-s_t$ obeys the exact recursion
\[
 e_{t+1} = q_{t+1}-(\beta s_t+\alpha q_{t+1}) = \beta(q_{t+1}-s_t) = \beta(e_t-\eta g_{t,s}) \qquad (t\ge 2).
\]
A pulse in coordinate $i$ moves $q_{t+1,i}$ directly to the opposite center $-KR\sign{s_{t,i}}$ up to the small prior discrepancy $e_{t,i}$. Following each pulse by $N=\Theta(1/\alpha)$ exact-gradient cooldown queries lets $s_{t,i}$ relax onto $q_{t+1,i}$ while residual momentum decays.

\paragraph{Objective assembly in ordinary and strong-momentum regimes.}
Adopt the matrix geometry of \cref{app:direct-geometry} for $m\ge 4$, with constants $K=16$, $R_0=100$, $C_A=1024$ from \eqref{eq:direct-constants}, coordinate maps $(s,z,w_0,w_1,w_2)$ from \eqref{eq:direct-coordinate-maps} ($k_j=|I_j|$, $k=\min_j k_j\ge m/6$), signals $\Phi_1,\Phi_2$ from \eqref{eq:direct-signals}, and coupled potential $Q(s,w_1,w_2)$ from \eqref{eq:direct-Q}. Define the momentum ratio $r:=\eta/\alpha$, and choose the geometric scales $(E,R)$ with $A:=E/C_A$ and $B:=A/u$ according to two regimes:
\begin{itemize}
    \item \textbf{Ordinary regime ($r\le 128$).} Set $(E,R):=(1,R_0)$ and $F(X):=Q(s,w_1,w_2)$, initialized at $\bar X_1$ with $s_i=KR$ ($i\in[m]$) and $w_1=w_2=0$.
    \item \textbf{Strong-momentum regime ($r>128$).} With $\ell=7/16$ and $a=1/\ell=16/7$, set
    \[
     D:=\frac{r}{14}, \qquad E:=D-2a=\frac{r}{14}-\frac{32}{7}, \qquad R:=R_0 r.
    \]
    Using $\mathsf V(z)$ from \eqref{eq:direct-driver} (with parameters $D,E$) and $J$ from \eqref{eq:direct-guard-potential} (with slope $u$), define the normalized coupling $C$, the startup guard $h(w_0)$, and the composite objective $F$ by
    \begin{equation}\label{eq:direct-strong-objective}
    \begin{aligned}
     C(s,w_1,w_2) &:= \frac{25}{32}+\frac{Q(s,w_1,w_2)}{E},
     \qquad B_0:=\frac{12u}{k_0\ell},
     \qquad h(w_0):=J\!\brac{B_0+\frac{u}{\ell}-w_0}, \\
     F(X) &:= \mathsf V(z)\,C(s,w_1,w_2)+h(w_0),
    \end{aligned}
    \end{equation}
    initialized at $\bar X_1$ with $z=w_0=w_1=w_2=0$ and $s_i=KR$ ($i\in[m]$).
\end{itemize}
In the strong-momentum regime ($r>128$), we have $D>128/14=4a$, which gives $E>2a>1$, $E>\frac12 D=r/28$ (so $r/E<28$), $R/E=R_0(r/E)<28R_0$, $R=R_0 r>R_0$, and $E/R^2=E/(R_0^2 r^2)<1/(14rR_0^2)<R_0^{-2}$. Together with the ordinary regime ($E=1,R=R_0$), both regimes satisfy
\[
 E\ge 1, \qquad R\ge R_0, \qquad \frac{E}{R^2}\le R_0^{-2}, \qquad \frac{r}{E}\le 128, \qquad \frac{R}{E}\le 28R_0.
\]
Hence both regimes satisfy the scale hypotheses \eqref{eq:direct-scales}, so \cref{lem:direct-global-geometry} applies to $Q$ whenever \eqref{eq:direct-geometry-condition} holds. To certify that the strong-momentum composite objective \eqref{eq:direct-strong-objective} still obeys normalized initial gap and unit operator-norm smoothness when invoked inside \cref{lem:direct-protected}, the next lemma extends those geometric bounds to $F(X)$.

\begin{lemma}[Strong-momentum composite geometry]\label{lem:direct-strong-geometry}
Suppose $r=\eta/\alpha>128$, $(E,R)=(r/14-32/7,\,R_0 r)$, $A=E/C_A$, $B=A/u$, and \eqref{eq:direct-geometry-condition} holds. Then the composite objective $F$ in \eqref{eq:direct-strong-objective} satisfies $\frac34<C(s,w_1,w_2)<\frac{13}{16}$ everywhere, $\inf F=0$, $F(\bar X_1)-\inf F\le 1$, and global operator-norm smoothness with constant $L_*\le 1$.
\end{lemma}

\paragraph{Guarded cooldown oracle.}
We now specify the conditional stochastic oracle globally on every history. The oracle maintains an $\F_{t-1}$-measurable cooldown counter $c_t\in\{0,\ldots,N\}$ (initialized at $c_1=0$) and a permanent fallback flag (initially inactive). The first query ($t=1$) always returns the exact gradient; in the strong-momentum regime ($r>128$), the oracle continues returning exact gradients throughout the startup phase until the driver first reaches its plateau $z\ge D$. Once startup has ended, at any query $t\ge 2$ we call the query \emph{ready} if permanent fallback is inactive, the cooldown counter satisfies $c_t=0$, the driver is on its plateau (when $r>128$), the slow coordinates satisfy $\abs{w_1},\abs{w_2}\le B$, and every column coordinate $i\in[m]$ satisfies
\begin{equation}\label{eq:direct-ready-guard}
 \abs{\abs{q_i}-KR}\le R, \qquad
 \abs{q_i-s_i}\le \frac{R}{100}.
\end{equation}
Set the cooldown length $N$ and startup budget $J_0$ to
\begin{equation}\label{eq:direct-cooldown}
 N:=\left\lceil\frac{\log(1000K)}{\alpha}\right\rceil,\qquad
 J_0:=\left\lceil\frac{1}{7\alpha}\right\rceil+2.
\end{equation}
Since $K=16$ and $\alpha\le 1$, these lengths satisfy
\[
 \alpha N<\log(16000)+1<11,\qquad \alpha J_0<\frac17+3<4.
\]
At a ready query, the oracle outputs the column pulse sample \eqref{eq:direct-sgd-pulse} (with probabilities $p_i$, $p:=\sum_{i=1}^m p_i$) and exact partial derivatives in all other entries:
\begin{itemize}
    \item If a nonzero pulse occurs, the cooldown counter is set to $N$, so the next $N$ queries return exact gradients while decrementing the counter to $0$ without consuming a pulse draw.
    \item If a zero column sample occurs, the cooldown counter stays at $0$ and the next query immediately undergoes another readiness check.
    \item If a post-startup query $t$ with $c_t=0$ fails the geometric guard (or any inconsistent history occurs), the oracle activates permanent fallback, returns the exact gradient $\nabla F(X_t)$ at query $t$ without consuming a pulse draw, and returns exact gradients at all subsequent queries.
\end{itemize}

Every ready query satisfies $(K-1-1/100)R\le\abs{s_i}\le(K+1+1/100)R$, which lies strictly inside the saturated core $[R,2KR]$ because $K=16$. Therefore $\nabla_s\Phi_1(s)=\nabla_s\Phi_2(s)=\vzero$, and in the strong-momentum regime $\mathsf V(z)/E=1$ on the plateau, so \eqref{eq:direct-H-saturated} gives $F_{s_i}(X)=H_{s_i}(s)=\frac{p_i}{\eta}[s_i+KR\sign{s_i}]$. This matches the conditional mean of \eqref{eq:direct-sgd-pulse} and establishes conditional unbiasedness on every history. When a pulse occurs at coordinate $i$ at a ready query $t\ge 2$, substituting $g_{t,s,i}=[s_{t,i}+KR\sign{s_{t,i}}]/\eta$ into \eqref{eq:direct-shadow} produces the exact shadow reset
\begin{equation}\label{eq:direct-reset}
 q_{t+1,i}
 = s_{t,i}-\frac{\eta\beta}{\alpha}m_{t-1,s,i}-\brac{s_{t,i}+KR\sign{s_{t,i}}}
 = -KR\sign{s_{t,i}}+(q_{t,i}-s_{t,i}),
\end{equation}
whereas every non-pulsed coordinate $j\neq i$ has $g_{t,s,j}=0$, leaving $q_{t+1,j}=q_{t,j}$ unchanged.

Because \eqref{eq:direct-reset} resets coordinate $i$'s shadow to its new center $-KR\sign{s_{t,i}}$ up to its prior tracking error $e_{t,i}$, subsequent drift in coordinate $i$ needs to be tracked only back to coordinate $i$'s own most recent pulse. To control this cumulative drift inside \cref{lem:direct-cooldown-tracking}, the next lemma bounds the number of cooling blocks between consecutive own resets of any coordinate with high probability.

\begin{lemma}[Gaps between own resets]\label{lem:direct-mark-gaps}
Let $M\ge 1$ and let $I_1,\ldots,I_M$ be independent coordinate marks drawn with probabilities $\hat p_i:=p_i/p$. Put $\Lambda:=\log(64m(M+1))$. With probability at least $63/64$, every run of marks avoiding any active coordinate $i$ ($\hat p_i>0$), including initial and final runs, has length strictly less than $\lceil\Lambda/\hat p_i\rceil$. Consequently, between consecutive own resets of coordinate $i$ (or before its first reset), the total number of cooling blocks affecting coordinate $i$ (including its own preceding cooling block, if any) is at most $1+\Lambda/\hat p_i$.
\end{lemma}

Consider the cooldown blocks triggered by other coordinates after coordinate $i$'s last block, or from the first ready query if $i$ has not yet pulsed. While $R\le\abs{s_i}\le 2KR$, we have $\psi'(s_i/R)=0$, so both signal derivatives in coordinate $i$ vanish regardless of the other coordinates. By \eqref{eq:direct-H-saturated}, each exact step then shifts its shadow by at most $\eta\abs{H_{s_i}}\le 4KRp_i$. If $\nu_i$ such blocks have begun and $p_i>0$, the mark-gap event gives $\nu_i\le\Lambda/\hat p_i$. Their total shadow drift, including any partial final block, is bounded by
\begin{equation}\label{eq:direct-local-drift}
 4KRp_iN\nu_i\le 4KRpN\Lambda\le R/32,
\end{equation}
where the last inequality uses \eqref{eq:direct-reset-condition}. For $p_i=0$, coordinate $i$ is never selected and its drift while saturated is zero. A coordinate need not remain saturated during its own cooldown; the next lemma controls that block using the global gradient bound.

\begin{lemma}[Cooldown tracking invariant]\label{lem:direct-cooldown-tracking}
Assume \eqref{eq:direct-geometry-condition}, \eqref{eq:direct-pulse-smallness}, and \eqref{eq:direct-reset-condition} hold, and work on the mark-gap event of \cref{lem:direct-mark-gaps} with at most $M$ pulses. Suppose the driver (when $r>128$) is on its plateau $z\ge D$ and the slow coordinates remain in $[-B,B]$. Suppose that at the first ready query $t_{\mathrm{ready}}\ge 2$, every coordinate $j\in[m]$ satisfies $\abs{q_{t_{\mathrm{ready}},j}-KR}\le R/100$ and $\abs{e_{t_{\mathrm{ready}},j}}\le R/100$. Then:
\begin{enumerate}
    \item Readiness \eqref{eq:direct-ready-guard} holds after every zero-sample ready query and at the end of every $N$-step cooldown block.
    \item A pulse at ready query $t$ in coordinate $i$ resets its shadow to the opposite side. At the next ready query $t+N+1$, its physical coordinate is in the saturated core with the opposite sign. Every other coordinate keeps its sign and stays in the saturated core throughout the block.
\end{enumerate}
\end{lemma}

During cooldown, $\Phi_1(s)$ can vanish as a coordinate crosses zero. The complementary signal $\Phi_2(s)$ protects the gradient at these queries.

\begin{lemma}[Startup phase and uniform gradient protection]\label{lem:direct-startup-protection}
Assume \eqref{eq:direct-geometry-condition} and \eqref{eq:direct-pulse-smallness} hold.
\begin{enumerate}
    \item \textbf{Startup readiness.} If the slow coordinates remain in $[-B,B]$ during startup, then the first ready query occurs at some index $t_{\mathrm{ready}}\in\{2,\ldots,J_0\}$ (with $z_t\ge D$ for all $t\ge t_{\mathrm{ready}}$ when $r>128$), and every column coordinate $i\in[m]$ satisfies $\abs{q_{t_{\mathrm{ready}},i}-KR}\le R/100$ and $\abs{e_{t_{\mathrm{ready}},i}}\le R/100$.
    \item \textbf{Uniform gradient lower bound.} At any query $t\in[T]$ where $\abs{w_{t,1}},\abs{w_{t,2}}\le B$ (whether during startup, at a ready query, or inside a cooldown block), $\norm{\nabla F(X_t)}_*\ge u/4$.
\end{enumerate}
\end{lemma}

The following bound reduces the residual regime to five parameter checks. Its proof combines startup protection with control of the slow coordinates over the full horizon.

\begin{lemma}[Protected matrix construction]\label{lem:direct-protected}
Let $M\ge\lceil 16(pT+1)\rceil$ and set $\Lambda:=\log(64m(M+1))$. In addition to \eqref{eq:direct-geometry-condition}, assume
\begin{align}
 &pT>4,\qquad p\le\frac14,\qquad \frac{p}{\alpha}\le(10^4K)^{-1},
 \label{eq:direct-pulse-smallness}\\
 &pN(1+\Lambda)\le(128K)^{-1}, \label{eq:direct-reset-condition}\\
 &\frac{\eta u^2}{kA}\brac{J_0+NM+\frac2\alpha}\le\frac14,
 \label{eq:direct-slow-condition}\\
 &\frac{\eta^2u^4T}{k^2A^2p}\le 2^{-12}.
 \label{eq:direct-occupation-condition}
\end{align}
Then the construction has normalized initial gap and unit smoothness, is conditionally unbiased on every history, and satisfies
\begin{equation}\label{eq:direct-protected-risk}
 \expect{\min_{t\le T}\norm{\nabla F(X_t)}_*} \ge \frac{u}{8}.
\end{equation}
Its centered noise is zero except at ready queries, where it occupies the first column and equals the centered version of \eqref{eq:direct-sgd-pulse}.
\end{lemma}

\subsubsection{Proof of the SGDM Lower Bound}
\label{app:direct-sgdm-interpolation}

\begin{proof}[Proof of \cref{thm:direct-matrix-sgdm}]
For $V>0$, the table covers $d_T$ for all $m\ge2$ and $s_T$ for $m\ge4$. Here $\eta_0:=\sqrt{m/(VT)}$, $u$ is the target in \eqref{eq:direct-interpolation-u}, and $\gamma:=\eta$ if $r=\eta/\alpha\le128$ or $\gamma:=\alpha$ otherwise.
\begin{center}
\footnotesize
\setlength{\tabcolsep}{5pt}
\renewcommand{\arraystretch}{1.25}
\begin{tabular}{lll}
\toprule
\textbf{Branch and regime ($V>0$)} & \textbf{Hard instance} & \textbf{Lower bound} \\
\midrule
$d_T$: $\eta\ge 4$ & Periodic (\cref{lem:direct-step-obstructions}) & $=1\ge d_T$ \\
$d_T$: $0<\eta<4$, $\eta\ge \eta_0$ & Circular oracle (\cref{lem:direct-circular}) & $\Omega(d_T)$ \\
$d_T$: $0<\eta<4$, $\eta<\eta_0$ & Huber (\cref{lem:direct-step-obstructions}) & $\Omega(d_T)$ \\
\midrule
$s_T$: $\eta\ge 4$ & Periodic (\cref{lem:direct-step-obstructions}) & $=1\ge u$ \\
$s_T$: $0<\eta<4$, $\eta V\ge u^2$ & Circular oracle (\cref{lem:direct-circular}) & $\Omega(u)$ \\
$s_T$: $0<\eta<4$, $r\le 128$, $\eta\le m/(u^2T)$ & Huber (\cref{lem:direct-step-obstructions}) & $\Omega(u)$ \\
$s_T$: $0<\eta<4$, $r>128$, $\alpha\le m/(u^2T)$ & Inertia (\cref{lem:direct-inertia}) & $\Omega(u)$ \\
$s_T$: $0<\eta<4$, $\frac{m}{u^2T}<\gamma\le\eta<\frac{u^2}{V}$ ($m\ge 4$) & Cooldown pulses (\cref{lem:direct-protected}) & $\ge u/8$ \\
\bottomrule
\end{tabular}
\end{center}

\paragraph{Step 1: Deterministic branch $b_T$ and simple stochastic branch $d_T$.}
For every $(\eta,\beta)$, \cref{lem:direct-step-obstructions} yields a normalized deterministic instance with minimum nuclear gradient norm at least a universal multiple of $b_T=\min\{1,\sqrt{m/T}\}$. When $V=0$, we have $d_T=s_T=0$, so this already proves the normalized form of \eqref{eq:direct-sgdm-all-horizon}.

Assume $V>0$. For $d_T=\min\{1,(mV/T)^{1/4}\}$, compare $\eta$ with $h$ when $\eta<4$:
\begin{itemize}
    \item If $\eta\ge 4$, the periodic instance (\cref{lem:direct-step-obstructions}) gives $\min_{t\le T}\norm{\nabla F(X_t)}_*=1\ge d_T$.
    \item If $0<\eta<4$ and $\eta\ge \eta_0$, the circular oracle (\cref{lem:direct-circular}) gives
    \[
     \expect{\min_{t\le T}\norm{\nabla F(X_t)}_*}
     \ge \frac{1}{32}\min\bbrac{1,\,\sqrt{\eta V}}
     \ge \frac{1}{32}\min\bbrac{1,\,\brac{\frac{mV}{T}}^{1/4}}
     = \frac{d_T}{32}.
    \]
    \item If $0<\eta<4$ and $\eta<\eta_0$, the Huber instance (\cref{lem:direct-step-obstructions}) with $d=m$ gives
    \[
     \min_{t\le T}\norm{\nabla F(X_t)}_*
     = \sqrt{\frac{m}{\eta T+m/2}}
     \ge \frac{1}{\sqrt{\sqrt{T/(mV)}+1/2}}
     \ge \frac{d_T}{\sqrt2},
    \]
    since the right-hand side is at least $1$ when $\sqrt{T/(mV)}\le 1/2$ and exceeds $2^{-1/2}(mV/T)^{1/4}$ otherwise.
\end{itemize}

\paragraph{Step 2: Matrix pulse branch $s_T$ in the residual regime ($m\ge 4$).}
Fix $m\ge 4$, set $c_0:=2^{-40}$, and define
\begin{equation}\label{eq:direct-interpolation-u}
 u := c_0\min\bbrac{\ell_m^{-1/2},\,\brac{\frac{mV}{T}}^{1/6},\,\brac{\frac{m^2V}{T}}^{1/4}} = c_0 s_T.
\end{equation}
The three branches of $u$ control, respectively, the drift between resets, the motion of the slow coordinates during cooldown, and the sign imbalance at ready queries. We will use them in the form
\[
 u^2\ell_m\le c_0^2,\qquad \frac{u^6T}{mV}\le c_0^6,\qquad
 \frac{u^4T}{m^2V}\le c_0^4.
\]
Since $\ell_m>1$, we have $u<c_0<2^{-12}$. The definition of $\gamma$ gives $\gamma\le\eta$ and $\alpha\ge\gamma/128$. Outside the residual region:
\begin{itemize}
    \item If $\eta\ge 4$, the periodic instance (\cref{lem:direct-step-obstructions}) gives risk $1>u$.
    \item If $0<\eta<4$ and $\eta V\ge u^2$, the circular oracle (\cref{lem:direct-circular}) gives risk at least $\frac{1}{32}\min\{1,\sqrt{\eta V}\}\ge u/32$.
    \item If $0<\eta<4$ and $\gamma\le m/(u^2T)$, applying the Huber instance (\cref{lem:direct-step-obstructions}) when $r\le 128$ yields risk at least $\frac12\min\{1,\sqrt{m/(\eta T)}\}\ge u/2$, while applying the inertia obstruction (\cref{lem:direct-inertia}) when $r>128$ yields risk at least $\frac{1}{32}\min\{1,\sqrt{m/(\alpha T)}\}\ge u/32$.
\end{itemize}
It therefore remains to cover the residual regime
\begin{equation}\label{eq:direct-interpolation-residual}
 0<\eta<4, \qquad \frac{m}{u^2T}<\gamma\le\eta<\frac{u^2}{V}.
\end{equation}

In this region, instantiate the construction of \cref{app:direct-geometry,app:direct-reset-oracle} with $c_f:=(64K^2R_0^2)^{-1}$ and
\[
 p_i:=\frac{\eta^2\sigma_i^2}{64K^2R^2}, \qquad
 p:=\sum_{i=1}^m p_i, \qquad
 M:=\lceil 16(pT+1)\rceil, \qquad
 \Lambda:=\log(64m(M+1)).
\]
In the ordinary regime $(\eta/R,\eta/E)=(\gamma/R_0,\gamma)$; in the strong-momentum regime $R=R_0r$ and $r/E<28$. Thus both regimes give
\[
 \frac{\eta}{R}=\frac{\gamma}{R_0},\qquad
 p=c_f\gamma^2V,\qquad \frac{\eta}{E}\le 28\gamma.
\]
By \eqref{eq:direct-interpolation-residual}, $\gamma$ lies between $m/(u^2T)$ and $u^2/V$. Substituting these endpoints into $pT$, and using the fourth-power budget from \eqref{eq:direct-interpolation-u}, gives
\begin{equation}\label{eq:direct-interpolation-pulses}
 4<c_fc_0^{-4}\le\frac{c_fm^2V}{u^4T}<pT
   <\frac{c_fu^4T}{V}\le c_fc_0^4m^2.
\end{equation}
Hence $M\le 17+16pT\le 18m^2$. For $m\ge 4$, $e\cdot64m(M+1)\le 1216e\,m^3\le m^9$ since $1216e<4^6$. Taking logarithms gives $1+\Lambda\le 10\ell_m$.

\emph{Geometry and pulse probabilities.}
Since $\gamma V<u^2$, the geometry condition \eqref{eq:direct-geometry-condition} and pulse smallness \eqref{eq:direct-pulse-smallness} follow from $u<2^{-12}$, $pT>4$, and
\begin{align*}
 \frac{12K^2R^2p}{\eta E}
 &=\frac{3\eta V}{16E}\le\frac{21}{4}\gamma V
 <\frac{21}{4}u^2<\frac1{128},\\
 p\le\frac{p}{\alpha}
 &\le 128c_f\gamma V<128c_fu^2<(10^4K)^{-1}<\frac14.
\end{align*}
With these probabilities the centered column noise satisfies the conditional PSD bound \eqref{eq:direct-PSD-calibration}.

\emph{Drift between resets.}
Using $\alpha N<11$ from \eqref{eq:direct-cooldown}, the first budget gives \eqref{eq:direct-reset-condition}:
\[
 pN(1+\Lambda)=\frac{p}{\alpha}(\alpha N)(1+\Lambda)
 <128c_f\cdot 11\cdot 10\,u^2\ell_m
 \le 14080c_fc_0^2<(128K)^{-1}.
\]

\emph{Slow-coordinate motion.}
Use $k\ge m/6$, $r/E\le 128$, $\alpha J_0<4$, and $M\le 17+16c_fu^4T/V$ from \eqref{eq:direct-interpolation-pulses}. The constant part of the time budget contributes $u^2/m$; the number of pulses contributes the sixth-power term:
\begin{align*}
 \frac{\eta u^2}{kA}\brac{J_0+NM+\frac2\alpha}
 &=\frac{C_A(r/E)u^2}{k}\brac{\alpha J_0+(\alpha N)M+2}\\
 &\le 768C_A\frac{u^2}{m}(6+11M)
 \le 768C_A\brac{193\frac{u^2}{m}+176c_f\frac{u^6T}{mV}}.
\end{align*}
With $u<c_0$ and $u^6T/(mV)\le c_0^6$, the last bound is less than $1/4$, as required by \eqref{eq:direct-slow-condition}.

\emph{Sign occupation.}
Substitute $A=E/C_A$ and the definition of $p$. Since $k\ge m/6$ and $R/E\le 28R_0$, the fourth-power budget gives \eqref{eq:direct-occupation-condition}:
\[
 \frac{\eta^2u^4T}{k^2A^2p}
 =64C_A^2K^2\brac{\frac{R}{E}}^2\frac{u^4T}{k^2V}
 \le (48C_AK\cdot 28R_0)^2 c_0^4<2^{-12}.
\]
\Cref{lem:direct-protected} therefore applies and delivers risk at least $u/8=(c_0/8)s_T$, which proves the $s_T$ branch for $m\ge 4$.

\paragraph{Step 3: Small dimensions ($m\in\{2,3\}$), branch assembly, and scale restoration.}
When $m\in\{2,3\}$, since $\ell_m>1$, we have
\[
 s_T \le \min\bbrac{1,\,\brac{\frac{m^2V}{T}}^{1/4}}
     \le m^{1/4}\min\bbrac{1,\,\brac{\frac{mV}{T}}^{1/4}}
     \le 3^{1/4}d_T,
\]
so the simple stochastic branch $d_T$ from Step~1 already yields a universal multiple of $s_T$. For every $m\ge 2$, selecting the hardest among the three complete normalized instances ($b_T,d_T,s_T$) gives $\Rt_T^{\mathrm{SGDM}}\ge c\max\{b_T,d_T,s_T\}$ for a universal constant $c>0$. Applying \cref{app:matrix-rescaling} restores arbitrary $L_*,\Delta>0$, $\vSigma\in\SBB_+^m$, and $\WB_1\in\R^{m\times n}$, which establishes \eqref{eq:direct-sgdm-all-horizon}.

\paragraph{Step 4: Large-horizon simplification.}
Suppose now that $T\ge T_0=\lceil\max\{m,m^4v\}\rceil$. Because $T\ge m$, we have $b_T=\sqrt{m/T}$. If $V=0$, then $d_T=s_T=0$ and \eqref{eq:direct-sgdm-large} follows immediately from \eqref{eq:direct-sgdm-all-horizon}. Assume $V>0$, so that $v/T\le m^{-4}$. We check both comparisons in $s_T$:
\begin{itemize}
    \item Raising $(m^2v/T)^{1/4}$ and $(mv/T)^{1/6}$ to the twelfth power gives
    \[
     \frac{(m^2v/T)^3}{(mv/T)^2} = \frac{m^4v}{T} \le 1,
     \qquad\text{so}\qquad
     \brac{\frac{m^2v}{T}}^{1/4} \le \brac{\frac{mv}{T}}^{1/6}.
    \]
    \item For every integer $m\ge 2$, we have $e+m\le e^m$ and hence $\ell_m=\log(e+m)\le m$, so
    \[
     \brac{\frac{m^2v}{T}}^{1/4}
     \le (m^2\cdot m^{-4})^{1/4}
     = m^{-1/2}
     \le \ell_m^{-1/2}.
    \]
\end{itemize}
Therefore $s_T=(m^2v/T)^{1/4}$, and $d_T\le(mv/T)^{1/4}\le s_T$. Applying $\max\{b_T,d_T,s_T\}=\max\{b_T,s_T\}\ge\frac12(b_T+s_T)$ in \eqref{eq:direct-sgdm-all-horizon} establishes \eqref{eq:direct-sgdm-large} with $c_{\mathrm L}:=\min\{c/2,1\}$.
\end{proof}

\subsection{Tight Iteration Complexity of SGDM}
\label{app:sgdm-complexity}

We now prove the complexity characterization \eqref{eq:direct-sgdm-complexity} in \cref{thm:sgdm-tight-complexity} for every target accuracy $0<\epsilon\le\epsilon_4$, where with $c_{\mathrm L}\le 1$ from \eqref{eq:direct-sgdm-large} and $T_0$ from \eqref{eq:direct-sgdm-horizon} we define
\begin{equation}\label{eq:direct-sgdm-small-accuracy}
 \epsilon_4 := \frac{c_{\mathrm L}}{2}\sqbrac{
 \sqrt{\frac{mA_0}{T_0}} + \brac{\frac{m^2A_0V}{T_0}}^{1/4}}.
\end{equation}

\paragraph{Step 1: Ruling out short horizons $T\le T_0$.}
Fix any candidate query budget $T\le T_0$ and any fixed hyperparameter pair $(\eta,\beta)\in(0,\infty)\times[0,1)$. Let $(F,\O)$ be the hard instance supplied by \cref{thm:direct-matrix-sgdm} for that same pair $(\eta,\beta)$ at horizon $T_0$. Because the minimum over the first $T$ iterates is pathwise no smaller than the minimum over all $T_0$ iterates, \eqref{eq:direct-sgdm-large} gives
\[
 \expect{\min_{t\in[T]}\norm{\nabla F(\WB_t)}_*}
 \ge \expect{\min_{t\in[T_0]}\norm{\nabla F(\WB_t)}_*}
 \ge c_{\mathrm L}\sqbrac{\sqrt{\frac{mA_0}{T_0}}+\brac{\frac{m^2A_0V}{T_0}}^{1/4}}
 = 2\epsilon_4 > \epsilon.
\]
Hence no fixed-parameter \SGDMRef{} algorithm can guarantee worst-case expected minimum nuclear gradient norm at most $\epsilon$ with a budget $T\le T_0$.

\paragraph{Step 2: Matching lower and upper complexity bounds for $T>T_0$.}
For any horizon $T>T_0$, achieving worst-case risk at most $\epsilon$ under \eqref{eq:direct-sgdm-large} requires both $c_{\mathrm L}\sqrt{mA_0/T}\le\epsilon$ and $c_{\mathrm L}(m^2A_0V/T)^{1/4}\le\epsilon$, i.e.,
\[
 T \ge \max\bbrac{\frac{c_{\mathrm L}^2 mL_*\Delta}{\epsilon^2},\,
                  \frac{c_{\mathrm L}^4 m^2\norm{\vSigma}_F^2 L_*\Delta}{\epsilon^4}}
   = \Omega\brac{\frac{mL_*\Delta}{\epsilon^2}
                +\frac{m^2\norm{\vSigma}_F^2 L_*\Delta}{\epsilon^4}}.
\]
Conversely, with the explicit constant $C=4$ from \cref{app:direct-sgdm-upper}, choosing
\[
 T := \left\lceil\frac{4C^2 mL_*\Delta}{\epsilon^2}
                 +\frac{16C^4 m^2\norm{\vSigma}_F^2 L_*\Delta}{\epsilon^4}\right\rceil
   = \bigO{\frac{mL_*\Delta}{\epsilon^2}
          +\frac{m^2\norm{\vSigma}_F^2 L_*\Delta}{\epsilon^4}}
\]
in \cref{thm:direct-sgdm-upper} guarantees an average (and hence minimum) expected nuclear gradient norm at most $\epsilon/2+\epsilon/2=\epsilon$ (and at most $\epsilon/2$ when $V=0$). Combining the lower and upper bounds establishes \eqref{eq:direct-sgdm-complexity}.

\subsection{Proofs of Supporting Lemmas}
\label{app:sgdm-supporting-lemmas}

\begin{proof}[Proof of \cref{lem:direct-inertia}]
\paragraph{Construction.}
Set $d_0:=d-1\ge 1$, partition the coordinates as $(s,y_1,\ldots,y_{d_0})\in\R^d$, and define the slow average $w:=d_0^{-1}\sum_{i=1}^{d_0}y_i$. Initialize at the origin $x_1=\vzero$. With $\ell=7/16$ and $a=1/\ell=16/7$, set
\[
 D_{\mathrm{drv}}:=\frac{\eta}{14\alpha}, \qquad
 E_{\mathrm{drv}}:=D_{\mathrm{drv}}-2a, \qquad
 u:=\frac{1}{16}\min\bbrac{1,\,\sqrt{\frac{d_0}{\alpha T}}}, \qquad
 B:=\frac{12u}{d_0\ell}.
\]
Because $\alpha<\eta/128$, we have $\eta/\alpha>128$, so $D_{\mathrm{drv}}>128/14=64/7=4a$, which gives $E_{\mathrm{drv}}>2a$ and $E_{\mathrm{drv}}>\frac12 D_{\mathrm{drv}}=\eta/(28\alpha)$. Instantiate $\mathsf V$ from \eqref{eq:direct-driver} with $(D,E)=(D_{\mathrm{drv}},E_{\mathrm{drv}})$ and $J$ from \eqref{eq:direct-guard-potential} with slope $u$. Define the $C^1$ cutoff
\[
 P(z) := \begin{cases}
  1, & z \le -1/2, \\
  1-(z+1/2)^2, & -1/2 \le z \le 0, \\
  3/4-z, & 0 \le z \le 1/2, \\
  (1-z)^2, & 1/2 \le z \le 1, \\
  0, & z \ge 1,
 \end{cases}
\]
which satisfies $0\le P\le 1$, $-1\le P'\le 0$, $\abs{P''}\le 2$ on smooth pieces, and $P(0)=3/4$. Assemble the objective
\[
 \varphi(w):=P\!\brac{\frac{uw}{E_{\mathrm{drv}}}}, \qquad
 h(w):=J\!\brac{B+\frac{u}{\ell}-w}, \qquad
 f(s,y):=\mathsf V(s)\varphi(w)+h(w).
\]

\paragraph{Verification of initial gap and smoothness.}
Both $\mathsf V(s)\varphi(w)$ and $h(w)$ are nonnegative and vanish simultaneously whenever $w\ge\max\{E_{\mathrm{drv}}/u,\,B+u/\ell\}$, so $\inf f=0$. At the origin, $\mathsf V(0)=a/2=8/7$, $\varphi(0)=3/4$, and $h(0)=J(B+u/\ell)=uB+\frac{u^2}{2\ell}=(\frac{12}{d_0}+\frac12)\frac{u^2}{\ell}$. Using $d_0\ge 1$, $u\le 1/16$, and $\ell=7/16$,
\[
 f(\vzero)
 = \frac{3a}{8}+\brac{\frac{12}{d_0}+\frac12}\frac{u^2}{\ell}
 \le \frac{6}{7}+\frac{25}{2}\cdot\frac{1/256}{7/16}
 = \frac{6}{7}+\frac{25}{224}
 = \frac{31}{32} < 1.
\]
For smoothness, all first partial derivatives agree across piece joins because $\mathsf V,P,J\in C^1$. On each smooth piece, the second partials with respect to $(s,w)$ satisfy
\[
 \abs{f_{ss}} = \abs{\mathsf V''(s)\varphi(w)} \le \ell, \qquad
 \abs{f_{sw}} = \abs{\mathsf V'(s)\varphi'(w)} \le \frac{u}{E_{\mathrm{drv}}},
\]
\[
 \abs{f_{ww}} = \abs{\mathsf V(s)\varphi''(w)+h''(w)}
              \le E_{\mathrm{drv}}\cdot\frac{2u^2}{E_{\mathrm{drv}}^2}+\ell
              = \frac{2u^2}{E_{\mathrm{drv}}}+\ell.
\]
Since $w=d_0^{-1}\sum_{i=1}^{d_0}y_i$, each $y_i$-derivative carries a factor of $1/d_0$, so summing all entries of the $d\times d$ Hessian $\nabla^2 f$ yields
\[
 \sum_{p,q=1}^d \abs{\partial_{pq}f}
 = \abs{f_{ss}}+2\abs{f_{sw}}+\abs{f_{ww}}
 \le 2\ell+\frac{2u(1+u)}{E_{\mathrm{drv}}}
 < 2\ell+\ell u(1+u)
 \le \frac78+\frac{7\cdot 17}{4096} < 1,
\]
where the strict inequality uses $E_{\mathrm{drv}}>2a=2/\ell$. Integrating along line segments establishes global $\ell_\infty$-to-$\ell_1$ smoothness with constant at most $1$.

\paragraph{Driver acceleration and slow-coordinate bound.}
Globally, $\mathsf V(s)\abs{\varphi'(w)}\le E_{\mathrm{drv}}(u/E_{\mathrm{drv}})=u$ and $h'(w)=-J'(B+u/\ell-w)\in[-u,0]$, so $-2u\le f_w\le 0$ everywhere. Each coordinate gradient satisfies $\partial_{y_i}f=f_w/d_0\in[-2u/d_0,0]$, and since $m_{t,y_i}$ is a convex combination of past gradients under $m_1=g_1$, we have $-2u/d_0\le m_{t,y_i}\le 0$ for all $t\ge 1$. Summing $y_{t+1,i}=y_{t,i}-\eta m_{t,y_i}$ from $y_{1,i}=0$ and averaging over $i\in[d_0]$ gives
\[
 0 \le w_t \le \frac{2\eta u(t-1)}{d_0} \le \frac{2\eta uT}{d_0} \qquad (\forall\,t\in[T]).
\]
The linear part of $P$ extends to $w=E_{\mathrm{drv}}/(2u)$. Comparing this distance with the displacement above gives the scale $u^2\lesssim d_0/(\alpha T)$, since $E_{\mathrm{drv}}$ is of order $\eta/\alpha$. With the chosen $u$, we have $u^2T/d_0\le 1/(256\alpha)$ and $E_{\mathrm{drv}}>\eta/(28\alpha)$, so
\[
 0 \le \frac{uw_t}{E_{\mathrm{drv}}}
 \le \frac{2\eta u^2T}{d_0 E_{\mathrm{drv}}}
 < \frac{2\eta}{256\alpha}\cdot\frac{28\alpha}{\eta}
 = \frac{7}{32} < \frac14.
\]
Since $P(z)=3/4-z$ on $[0,1/2]$, it follows that
\[
 \varphi(w_t)\ge\frac12 \qquad\text{and}\qquad \varphi'(w_t)=-\frac{u}{E_{\mathrm{drv}}}
 \qquad (\forall\,t\in[T]).
\]

For the driver $s$, the samples are exact: $g_{t,s}=\mathsf V'(s_t)\varphi(w_t)$. Since $\varphi(0)=3/4$ and $0\le\varphi\le1$, the initial sample is $-3/4$, every sample is at most $1$, and each is zero on the plateau. The driver estimate in \cref{app:direct-inertia} therefore applies with $z=s$ and $D=D_{\mathrm{drv}}$.

\paragraph{Uniform gradient lower bound across all three regions.}
Since $\partial_{y_i}f=f_w/d_0$, every iterate satisfies $\norm{\nabla f(x_t)}_1=\abs{f_s(x_t)}+\abs{f_w(x_t)}$. We check three exhaustive regions for $s_t$ at any query $t\in[T]$:
\begin{itemize}
    \item \textbf{Region 1 ($s_t<2a$).} The driver estimate gives $t-1<4/(\eta\ell)$. The slow-coordinate bound then gives
    \[
     w_t \le \frac{2\eta u(t-1)}{d_0} < \frac{8u}{d_0\ell} < \frac{12u}{d_0\ell} = B.
    \]
    Hence $B+u/\ell-w_t>u/\ell$, so $h'(w_t)=-u$. Since $\mathsf V(s_t)\varphi'(w_t)\le 0$ has the same sign as $h'(w_t)$, we obtain $\abs{f_w(x_t)}\ge u$.

    \item \textbf{Region 2 ($2a\le s_t<D_{\mathrm{drv}}-a$).} Here $\mathsf V'(s_t)=1$, so $\abs{f_s(x_t)}=\varphi(w_t)\ge 1/2\ge u$.

    \item \textbf{Region 3 ($s_t\ge D_{\mathrm{drv}}-a$).} Because $\mathsf V$ is nondecreasing on $[a,\infty)$ and $E_{\mathrm{drv}}>2a$, we have $\mathsf V(s_t)\ge\mathsf V(D_{\mathrm{drv}}-a)=E_{\mathrm{drv}}-a/2>\frac34 E_{\mathrm{drv}}$. Since $\varphi'(w_t)=-u/E_{\mathrm{drv}}$ and $h'(w_t)\le 0$ share the same sign,
    \[
     \abs{f_w(x_t)} \ge \mathsf V(s_t)\abs{\varphi'(w_t)} > \frac34 E_{\mathrm{drv}}\cdot\frac{u}{E_{\mathrm{drv}}} = \frac{3u}{4}.
    \]
\end{itemize}
Across all queries $t\in[T]$, $\norm{\nabla f(x_t)}_1\ge 3u/4$. Because $d\ge 2$ implies $d_0=d-1\ge d/2$ and $3/(64\sqrt{2})>1/32$,
\[
 \min_{t\le T}\norm{\nabla f(x_t)}_1
 \ge \frac{3}{64}\min\bbrac{1,\,\sqrt{\frac{d_0}{\alpha T}}}
 \ge \frac{3}{64\sqrt{2}}\min\bbrac{1,\,\sqrt{\frac{d}{\alpha T}}}
 > \frac{1}{32}\min\bbrac{1,\,\sqrt{\frac{d}{\alpha T}}},
\]
which proves \eqref{eq:direct-inertia-risk}. Embedding $f$ along the leading $m$ diagonal entries of $\R^{m\times n}$ yields the corresponding normalized matrix obstruction.
\end{proof}

\begin{proof}[Proof of \cref{lem:direct-strong-geometry}]
Because $E\ge 1$, $R\ge R_0$, $E/R^2\le R_0^{-2}$, and \eqref{eq:direct-geometry-condition} holds, \cref{lem:direct-global-geometry} yields $Q/E\in[-3/C_A,\,1/128+3/C_A]$, $\norm{\nabla_s Q}_2\le h_g+3A\kappa$, $\abs{Q_{w_j}}\le u$ ($j\in\{1,2\}$), and $\operatorname{Lip}(\nabla Q)\le h_2+6A\kappa^2+4u\kappa+2u^2/A<1/200$, where $h_g:=4KRp/\eta$, $h_2:=10Kp/\eta$, and $\kappa:=\pi/(2R)$.

With $C_A=1024$, we obtain
\[
 C = \frac{25}{32}+\frac{Q}{E} \ge \frac{25}{32}-\frac{3}{1024} = \frac{797}{1024} > \frac34,
 \qquad
 C \le \frac{25}{32}+\frac{1}{128}+\frac{3}{1024} = \frac{811}{1024} < \frac{13}{16}.
\]
Both $\mathsf V(z)C$ and $h(w_0)$ are nonnegative and vanish at $z=a$, $w_0\ge B_0+u/\ell$, so $\inf F=0$. At $\bar X_1$, we have $w_1=w_2=0$, so $Q(\bar X_1)=H(KR\mathbf 1)\le 12K^2R^2p/\eta\le E/128$ by \eqref{eq:direct-H-bounds} and \eqref{eq:direct-geometry-condition}, and hence $C(\bar X_1)\le 25/32+1/128=101/128$. Using $\mathsf V(0)=a/2=8/7$, $k_0\ge 1$, $u\le 2^{-12}$, and $\ell=7/16$,
\[
 F(\bar X_1)
 \le \frac{101}{128}\cdot\frac87+\brac{\frac{12}{k_0}+\frac12}\frac{u^2}{\ell}
 \le \frac{101}{112}+\frac{25}{2}\cdot\frac{16}{7}\cdot 2^{-24} < 1.
\]

For smoothness, write $y:=(s,w_1,w_2)$ and fix any two matrix directions $\Delta X,\widetilde{\Delta X}\in\R^{m\times n}$ with $\norm{\Delta X}_{\op},\norm{\widetilde{\Delta X}}_{\op}\le 1$. Because each coordinate map in \eqref{eq:direct-coordinate-maps} contracts $\norm{\cdot}_{\op}$ into the Euclidean or absolute-value norm, the induced increments satisfy $\norm{\Delta s}_2,\abs{\Delta z},\abs{\Delta w_j}\le 1$ (and similarly for $\widetilde{\Delta X}$). On any smooth piece, differentiating $F(X)=\mathsf V(z)C(y)+h(w_0)$ by the product rule gives
\begin{align*}
 \mathrm D^2 F(X)[\Delta X,\widetilde{\Delta X}]
 &= \mathsf V''(z)C(y)\,\Delta z\,\widetilde{\Delta z}
    + h''(w_0)\,\Delta w_0\,\widetilde{\Delta w_0} \\
 &\quad + \mathsf V'(z)\brac{\Delta z\,\inner{\nabla C(y)}{\widetilde{\Delta y}}
        + \widetilde{\Delta z}\,\inner{\nabla C(y)}{\Delta y}} \\
 &\quad + \mathsf V(z)\,\mathrm D^2 C(y)[\Delta y,\widetilde{\Delta y}].
\end{align*}
Here $\abs{\mathsf V''}C\le\frac{13}{16}\ell$, $\abs{h''}\le\ell$, $\abs{\mathsf V'}\le 1$, and $\abs{\inner{\nabla C(y)}{\Delta y}}\le E^{-1}(\norm{\nabla_s Q}_2+\abs{Q_{w_1}}+\abs{Q_{w_2}})\le E^{-1}(h_g+3A\kappa+2u)$. For the last term, $\mathrm D^2 C=E^{-1}\mathrm D^2 Q$ and $0\le\mathsf V(z)/E\le 1$, while summing the $(s,s)$, $(s,w_j)$, $(w_j,s)$, and $(w_j,w_j)$ Hessian blocks of $Q$ via \eqref{eq:direct-signal-derivatives} and \eqref{eq:direct-H-bounds} yields
\[
 \abs{\mathsf V(z)\,\mathrm D^2 C(y)[\Delta y,\widetilde{\Delta y}]}
 \le \abs{\mathrm D^2 Q(y)[\Delta y,\widetilde{\Delta y}]}
 \le h_2+6A\kappa^2+4u\kappa+\frac{2u^2}{A},
\]
which is the block bound on $\operatorname{Lip}(\nabla Q)$ in \cref{lem:direct-global-geometry} and is strictly less than $1/200$. Combining all four terms gives
\begin{equation}\label{eq:direct-strong-hessian-bound}
 \abs{\mathrm D^2 F(X)[\Delta X,\widetilde{\Delta X}]}
 \le \ell\brac{1+\frac{13}{16}} + h_2+6A\kappa^2+4u\kappa+\frac{2u^2}{A}
     + \frac{2(h_g+3A\kappa)+4u}{E}.
\end{equation}
The terms after $\ell(1+13/16)=203/256$ in \eqref{eq:direct-strong-hessian-bound} form exactly the sum bounded in \eqref{eq:direct-hessian-bound}. Hence
\[
 \abs{\mathrm D^2 F(X)[\Delta X,\widetilde{\Delta X}]}
 < \frac{203}{256}+\frac{1}{200}<\frac45.
\]
Since all first derivatives agree across joining boundaries, integrating this bilinear bound along generic line segments (and passing to the limit via a vanishing generic translation for any segment contained in a joining hyperplane by continuity of $\nabla F$) establishes $\norm{\nabla F(X')-\nabla F(X)}_*\le\norm{X'-X}_{\op}$ globally.
\end{proof}

\begin{proof}[Proof of \cref{lem:direct-mark-gaps}]
For any active coordinate $i$ ($\hat p_i>0$), set $L_i:=\lceil\Lambda/\hat p_i\rceil$. If $L_i>M$, then every avoiding run within $I_1,\ldots,I_M$ has length at most $M<L_i$ deterministically. Otherwise, for every starting index $j_0\in\{0,1,\ldots,M-L_i\}$, the probability that the $L_i$ consecutive marks $I_{j_0+1},\ldots,I_{j_0+L_i}$ all avoid $i$ is at most
\[
 (1-\hat p_i)^{L_i} \le e^{-\hat p_i L_i} \le e^{-\Lambda} = \frac{1}{64m(M+1)}.
\]
Because there are at most $M+1$ valid starting indices for each $i\in[m]$, a union bound over all active coordinate-start pairs bounds the failure probability by $1/64$. On the complementary event, any avoiding run within $I_1,\ldots,I_M$ has integer length strictly less than $L_i$, hence at most $\Lambda/\hat p_i$. Adding at most one preceding cooling block from coordinate $i$'s own last pulse yields at most $1+\Lambda/\hat p_i$ cooling blocks. Coordinates with $\hat p_i=0$ have $p_i=0$, never receive pulses, and have zero saturated potential gradient.
\end{proof}

\begin{proof}[Proof of \cref{lem:direct-cooldown-tracking}]
At the first ready query, give every coordinate the target center $KR$. Each pulse in coordinate $i$ reverses its target sign $\sigma_i$. We show by induction that at each ready query $t$,
\[
 \abs{e_{t,j}}\le R/100,\qquad \abs{q_{t,j}-\sigma_j KR}<R/4
 \qquad (j\in[m]).
\]
The initial bounds hold by hypothesis. A zero column sample leaves $q_t$ fixed and sends $e_t$ to $\beta e_t$, so it suffices to consider a pulse in coordinate $i$ followed by the exact steps $t+1,\ldots,t+N$.

\paragraph{The coordinate hit by the pulse.}
Write $c_i=-\sigma_i KR$ for its new center. By \eqref{eq:direct-reset}, $q_{t+1,i}=c_i+e_{t,i}$. The ready guard gives $\abs{s_{t,i}}\le(K+1+1/100)R$, so
\[
 \abs{q_{t+1,i}-c_i}\le R/100,\qquad
 \abs{e_{t+1,i}}=\beta\abs{q_{t+1,i}-s_{t,i}}<(2K+2)R.
\]
Throughout the cooldown, $F=Q$ in the ordinary regime and $\mathsf V(z)/E=1$ in the strong-momentum regime. Thus the column gradient is $\nabla_s Q$ even when coordinate $i$ is not saturated. By \cref{lem:direct-global-geometry}, its norm is at most $G:=4KRp/\eta+5A/R$, since $3\kappa<5/R$. Both scale choices give $rE/R^2\le 128/R_0^2$: for $r\le 128$ the ratio is $r/R_0^2$, and for $r>128$ it is $E/(rR_0^2)<1/(14R_0^2)$. Hence \eqref{eq:direct-pulse-smallness} gives
\[
 \frac{\eta G}{\alpha R}
 =\frac{4Kp}{\alpha}+\frac{5rE}{C_A R^2}
 \le\frac4{10^4}+\frac{5\cdot128}{C_A R_0^2}<\frac1{2000}.
\]
The shadow moves by at most $\eta G$ per exact step. Since $\alpha N<11$, it stays close to the new center throughout the block:
\[
 \max_{1\le \iota\le N+1}\abs{q_{t+\iota,i}-c_i}
 \le R/100+\eta N G < (1/100+11/2000)R<R/50.
\]
Unrolling $e_{\tau+1,i}=\beta(e_{\tau,i}-\eta g_{\tau,s,i})$ over the same $N$ steps and using $\beta^N\le e^{-\alpha N}\le 1/(1000K)$ gives
\[
 \frac{\abs{e_{t+N+1,i}}}{R}
 \le (2K+2)\beta^N+\frac{\eta G}{R}\sum_{\iota=1}^{N}\beta^\iota
 \le \frac{2K+2}{1000K}+\frac1{2000}<\frac1{100}.
\]
Thus $\abs{s_{t+N+1,i}-c_i}<3R/100$: the physical coordinate has the new sign, lies in the saturated core, and satisfies the tracking guard.

\paragraph{Coordinates not hit by the pulse.}
For $j\ne i$, the pulse step leaves $q_{t+1,j}=q_{t,j}$ and $\abs{e_{t+1,j}}=\beta\abs{e_{t,j}}\le R/100$. Hence $s_{t+1,j}$ is within $(1/4+1/100)R$ of $\sigma_j KR$ and is saturated. Suppose it is still saturated at an exact step $\tau$ and $\abs{e_{\tau,j}}\le R/100$. Then $\eta\abs{g_{\tau,s,j}}\le 4KRp_j$ and \eqref{eq:direct-pulse-smallness} gives
\[
 \abs{e_{\tau+1,j}}
 \le \beta R/100+4\beta KRp_j
 \le (\beta+\alpha)R/100=R/100.
\]
At the end of its last own block, $j$'s shadow was within $R/50$ of its center by the pulsed-coordinate estimate (within $R/100$ at startup if it has never pulsed). The accumulated drift from all later blocks, up to and including step $\tau$, is at most $R/32$ by \eqref{eq:direct-local-drift}; it is zero for $p_j=0$. Thus
\[
 \abs{q_{\tau+1,j}-\sigma_j KR}\le R/50+R/32<R/4.
\]
The physical coordinate stays within $(1/4+1/100)R$ of its center, so it remains saturated with the same sign. This closes the induction through the end of the block. All coordinates satisfy \eqref{eq:direct-ready-guard} at $t+N+1$.
\end{proof}

\begin{proof}[Proof of \cref{lem:direct-startup-protection}]
\textbf{Part 1: Startup phase.}\enspace
In the ordinary regime ($r\le 128$), only $t=1$ is an exact startup query. At $\bar X_1$, $s_{1,i}=KR$ and $w_{1,1}=w_{1,2}=0$, so \eqref{eq:direct-H-saturated} gives $g_{1,s,i}=H_{s_i}(KR\mathbf 1)=2p_iKR/\eta$. With $m_{1,s}=g_{1,s}$,
\begin{align*}
 s_{2,i} &= KR-\eta g_{1,s,i} = KR(1-2p_i), \\
 q_{2,i} &= s_{2,i}-\frac{\eta\beta}{\alpha}g_{1,s,i} = KR-\frac{\eta}{\alpha}g_{1,s,i} = KR\brac{1-\frac{2p_i}{\alpha}}.
\end{align*}
By \eqref{eq:direct-pulse-smallness}, both $\abs{q_{2,i}-KR}=2Kp_iR/\alpha\le 2R/10^4<R/100$ and $\abs{e_{2,i}}=2\beta Kp_iR/\alpha<R/100$, so $t_{\mathrm{ready}}=2\le J_0$.

In the strong-momentum regime, the oracle perturbs only the control column, so its driver sample is always $g_{t,z}=F_z(X_t)=\mathsf V'(z_t)C(X_t)$. By \cref{lem:direct-strong-geometry}, $3/4<C<13/16$. Thus $m_{1,z}=-C(\bar X_1)<-3/4$, $g_{t,z}\le1$ globally, and $g_{t,z}=0$ on the plateau. The driver estimate in \cref{app:direct-inertia} gives a first plateau query $2\le\tau_D\le K_{\mathrm{acc}}+1<J_0$ and $z_t\ge D$ for all $t\ge\tau_D$.

Because $z_t<D$ for every $1\le t<\tau_D$, the oracle deterministically returns exact gradients on all updates $1\le t<\tau_D$. To verify column stability before any stochastic pulse can occur, suppose inductively that for a given $t\in\{1,\ldots,\tau_D-1\}$, every $\tau\in\{1,\ldots,t\}$ satisfies $\abs{s_{\tau,i}}\in[R,2KR]$ (which holds at $\tau=1$ since $s_{1,i}=KR$). For each such $\tau\le t$, saturation gives $\partial_{s_i}\Phi_1=\partial_{s_i}\Phi_2=0$, and $0\le\mathsf V(z_\tau)/E\le 1$ together with \eqref{eq:direct-H-saturated} gives $\abs{g_{\tau,s,i}}=\frac{\mathsf V(z_\tau)}{E}\abs{H_{s_i}(s_\tau)}\le 4KRp_i/\eta$. Since $m_{\tau,s,i}$ is a convex combination of $\{g_{1,s,i},\ldots,g_{\tau,s,i}\}$, we also have $\abs{m_{\tau,s,i}}\le 4KRp_i/\eta$. Using $t\le\tau_D-1<J_0$, $\alpha J_0\le 4$, and \eqref{eq:direct-pulse-smallness},
\[
 \abs{s_{t+1,i}-KR} \le \eta\sum_{\tau=1}^t\abs{m_{\tau,s,i}} \le 4KRp_i J_0 \le \frac{16KRp_i}{\alpha} \le \frac{16}{10^4}R,
\]
\[
 \abs{e_{t+1,i}} = \frac{\eta\beta}{\alpha}\abs{m_{t,s,i}} \le \frac{4KRp_i}{\alpha} \le \frac{4}{10^4}R < \frac{R}{100},
\]
and therefore $\abs{q_{t+1,i}-KR}\le\abs{s_{t+1,i}-KR}+\abs{e_{t+1,i}}\le\frac{20KRp_i}{\alpha}\le\frac{20}{10^4}R<\frac{R}{100}$. In particular, $s_{t+1,i}$ stays inside $[R,2KR]$, closing the induction through $t=\tau_D-1$. Its final update establishes $\abs{q_{\tau_D,i}-KR}<R/100$ and $\abs{e_{\tau_D,i}}<R/100$ before the sample at query $\tau_D$ is drawn, so the state at $\tau_D$ is ready and $t_{\mathrm{ready}}=\tau_D\in\{2,\ldots,J_0\}$.

\paragraph{Part 2: Uniform gradient protection.}
When $\abs{w_{t,1}},\abs{w_{t,2}}\le B$, both ramps have derivative one and $A/B=u$. The signal bound following \eqref{eq:direct-signals} therefore gives
\[
 \abs{Q_{w_1}(X_t)}+\abs{Q_{w_2}(X_t)}
 =u\brac{\abs{\Phi_1(s_t)}+\Phi_2(s_t)}\ge u/2
\]
even if a column coordinate is crossing zero. In the ordinary regime $F=Q$, so \eqref{eq:direct-gradient-compression} gives $\norm{\nabla F(X_t)}_*\ge u/2$.

In the strong-momentum regime ($r>128$), consider three exhaustive regions for $z_t$:
\begin{itemize}
    \item \textbf{If $z_t<2a$.} The driver estimate gives $t-1<4/(\eta\ell)$. Since $-u\le F_{w_0}=h'(w_0)\le0$, each diagonal coordinate $i\in I_0$ has momentum in $[-u/k_0,0]$. Hence
    \[
     w_{t,0} \le \frac{\eta u(t-1)}{k_0} < \frac{4u}{k_0\ell} < \frac{12u}{k_0\ell} = B_0.
    \]
    Thus $B_0+u/\ell-w_{t,0}>u/\ell$, so $\abs{F_{w_0}(X_t)}=u$, and \eqref{eq:direct-gradient-compression} yields $\norm{\nabla F(X_t)}_*\ge u>u/4$.
    \item \textbf{If $2a\le z_t<D-a$.} Here $\mathsf V'(z_t)=1$, so \eqref{eq:direct-gradient-compression} gives $\norm{\nabla F(X_t)}_*\ge\abs{F_z(X_t)}=C(X_t)>3/4>u/4$.
    \item \textbf{If $z_t\ge D-a$.} Since $\mathsf V$ is nondecreasing on $[a,\infty)$ and $E=D-2a>2a$, we have $\mathsf V(z_t)/E\ge(E-a/2)/E>3/4$. Applying \eqref{eq:direct-gradient-compression} yields
    \begin{align*}
     \norm{\nabla F(X_t)}_*
     &\ge \abs{F_{w_1}(X_t)}+\abs{F_{w_2}(X_t)}
      = \frac{\mathsf V(z_t)}{E}\brac{\abs{Q_{w_1}(X_t)}+\abs{Q_{w_2}(X_t)}} > \frac34\cdot\frac{u}{2}>\frac{u}{4}.
    \end{align*}
\end{itemize}
This covers every query $t\in[T]$.
\end{proof}

\begin{proof}[Proof of \cref{lem:direct-protected}]
Normalized initial gap and unit operator-norm smoothness follow from \cref{lem:direct-global-geometry} (for $r\le 128$) and \cref{lem:direct-strong-geometry} (for $r>128$) under \eqref{eq:direct-geometry-condition}, while global conditional unbiasedness was verified in \cref{app:direct-reset-oracle}.

\paragraph{Step 1: Coupling and high-probability event.}
Pre-sample i.i.d.\ $\operatorname{Bernoulli}(p)$ pulse indicators $\{\xi_j\}_{j\ge 1}$ on the auxiliary ready-query clock (with filtration $\mathcal H_0$ trivial and $\mathcal H_l:=\sigma(\xi_1,\ldots,\xi_l)$ for $l\ge 1$) together with an independent infinite sequence of i.i.d.\ coordinate marks $\{I_r\}_{r\ge 1}$ drawn with probabilities $\hat p_i=p_i/p$ (unused marks remain unrevealed; the $r$-th nonzero pulse uses $I_r$). Define $\zeta_1:=1$, $\zeta_{j+1}:=(-1)^{\xi_j}\zeta_j$ for $j\ge 1$, and partial sums $S_0:=0$ and $S_l:=\sum_{j=1}^l\zeta_j$ ($l\ge 1$). Define three events:
\begin{itemize}
    \item $\mathcal E_1:=\bbrac{\sum_{j=1}^T \xi_j\le M}$. Since $\E[\sum_{j=1}^T \xi_j]=pT$ and $M\ge 16pT$, Markov's inequality gives $\Pr(\mathcal E_1^c)\le pT/M\le 1/16$.
    \item $\mathcal E_2:=\bbrac{\max_{0\le l\le T}\abs{S_l}\le\frac{kA}{4\eta u^2}}$. Since $S_0=0$, applying \eqref{eq:direct-sign-maximal} on the auxiliary filtration $\{\mathcal H_l\}_{l=0}^T$ together with $pT>4$ from \eqref{eq:direct-pulse-smallness} gives
    \[
     \E\max_{l\le T}\abs{S_l}^2 \le \frac{2}{p^2}+\frac{8T}{p} < \frac{T}{2p}+\frac{8T}{p} = \frac{17T}{2p}.
    \]
    Applying Markov's inequality and \eqref{eq:direct-occupation-condition} yields
    \[
     \Pr(\mathcal E_2^c)
     \le \frac{16\eta^2u^4}{k^2A^2}\cdot\frac{17T}{2p}
     = \frac{136\eta^2u^4T}{k^2A^2p}
     \le \frac{136}{4096} < \frac{1}{16}.
    \]
    \item $\mathcal E_3$: the mark-gap event of \cref{lem:direct-mark-gaps} applied to the first $M$ marks $I_1,\ldots,I_M$, which satisfies $\Pr(\mathcal E_3^c)\le 1/64$.
\end{itemize}
By the union bound, the intersection $\mathcal E:=\mathcal E_1\cap\mathcal E_2\cap\mathcal E_3$ has probability
\[
 \Pr(\mathcal E) \ge 1-\brac{\frac{1}{16}+\frac{1}{16}+\frac{1}{64}} = \frac{55}{64}.
\]

\paragraph{Step 2: Exact EMA summation identity.}
For each slow signal group $j\in\{1,2\}$, write $a_{t,j}:=F_{w_j}(X_t)$ and let $\mu_{t,j}$ be its EMA ($\mu_{1,j}=a_{1,j}$ and $\mu_{t,j}=\beta\mu_{t-1,j}+\alpha a_{t,j}$ for $t\ge 2$). Because every diagonal entry $i\in I_j$ receives exact gradient $a_{t,j}/k_j$, its momentum is $\mu_{t,j}/k_j$, and averaging $X_{t+1,ii}=X_{t,ii}-\frac{\eta}{k_j}\mu_{t,j}$ over $i\in I_j$ gives $w_{t+1,j}=w_{t,j}-\frac{\eta}{k_j}\mu_{t,j}$. Rewriting $\mu_{t,j}=\beta\mu_{t-1,j}+\alpha a_{t,j}$ as $\alpha\mu_{t,j}=\alpha a_{t,j}+\beta(\mu_{t-1,j}-\mu_{t,j})$ for $t\ge 2$, dividing by $\alpha$, summing over $t=2,\ldots,l$, and adding $\mu_{1,j}=a_{1,j}$ yields
\begin{align}
 w_{t+1,j} &= w_{t,j}-\frac{\eta}{k_j}\mu_{t,j}, \nonumber\\
 \sum_{t=1}^l\mu_{t,j}
 &= a_{1,j}+\sum_{t=2}^l\brac{a_{t,j}+\frac{\beta}{\alpha}(\mu_{t-1,j}-\mu_{t,j})}
  = \sum_{t=1}^l a_{t,j}+\frac{\beta}{\alpha}(a_{1,j}-\mu_{l,j}).
 \label{eq:direct-ema-sum}
\end{align}

\paragraph{Step 3: Slow-coordinate safety and risk bound.}
Fix a sample path in $\mathcal E$. Globally, $\abs{a_{\tau,j}}\le u$ and hence $\abs{\mu_{\tau,j}}\le u$ for all $\tau\ge 1$ and $j\in\{1,2\}$. Because $w_{1,j}=0$ and $k_j\ge k$, summing $w_{\tau+1,j}-w_{\tau,j}=-\frac{\eta}{k_j}\mu_{\tau,j}$ immediately gives, for every $1\le h\le J_0$,
\[
 \abs{w_{h,j}}
 \le \frac{\eta u(h-1)}{k_j}
 \le \frac{\eta u J_0}{k}
 \le \frac{B}{4} < B,
\]
where the third inequality uses \eqref{eq:direct-slow-condition} with $B=A/u$. Hence both slow coordinates stay strictly inside $[-B,B]$ throughout startup, so \cref{lem:direct-startup-protection} (Part~1) applies directly and guarantees that the first ready query occurs at some $t_{\mathrm{ready}}\in\{2,\ldots,J_0\}$ with $\abs{q_{t_{\mathrm{ready}},i}-KR}\le R/100$ and $\abs{e_{t_{\mathrm{ready}},i}}\le R/100$ for all $i\in[m]$.

Now consider any query $t\in\{2,\ldots,T\}$ up to a putative first exit of $(w_{\tau,1},w_{\tau,2})$ from $[-B,B]$, so that $\abs{w_{\tau,1}},\abs{w_{\tau,2}}\le B$ for all $1\le\tau\le t-1$. By \cref{lem:direct-startup-protection} (Part~1) and \cref{lem:direct-cooldown-tracking}, readiness never fails during queries $1,\ldots,t-1$: at most $M$ pulses occur (on $\mathcal E_1$), and each pulse flips the sign of exactly one column coordinate by the next ready query, so at the $j$-th ready query the column sign product equals $\zeta_j$. At every ready query $\tau$ (where $s_\tau$ is saturated and $\mathsf V(z_\tau)/E=1$ when $r>128$), it follows that $a_{\tau,1}=u\Phi_1(s_\tau)=u\zeta_j$ and $a_{\tau,2}=u\Phi_2(s_\tau)=0$.

Among queries $1,\ldots,t-1$, at most $(t_{\mathrm{ready}}-1)+NM<J_0+NM$ are non-ready. Let $l_t\in\{0,1,\ldots,t-1\}$ denote the number of ready queries among $\{1,\ldots,t-1\}$ (with $l_t=0$ if $t\le t_{\mathrm{ready}}$, in which case $S_{l_t}=S_0=0$ and all $t-1<J_0$ queries are non-ready). Since $t-1\ge 1$, applying \eqref{eq:direct-ema-sum} with $l=t-1$, $w_{1,j}=0$, $k_j\ge k$, and $\abs{\frac{\beta}{\alpha}(a_{1,j}-\mu_{t-1,j})}\le 2u/\alpha$ gives
\begin{align*}
 \abs{w_{t,1}}
 &= \frac{\eta}{k_1}\abs{\sum_{\tau=1}^{t-1}\mu_{\tau,1}}
  \le \frac{\eta u}{k}\sqbrac{\abs{S_{l_t}}+J_0+NM+\frac{2}{\alpha}} \\
 &\le \frac{\eta u}{k}\max_{0\le l\le T}\abs{S_l} + \frac{\eta u}{k}\brac{J_0+NM+\frac{2}{\alpha}}
  \le \frac{B}{4}+\frac{B}{4} = \frac{B}{2},
\end{align*}
where the last inequality uses $\mathcal E_2$ for the first term and \eqref{eq:direct-slow-condition} (with $B=A/u$) for the second. Similarly, since $a_{\tau,2}=0$ at every ready query,
\[
 \abs{w_{t,2}} \le \frac{\eta u}{k}\brac{J_0+NM+\frac{2}{\alpha}} \le \frac{B}{4}.
\]
Because $\abs{w_{t,1}}\le B/2<B$ and $\abs{w_{t,2}}\le B/4<B$, a first exit from $[-B,B]$ cannot occur. Therefore $\abs{w_{t,1}},\abs{w_{t,2}}\le B$ holds for every $t\in[T]$ on $\mathcal E$, and \cref{lem:direct-startup-protection} (Part~2) guarantees $\min_{t\le T}\norm{\nabla F(X_t)}_*\ge u/4$ everywhere on $\mathcal E$. Taking expectations gives
\[
 \expect{\min_{t\le T}\norm{\nabla F(X_t)}_*}
 \ge \Pr(\mathcal E)\cdot\frac{u}{4}
 \ge \frac{55}{64}\cdot\frac{u}{4} > \frac{u}{8},
\]
which proves \eqref{eq:direct-protected-risk}.
\end{proof}

\section{Experimental Details}\label{sec:experimental-setup}

\paragraph{Architectures and training configurations.}
All pretraining runs use a fixed random seed of 42 and follow standard benchmark protocols:
\begin{itemize}
    \item \textbf{ResNet-20 on CIFAR-10.} Following~\citet{bernstein2018signsgd}, we train ResNet-20~\citep{he2016deep} on CIFAR-10~\citep{krizhevsky2009learning} for 160 epochs with a batch size of 128, momentum $\beta=0.9$, and a step-decay learning rate schedule (decayed at epoch 80). For each optimizer, the peak learning rate is tuned over $\{10^{-4}, 5\times 10^{-4}, 10^{-3}, 5\times 10^{-3}, 10^{-2}\}$. Each 4D convolutional kernel is reshaped into a 2D matrix by flattening spatial dimensions into the input-channel axis before applying \MuonRef{}.
    \item \textbf{nanoGPT-124M and LLaMA-130M on C4.} Both models are trained on the C4 dataset~\citep{JMLR:v21:20-074} using the GPT-2 tokenizer (vocabulary size $50{,}304$) for $20{,}000$ steps with sequence length $512$ and global batch size $256$ ($2.6$B training tokens). nanoGPT~\citep{nanogpt} comprises 12 Transformer blocks with hidden dimension $768$ and 12 attention heads; LLaMA-130M~\citep{touvron2023llama,lialin2023relora,zhao2024galore} comprises 13 blocks of the same width and head count, using RMSNorm, rotary position embeddings (RoPE), and SwiGLU feed-forward layers. Both models tie input and output embeddings. Following a grid search over $\{10^{-4}, 5\times 10^{-4}, 10^{-3}, 5\times 10^{-3}, 10^{-2}\}$, the peak learning rates are set to $10^{-2}$ for \SGDMRef{} and $10^{-3}$ for \MuonRef{}, with momentum $\beta=0.9$, a $1{,}000$-step linear warmup, and cosine decay.
    \item \textbf{LLaMA-1.34B on C4.} The 1.34B model~\citep{touvron2023llama,zhao2024galore} consists of 24 Transformer blocks with hidden dimension $2048$, 32 attention heads, and SwiGLU intermediate dimension $5632$ (with RMSNorm, RoPE, and tied embeddings). We train from scratch for $7{,}630$ steps with sequence length $512$ and global batch size $256$ ($1.0$B training tokens), using momentum $\beta=0.9$, peak learning rate $10^{-3}$, a $400$-step warmup, and cosine decay to $10^{-5}$.
\end{itemize}
Across all benchmarks, \MuonRef{} approximates $\msign{\MB_t}$ using five Newton--Schulz iterations~\citep{jordan2024muon} on all 2D hidden weight matrices (with update RMS scaling of $0.2$ on the $1.34$B model), while 1D parameters, embeddings, and output heads are updated using standard vector optimizers (Adam with $\beta_1=0.9, \beta_2=0.95$ on the language models). Training and validation losses report token-averaged cross-entropy on held-out C4 validation sequences.

\paragraph{Matrix-noise measurement protocol.}
To compute $\widehat{C}_b$ and $\chi_b$ in~\eqref{eq:empirical-matrix-noise}, we freeze model checkpoints at $25\%$, $50\%$, and $75\%$ of training for ResNet-20, nanoGPT-124M, and LLaMA-130M, and at $25\%$, $50\%$, $75\%$, and $100\%$ of training for LLaMA-1.34B. At each checkpoint, we disable dropout and evaluate raw minibatch gradients of the data loss at the training batch size ($128$ images for ResNet-20 and $256$ sequences for the C4 models), aggregating $K = 10{,}557$ minibatch gradients per checkpoint for ResNet-20, $K = 10{,}240$ for nanoGPT-124M and LLaMA-130M, and $K = 256$ global-batch gradients for LLaMA-1.34B. Biases, normalization parameters, embeddings, and output heads are excluded, yielding $22$ weight matrices in ResNet-20, $48$ matrices in nanoGPT-124M ($m_b = 768$), $65$ matrices in LLaMA-130M ($m_b = 768$), and $120$ matrices in LLaMA-1.34B ($m_b = 2048$, comprising \texttt{QKV}, \texttt{Out}, \texttt{w1}, \texttt{w2}, and \texttt{Down} in each layer).

\paragraph{Open-source Pythia-6.9B scaling probe.}
To examine matrix-noise concentration at the 7B scale ($m_b = 4096$) over a full pretraining horizon, we evaluate open checkpoints of Pythia-6.9B~\citep{biderman2023pythia} at mid-training (\texttt{step72000}) and completion (\texttt{step143000}). At each frozen checkpoint, we disable dropout, compute raw data-loss minibatch gradients on The Pile~\citep{gao2020pile} (batch size $256$ and sequence length $2048$), and aggregate $K = 256$ merged batches (from two independent pools of $128$ samples) across all 32 layers ($4$ matrices per layer, namely \texttt{QKV}, \texttt{Out}, \texttt{Up}, and \texttt{Down}, totaling $128$ matrices per checkpoint). Defining the normalized cumulative singular-value spectrum
\[
S(k) = \frac{\sum_{j \le k} \sqrt{\lambda_j(\widehat C_b)}}{\sum_{j \le 4096} \sqrt{\lambda_j(\widehat C_b)}},
\]
we find that while \texttt{Out}, \texttt{Up}, and \texttt{Down} closely track the $4096$-dimensional isotropic baseline, \texttt{QKV} concentrates $6.0\%$ (\texttt{step72000}) and $6.2\%$ (\texttt{step143000}) of its total nuclear norm in the single leading singular direction (compared with $0.024\%$ for a uniform spectrum) and $12.7\%\text{--}13.9\%$ within the top 64 directions. Subsampling diagnostics confirm strong estimation stability, as the two independent $K=128$ pools agree within $10\%$ relative variation on $118/128$ matrices at \texttt{step72000} and $128/128$ matrices at \texttt{step143000}, and over $90\%$ of matrices ($116/128$ and $119/128$, respectively) remain within $10\%$ across all $K=64$, $K=128$, and merged $K=256$ comparisons.

\clearpage
\bibliography{ref}

\end{document}